\newcommand{\coloption}{2}

\ifnum\coloption=1
  \documentclass[onecolumn,draftclsnofoot,a4paper, 12pt]{IEEEtran}
  \usepackage[a4paper,%
            top=20mm,    
            bottom=20mm, 
            left=20mm,   
            right=20mm   
           ]{geometry}
\else\ifnum\coloption=2
  \documentclass[a4paper]{IEEEtran}
  \usepackage[a4paper,%
            top=18mm,    
            bottom=18mm, 
            left=18mm,   
            right=18mm   
           ]{geometry}
\else
  \PackageError{ColumnOption}{Invalid column option '\coloption'. Use 1 or 2}{}
\fi\fi

\usepackage{graphicx}
\usepackage{amsmath,amssymb}
\usepackage{cite}
\usepackage[dvipsnames,table]{xcolor}  

\usepackage{caption}
\usepackage{siunitx}
\usepackage{longtable}
\usepackage{array}
\usepackage{kotex}

\usepackage{soul}
\usepackage{booktabs}
\usepackage{makecell}
\usepackage{multirow}
\usepackage{array}
\usepackage{ragged2e} 

\usepackage{lipsum}
\usepackage{setspace}
\usepackage{pifont}          
\usepackage{tikz}            
\usetikzlibrary{shapes.geometric, arrows, positioning, calc}
\usepackage{enumitem} 
\setlist[itemize]{leftmargin=3pt}

\usepackage{subcaption}
\usepackage{adjustbox}

\usepackage[normalem]{ulem}

\usepackage{longtable}
\usepackage{tabularx} 
\usepackage[most]{tcolorbox}       
\usepackage{upgreek}
\tcbuselibrary{skins}

\newcommand{\tcb}{\textcolor{black}}

\usepackage{multirow}

\begin{document}

\title{Advancing Multi-Robot Networks via MLLM-Driven Sensing, Communication, and Computation: A Comprehensive Survey}

\author{Hyun~Jong~Yang, \textit{Senior Member,~IEEE}, Howon~Lee, \textit{Senior Member,~IEEE}, Kyuhong~Shim, \textit{Member,~IEEE}, Jeongho~Kwak, \textit{Member,~IEEE}, Hyunsoo~Kim, \textit{Student Member,~IEEE}, Donghoon~Kim, \textit{Student Member,~IEEE}, Khoa~Anh~Ngo, \textit{Student Member,~IEEE}, Sehyun~Ryu, Jaehyun~Choi, Youbin~Kim, \textit{Student Member,~IEEE}, Chanjun~Moon, Michael~Ryoo, and Byonghyo~Shim,~\textit{Fellow,~IEEE}\\
\thanks{%
H. J. Yang, H. Kim, D. Kim, K. A. Ngo, J. Choi, and B. Shim are with the Department of Electrical and Computer Engineering, Seoul National University, Seoul, South Korea (Emails: H. J. Yang, K. A. Ngo, J. Choi, and B. Shim (\texttt{\{hjyang, 2019-20227, jhchoi0226, bshim\}@snu.ac.kr}), H. Kim and D. Kim (\texttt{\{hskim, dhkim\}@islab.snu.ac.kr})).
H. Lee is with the Department of Electrical and Computer Engineering, Ajou University, Suwon, South Korea (Email: \texttt{howon@ajou.ac.kr}).
Y. Kim and C. Moon are with the Department of AI Convergence Network, Ajou University, Suwon, South Korea 
(Emails: \texttt{\{youbin1323, mcjmcj0412\}@ajou.ac.kr}).
K. Shim is with the Department of Intelligent Software, College of Computing and Informatics, Sungkyunkwan University, Suwon, South Korea (Email: \texttt{khshim@skku.edu}).
J. Kwak is with the Department of Computer Science and Engineering, Korea University, Seoul, South Korea (Email: \texttt{jeonghokwak@korea.ac.kr}).
S. Ryu is with the Department of Electrical Engineering, POSTECH, Pohang, South Korea (email: \texttt{sh.ryu@postech.ac.kr}).
M. Ryoo is with the Department of Computer Science and the AI Institute, Stony Brook University, Stony Brook, NY 11794, USA (email: \texttt{mryoo@cs.stonybrook.edu}).%
}
\thanks{Corresponding authors: B. Shim and H. Lee.}}

\maketitle

\begin{abstract}
Imagine a near future where advanced humanoid robots, powered by multimodal large language models (MLLMs), effortlessly interpret real-time sensing data, not just from their own sensors, but also from neighboring drones, autonomous vehicles, or underwater vehicles. Such robots for the physical AI are already taking shape in laboratories, hinting at imminent deployment across industries to tackle tasks like warehouse logistics, manufacturing, factory assembly, precision agriculture, public-space assistance, and on-site medical or safety rescue. While a single robot can demonstrate impressive local autonomy, realistic missions demand holistic coordination among multiple agents, compelling them to share and jointly interpret vast streams of sensor information.
In these high-stakes scenarios, communication is indispensable, since without robust links, each robot remains blind to the broader mission context and cannot leverage the combined intelligence offered by a collective MLLM. Yet transmitting comprehensive sensor data from dozens or hundreds of robots, each with its own bandwidth and latency constraints, can overwhelm networks. This challenge is exacerbated when a system-level orchestrator or cloud-based MLLM needs to fuse multimodal inputs to generate holistic decisions, like route planning, anomaly detection, or real-time adjustments to complex collaborative tasks. Crucially, these tasks are often initiated by high-level natural language instructions (e.g., ``Search for the yellow bin''). This text-based intent serves as a powerful filter for resource optimization: by understanding the specific goal via MLLMs, the system can selectively activate only the relevant sensing modalities, dynamically allocate communication bandwidth, and determine the optimal computation placement. This ``intent-to-resource'' mapping capability is a fundamental motivation for the proposed unified architecture. Moreover, many real deployments require open-vocabulary perception and language-grounded action (e.g., recognizing previously unseen objects or infrastructure outside the robot's field-of-view), which is difficult to achieve with closed-set on-device perception alone. Viewed this way, R2X is fundamentally an \emph{intent-to-resource orchestration} problem: given a high-level language command and system context, the network must \emph{jointly optimize} sensing, wireless communication, and computation so that task-level success is maximized under resource constraints.

This survey examines how {integrated sensing, communication, and computation} design paves the way for effective multi-robot coordination under MLLM guidance. We begin by reviewing state-of-the-art sensing modalities (from bounding-box LiDAR to hyperspectral imaging) and their interplay with semantic or partial-compression techniques. Next, we analyze communication strategies, including low latency protocols, edge-based orchestration, and adaptive resource allocation, to ensure scalable and timely data delivery at scale. We then explore distributed, hybrid, and fully centralized computing approaches, highlighting how large-model reasoning can be split among on-device distillation and powerful edge/cloud servers.
To ground these concepts in measurable outcomes, we further present \emph{four} end-to-end demonstrations that make the orchestration loop explicit (sense $\rightarrow$ communicate $\rightarrow$ compute $\rightarrow$ act): (i) digital-twin warehouse navigation with semantic sensing and predictive link context, (ii) mobility-driven proactive MCS control under delayed feedback, (iii) a real FollowMe robot with a practical semantic-sensing switch over WiFi, and (iv) real-hardware open-vocabulary trash sorting where an edge-assisted MLLM grounds text instructions to unseen objects and out-of-FOV bins via multi-view sensing.

Ultimately, our goal is to guide the robotics research community and industry practitioners in devising end-to-end solutions that integrate sensing, communication, and computation for real-time and large-scale multi-robot operations. By uniting advanced sensors, flexible network design, and MLLM-based intelligence, autonomous robot teams can achieve a truly holistic awareness and responsiveness in the complex environments of tomorrow.
Across the demonstrations, we emphasize system-level metrics---payload size, end-to-end latency, reliability, and task success---to clarify when and why R2X-style edge/communication-assisted orchestration provides tangible advantages over purely on-device baselines.
\end{abstract}


\section{Introduction} \label{sec:intro}

Humanoid robots equipped with large language model (LLM) have advanced to a stage where they can interpret textual instructions, multimodal sensor data, and contextual cues to operate with remarkable adaptability. Figure AI’s humanoid \cite{figure02} and Tesla’s Optimus \cite{tesla_optimus} are leading examples of such systems. By leveraging LLM-based semantics and real-time feedback, robots can navigate highly dynamic environments and handle complicated tasks ranging from logistics, manufacturing, to human assistance. While single-robot autonomy has improved significantly, real-world industrial demands require fleets of robots cooperating in real time, inevitably bringing communication-enabled collaboration to the forefront. Crucially, multi-robot missions are often specified as high-level \emph{language} objectives (e.g., ``find the yellow bin and bring it here''), not as low-level control scripts.
Such language intent should be translated into concrete system decisions---\emph{what to sense}, \emph{what to transmit}, and \emph{where to compute}---so that limited sensing/compute/wireless resources are used efficiently to maximize task success.

\begin{figure}[t]
    \centering
    \includegraphics[width=0.95\linewidth]{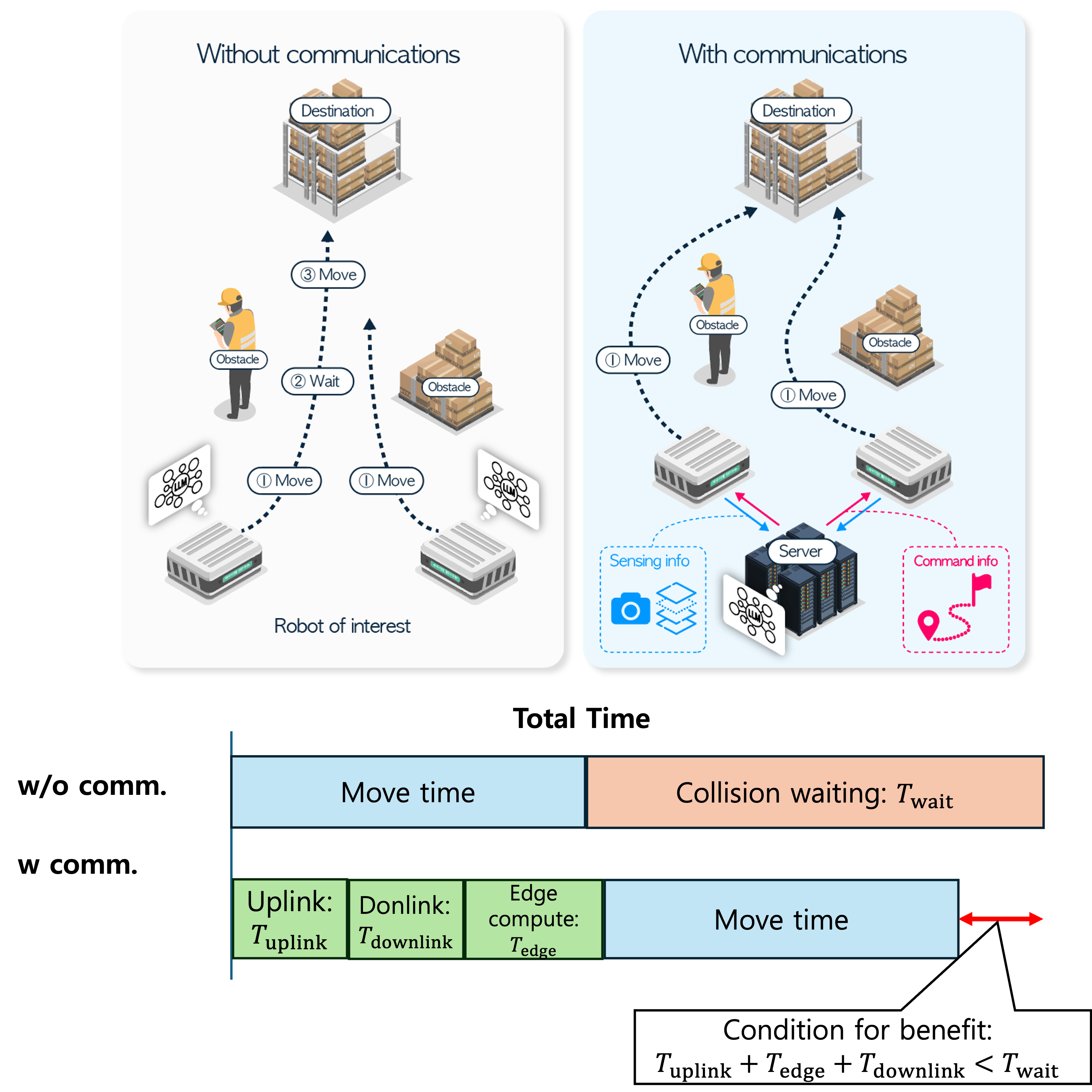}
    \caption{A scenario comparing (left) individually planned routes vs.\ (right) communication-based collaboration. The lower timeline highlights that the gain is conditional: ``with communications'' outperforms when $T_{\mathrm{uplink}}+T_{\mathrm{edge}}+T_{\mathrm{downlink}} < T_{\mathrm{wait}}$, where $T_{\mathrm{wait}}$ is the expected collision-induced waiting time without coordination.}
    \label{fig:showcase}
\end{figure}

Figure~\ref{fig:showcase} illustrates how such collaboration significantly reduces total task completion time. On the left, two autonomous robots plan their routes independently, causing one robot to stop due to the overlapping paths. Obviously, this idle delay will increase overall latency. On the other hand, each robot in the right-hand uploads sensing data (e.g., depth maps of the environment) to the centralized computing node to perform the path optimization. Although an extra step of transmitting sensing information to the centralized node is needed due to the offloading of computation, two non-intersecting trajectories can be generated, thereby minimizing the latency of robot movement. The lower portion of Figure~\ref{fig:showcase} illustrates timelines, demonstrating ``with communications" outperforms ``without communications".
Importantly, this advantage is \emph{conditional}: as visualized by the callout in Fig.~\ref{fig:showcase}, communication-assisted offloading is beneficial when the offloading critical path satisfies $T_{\mathrm{uplink}}+T_{\mathrm{edge}}+T_{\mathrm{downlink}} < T_{\mathrm{wait}}$. Here, $T_{\mathrm{edge}}$ and $T_{\mathrm{wait}}$ denote the edge-side planning/optimization time and  expected collision-induced waiting time without coordination, respectively.

Despite the seemingly straightforward benefit of coordinated routing, design of multi-robot systems is quite involved since it should consider sensing, communication, and computation as a whole. The sensing side dictates how raw data (e.g., camera feeds, LiDAR scans, haptic readings) is collected and then compressed for efficient transmission. In this step, semantic sensing and feature vector encoding might be used to reduce the bandwidth while preserving essential information. The communication architecture, in turn, must handle dynamic multimodal data flows to ensure low-latency communications of time-critical tasks. Whether a robot should offload large-scale perception tasks or keep them local depends on the complexity of given task and resource constraints, such as CPU/GPU/memory capacity, bandwidth, and power/energy budget. Task-centric communication requirements should thus be adapted dynamically, to guarantee that high-priority data and commands are delivered in time. To make this interplay possible, all system layers including physical and network layers as well as computational layer need to be properly designed to judge whether given task is served at the edge or cloud, while reactive control loops remain on-device. Importantly, the objective in R2X is not to maximize link throughput per se, but to \emph{execute the user/task intent as reliably as possible}.
\tcb{The system-level orchestrator therefore solves this \emph{intent-conditioned joint optimization} by jointly configuring sensing, communication, and computation to satisfy task-facing QoS under time-varying resource constraints.}

These are some recent studies addressing parts of this problem, but they do not provide holistic and satisfactory results. For instance, machine-to-machine (M2M) \cite{Verma201683,Zhao2017M2MChallenges} focuses on generic device connectivity but excludes large-model orchestration or goal-driven collaboration. Vehicle-to-everything (V2X) \cite{Mande2024V2X6G} addresses low-latency traffic safety issue but the scope is limited to the vehicles. Ultra-reliable low-latency communication (URLLC) \cite{Haque2024URLLC,Liu2023MLfor6GURLLC,Shim_URLLC} promises sub-millisecond latency but is not tightly coupled with AI-based decision-making or multi-robot sensor fusion.
(We use the standard 3GPP term \emph{URLLC} throughout this paper to avoid ambiguity.)
Robot collaboration works \cite{wu2024stateoftheartrobotlearningmultirobot,nair2024collaborativeperceptionmultirobotsystems,machines12080540,baratta2023hrc} highlight distributed learning and cooperative perception but do not consider the physical layer and network layer design issues in communications systems.
\tcb{Table~\ref{tab:table_r2x_comparison} systematically compares these paradigms across six criteria that collectively define the scope of a robotic networking framework. The central observation is that no existing paradigm simultaneously satisfies all six criteria: robot collaboration lacks communication-layer design; V2X and M2M lack multi-robot AI synergy; and URLLC lacks task-facing orchestration. R2X is the only paradigm that spans the full stack from physical-layer wireless protocols to MLLM-based semantic reasoning, making a unified survey and framework essential.}

\begin{table*}[t]
\centering
{\fontsize{12pt}{14.4pt}\selectfont
\caption{Comparison of robot collaboration, V2X, M2M, URLLC, and the proposed R2X Paradigm.}
\label{tab:table_r2x_comparison}
\resizebox{\textwidth}{!}{
\begin{tabular}{p{2.2cm} p{10.8cm} c c c c c c p{3.0cm}}
\toprule
\textbf{Paradigm} 
& \textbf{Defining scope and features}  
& \parbox[c]{2.2cm}{\centering \textbf{R2R comm.}\\ \textbf{protocol}} 
& \parbox[c]{2.2cm}{\centering \textbf{Ultra-low} \\ \textbf{latency} \\ \textbf{support}} 
& \parbox[c]{2.2cm}{\centering \textbf{Consideration} \\ \textbf{of multiple} \\ \textbf{robot collab.}} 
& \parbox[c]{2.2cm}{\centering \textbf{Task-central} \\ \textbf{comm.} \\ \textbf{reqs. design}} 
& \parbox[c]{2.2cm}{\centering \textbf{Integrated} \\ \textbf{sensing/} \\ \textbf{computing/} \\ \textbf{comm. design}} 
& \parbox[c]{2.2cm}{\centering \textbf{Inclusion of} \\ \textbf{LLM-Based} \\ \textbf{AI synergy}} 
& \textbf{Relevant surveys} \\
\midrule

\textbf{Robot collaboration} 
& Emphasizes multi-robot synergy through shared perception and distributed planning, yet rarely integrates underlying communication protocols or edge/cloud offloading mechanisms.
& \textbf{X}
& \textbf{X}
& \textbf{O}
& \textbf{X}
& \textbf{X}
& \textbf{X}
& \cite{wu2024stateoftheartrobotlearningmultirobot,nair2024collaborativeperceptionmultirobotsystems,machines12080540,baratta2023hrc} \\
\midrule

\textbf{V2X}
& Concentrates on vehicle-to-everything links for low-latency road safety and traffic coordination, lacking a broader focus on multi-robot contexts or advanced AI-driven orchestration.
& \textbf{X}
& \textbf{O}
& \textbf{X}
& \textbf{X}
& \textbf{X}
& \textbf{X}
& \cite{Mande2024V2X6G} \\
\midrule

\textbf{M2M}
& A device-centric communication protocol for automated IoT interactions, including R2R, yet generally omits goal-oriented design or large-model–assisted collaboration.
& \textbf{O}
& \textbf{X}
& \textbf{X}
& \textbf{X}
& \textbf{X}
& \textbf{X}
& \cite{Verma201683,Zhao2017M2MChallenges} \\
\midrule

\textbf{URLLC}
& Promises ultra-reliable, sub-ms network links for mission-critical 5G/6G, but seldom integrates multi-robot perception or large-scale AI inference into the communication structure.
& \textbf{X}
& \textbf{O}
& \textbf{X}
& \textbf{X}
& \textbf{X}
& \textbf{X}
& \cite{Haque2024URLLC,Liu2023MLfor6GURLLC,Shim_URLLC} \\

\midrule

\textbf{R2X (Proposed)} 
& Offers a full-stack paradigm coupling LLM-based feature vector encoding, multimodal sensing (incl. emergency audio/visual cues), edge/cloud resource management, and multi-robot collaboration from physical-layer protocols to high-level orchestration, enabling \emph{intent-to-resource} orchestration from language goals to sensing/communication/computation decisions.
& \textbf{O}
& \textbf{O}
& \textbf{O}
& \textbf{O}
& \textbf{O}
& \textbf{O}
& \emph{\textbf{This work.}} \\

\bottomrule
\end{tabular}
}
}
\end{table*}

\tcb{As Table~\ref{tab:table_r2x_comparison} makes explicit, the key differentiator of R2X lies not in any single capability in isolation, but in the \emph{simultaneous integration} of all six dimensions under a unified intent-to-resource orchestration framework. In particular, the combination of task-centric communication requirements design and LLM-based AI synergy---absent in all prior paradigms---is what enables the system to translate high-level language goals into concrete sensing, wireless, and computation decisions.}
The primary goal of this paper is to undertake a comprehensive study of the robot-to-everything (R2X) paradigm-defined as an integrated orchestrator for communication, sensing, and computing tailored to LLM-based collaborative robots \cite{wu2024stateoftheartrobotlearningmultirobot,nair2024collaborativeperceptionmultirobotsystems,machines12080540,baratta2023hrc}-which holistically combines these features into a unified framework.
In Table~\ref{tab:table_r2x_comparison}, we summarize R2X in comparison with conventional frameworks, underscoring how R2X uniquely spans from low-level wireless protocols up to LLM-based semantics for multi-robot decision-making. Robots exchange not only raw sensor data but also contextual representations, enabling on-the-fly adjustments of tasks, routes, and resource allocations.

Specifically, R2X emphasizes {\it{multi-modal and task-aware sensing}}, where sensing information is selectively extracted and compressed using feature compression to minimize bandwidth usage while preserving task-relevant details.
From a communication perspective, R2X adopts {\it{semantic and task-oriented communication}} policies in which bandwidth, latency, and reliability requirements are dynamically aligned with each robot's current priority and the semantic importance of exchanged information, rather than being optimized in isolation.
Furthermore, R2X enables {\it{a dynamic and hierarchical division of computation}} between robots and edge/cloud servers.
Simple reflex actions (e.g., immediately stopping upon obstacle contact, adjusting grip force when an object slips, or pulling back a manipulator arm if it encounters excessive resistance) are processed locally, while heavier tasks (e.g., joint map fusion and route optimization) are offloaded to edge or centralized servers, especially when stringent timing constraints need to be satisfied.

\begin{figure*}[t]
    \centering
  \ifnum\coloption=1
    \includegraphics[width=1\textwidth]{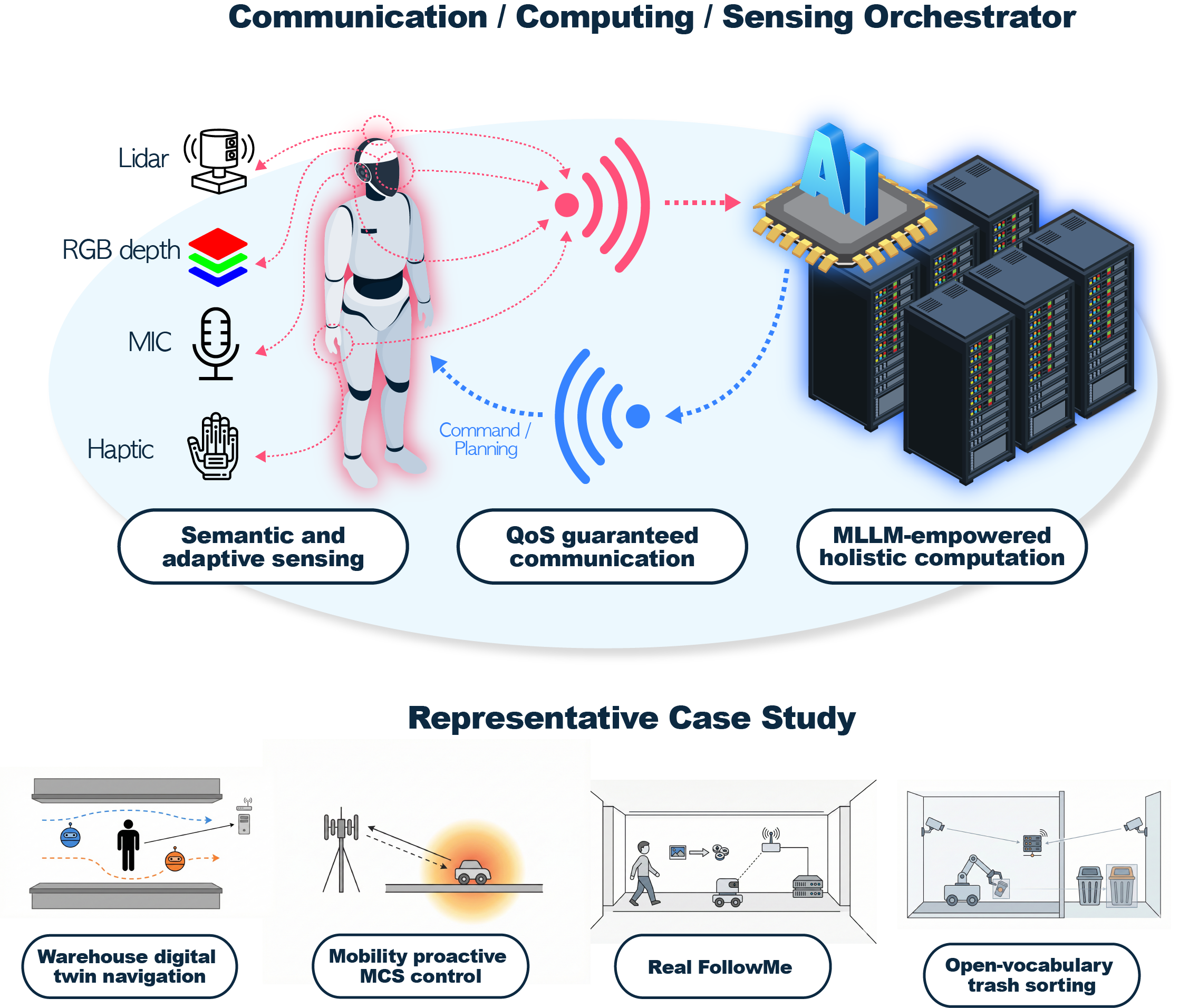}
  \else
    \includegraphics[width=0.7\textwidth]{figures/overview_demos.png}
  \fi
    \caption{High-level overview of this paper's key concept, four end-to-end demonstrations (Demo~I--IV), and overall organization.}
    \label{fig:overview2}
\end{figure*}

As illustrated in Figure \ref{fig:overview2}, this paper explores the unification of communication, sensing, and computing orchestration on multimodal LLM (MLLM)-driven intelligence and multi-robot systems. 
Fig.~\ref{fig:overview2} now summarizes \emph{four} end-to-end demonstrations (Demo~I--IV) that instantiate the same orchestration loop (sense $\rightarrow$ communicate $\rightarrow$ compute $\rightarrow$ act) with measurable system-level metrics.
\tcb{These four demonstrations progressively increase realism from a ray-tracing digital twin, through a simulated mobility scenario, to two real-hardware deployments; full details are provided in Section~\ref{sec:apps_intro}.}

The subsequent sections of the paper discuss the following subjects: Section~II introduces LLM and MLLM technologies for R2X networks.
In Section~III, we examine the fundamental building blocks, including sensing, R2X communications, and the distinction between distributed and centralized computing. 
\tcb{In Section~IV, we present four end-to-end demonstrations (Demo~I--IV) that span digital-twin simulation, mobility simulation, and real hardware, each reporting measurable system-level KPIs.}
Finally, Section~V proposes open research challenges.

\section{Basics of Multimodal LLM-based R2X Networks} \label{sec:mlm_basics}
In R2X, MLLMs are not only perception/reasoning engines but also an \emph{intent interface} that converts high-level language goals (e.g., urgency, safety, priority, and target objects) into system-level decisions~\cite{Kim2026LMMAidedScheduling}.
This intent-conditioned control is the key enabler for \emph{jointly optimizing} sensing, wireless communication, and computation: the same physical task can demand very different sensing modalities, payload representations, and compute placement depending on what the user actually wants to achieve.

\tcb{Throughout this paper, we use \emph{LLM} to denote text-only (unimodal) large language models that process structured or natural-language inputs exclusively, and \emph{MLLM} to denote models that additionally ingest and reason over non-textual modalities (e.g., visual features, depth maps, or multimodal sensor embeddings). The term \emph{LMM} (Large Multimodal Model), sometimes used interchangeably with MLLM in the literature, is avoided hereafter to prevent ambiguity; we standardize on \emph{MLLM} for multimodal models throughout this work. Concretely, the server-side orchestrator in Demo~I (Section~\ref{sec:demo1_warehouse}) is a \emph{text-only LLM} (LLaMA~3~8B) that maps structured system context and operator intent to a schema-valid JSON configuration—no visual or sensor stream is fed directly into the model. In contrast, the orchestrators in Demo~II (Section~\ref{sec:demo2_mcs}) and Demo~IV (Section~\ref{sec:demo4_trashsorting}) consume multimodal semantic features or visual observations alongside text, and are therefore referred to as \emph{MLLM}-based orchestrators.}

\subsection{LLM and MLLM for R2X Networks}

The field of generative AI has seen remarkable growth in recent years with the advancement of LLMs and MLLMs technologies, generating a bewildering variety of applications in both consumer and industrial domains. 
Generative AI, through models such as LLMs and MLLMs, enables machines to interpret, generate, and interact with both textual and multimodal data.
Recently, LLM-assisted robotic systems have leveraged multi-modal prompts and foundation models to integrate the language understanding and decision-making capabilities of LLMs and MLLMs, enabling sophisticated interactions and complex task execution across diverse scenarios. 
For example, Google's Robot Transformer (RT) is designed to comprehend and execute physical tasks using natural language command \cite{RT1, RT2}. Using RT, robot-to-robot and robot-to-human interactions in shared environments can be possible, allowing robots to respond contextually and execute cooperative tasks.

\begin{figure}[t]
    \centering
    \includegraphics[width=1\columnwidth]{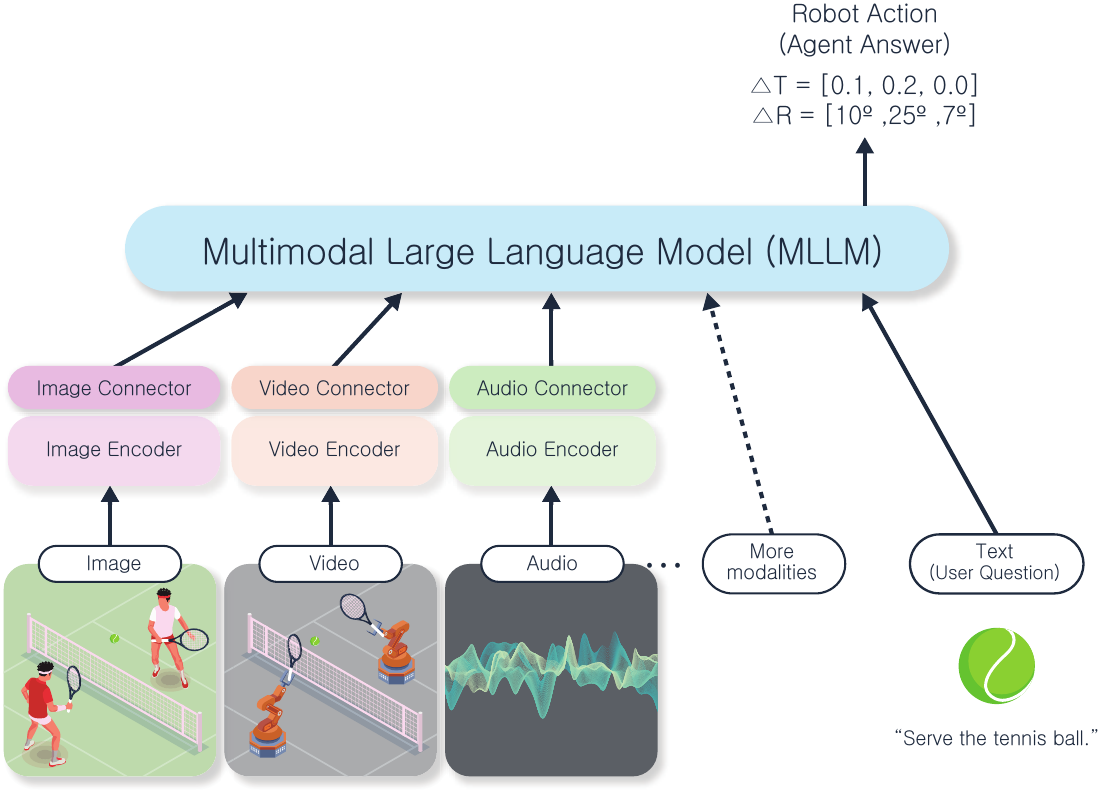} 
    \caption{Common block diagram of MLLMs.}
    \label{fig:lmm_concept} 
\end{figure}

\subsubsection{MLLM Architectures}
MLLM leverages the power of LLMs to process and generate \tcb{multimodal} information~\cite{liu2024visual,liu2024improved}. This model typically consists of three key components: modality encoders, modality connectors, and an LLM core (see Figure \ref{fig:lmm_concept}). The modality encoder extracts modality-specific features from the input sensing image, video, or speech. Popular choices for the visual input data include Vision Transformers (ViT)~\cite{dosovitskiy2021an} and Resnet~\cite{he2016deep}.
The modality connector bridges the gap between visual/speech and textual modalities by translating heterogeneous sensing inputs into a unified semantic representation space.
This alignment process can be achieved through various techniques, including linear projections, cross-attention, Q-Formers, and multi-layer perceptrons (MLP)~\cite{liu2024visual,jaegle2021perceiver,li2023blip}.

As a core component, LLM serves as the central processing unit of the MLLM, receiving the aligned multimodal information and generating responses based on the input~\cite{tong2024cambrian}. Some MLLM also incorporate a generator to produce outputs in different modalities besides text~\cite{team2024chameleon}. 
For example, MLLM can be integrated with Stable Diffusion models~\cite{rombach2022high,peebles2023scalable,podell2024sdxl,kimvisually} to handle image generation and editing capabilities or can be integrated with text-to-speech (TTS) modules~\cite{ren2021fastspeech,popov2021grad} to generate synthetic voice.

\subsubsection{MLLM Training}
Training a fully functional and human-friendly MLLM usually includes a three-stage process~\cite{yin2024survey}: pre-training, instruction tuning, and alignment tuning. Each stage serves a distinct purpose and utilizes different types of data.

\begin{itemize}
\item \textbf{Stage 1: Pre-training}

The primary goal of pre-training is to align different modalities and instill world knowledge in the MLLM.
This stage relies heavily on large-scale, text-paired datasets, such as caption data that describe images, audio, or videos in natural language sentences.
For example, the data used in vision-text pre-training is typically large-scale, publicly available, and less curated, such as Conceptual Captions 3M (CC3M)~\cite{sharma-etal-2018-conceptual} and the LAION family datasets~\cite{schuhmann2022laion}.
These datasets offer a broad spectrum of styles and content, valuable for initial alignment.
Main target of the pre-training is the \tcb{multimodal} connector but the \tcb{multimodal} encoders and MLLM can also be trained optionally.
Similar to the language model training, the training objective is to predict the next token in a sequence, guided by multimodal context.

\item \textbf{Stage 2: Instruction tuning}

Instruction-tuning aims to train models to understand user instructions and complete specified tasks. This is akin to adapting a communication protocol to support higher-level services or user-oriented functionalities. This stage significantly enhances the zero-shot performance of MLLM, allowing the model to generalize to unseen tasks by following new instructions~\cite{bai2024survey}. As in modern AI chatbot systems, instruction tuning assumes a realistic user-agent conversation. For example, in a visual question answering (VQA) task, the instruction might be ``What is the person holding in this image?" and the input would be an image showing a person holding a book. The model generates the response, ``The person is holding a book". \tcb{Multimodal} instruction samples usually contain an optional instruction and an input-output pair. The instruction is a natural language sentence describing the task; for example, the input might be an image-text pair (as in visual question-answering tasks) or just an image (as in image captioning tasks). The output is the model's response to the input instruction. 

\item \textbf{Stage 3: Alignment tuning}

Alignment tuning can be seen as a process of refining the model's ability to generate responses that align well with human preferences and expectations.
This stage may involve techniques like reinforcement learning from human feedback (RLHF)~\cite{kaufmann2024surveyreinforcementlearninghuman}, where human judges evaluate the model's responses and provide feedback to guide further fine-tuning.
This step improves the user-defined quality metrics, such as transmission quality in communication systems and the quality of user experience.

\end{itemize}

\subsubsection{Collaborative MLLMs}
As AI systems become more pervasive, approaches using multiple AI agents working together—much like nodes in a communication network—have gained attraction~\cite{xi2023rise}.
In these collaborative settings, multiple MLLMs (or specialized agents) cooperate to solve tasks, exchange information, and leverage their diverse skill sets~\cite{qian2024scaling,qian2024chatdev}. 
Understanding these interactions is highly relevant to the communication domain, as they resemble distributed communication protocols and multi-agent coordination strategies.
One major challenge of multi-agent collaborative systems is information asymmetry, a problem where each agent can only access the information of its task and nearby environment~\cite{liu2024autonomous}. 
Addressing this challenge requires agents to 1) proactively exchange information, 2) access and retrieve information from collective servers, and 3) make decision under insufficient information.

\begin{itemize}
\item \textbf{Centralized AI agents (ordered co-op)}

Recent works focus on ordered cooperation, a type of collaborative multi-agent system where agents follow specific rules and interact in a structured manner. Analogously, this can be viewed as a centralized system, where a central authority or protocol dictates the flow of information and tasks.
ChatDev~\cite{qian2024chatdev}, MetaGPT~\cite{hong2023metagpt}, and MACNET~\cite{qian2024scaling} use predefined agent-agent interaction protocols that follow a tightly controlled sequence, similar to a well-defined routing protocol in a communication network.
These systems benefit from the organized workflow and efficient communication inherent in a centralized approach. 
While such centralized protocols work well in strictly controlled environments, they face significant challenges when applied to real-world agents like robots or vehicles. 
For instance, in a warehouse scenario where multiple mobile robots navigate unpredictable obstacles or in an autonomous driving fleet scenario where dynamic traffic and sensor noise exist, it would be very difficult to make the coordinated decisions that ensure both efficiency and safety.

\item \textbf{Distributed AI agents (disordered co-op)}

Disordered cooperation often occurs in multi-agent systems where three or more agents communicate openly, without a predefined order or protocol.
Each agent can freely express their perspectives, offer feedback, and suggest modifications related to the task at hand.
This open-ended dialog-based approach is similar in spirit to the dynamic ad-hoc networks or peer-to-peer systems.
While disordered cooperation offers flexibility and adaptability, it also causes difficulty in consolidating feedback and potential inefficiency caused by redundant communication.
Studies like iAgents~\cite{liu2024autonomous} and RoCo~\cite{mandi2024roco} tackle these issues by exchanging the necessary information and responding to uncertainties and asymmetries in their local views of the environment.

\end{itemize}

\subsubsection{\tcb{Multimodal} Prompting and Robot Foundation Models for LLM/MLLM}
LLM-based robot control is rapidly advancing, with a focus on incorporating \tcb{multimodal} prompts and foundation models \cite{Celemin22, Wang23, Vasan24}.
Recent research demonstrates how \tcb{multimodal} prompts, integrating text and visual information, are used to control robots.
For example, this approach has proven effective in imitation learning \cite{Celemin22}, following language instructions \cite{Wang23}, and achieving visual goals \cite{Vasan24}.

In addition, robot foundation models, trained on massive datasets, have immense potential for creating more versatile and capable robots~\cite{zhao2023learning,team2024octo,yang2025octopus}.
These models can be applied to various robot operations, including perception, planning, and control~\cite{kawaharazuka2024real}. 
For example, by employing these models, automated manufacturing plants can perform dynamic assembly line tasks, such as precise component placement and real-time quality control.
In healthcare, these models can be used for robotic surgical assistants that adapt to live imaging data and surgeon commands.

Recently, vision-language-action (VLA) models integrating holistic visual understanding with large-scale pre-training have received much attention since VLA models improve the capability of robots to interpret more challenging environments and support difficult tasks~\cite{Kim2026AdaptiveCapacityAllocation}.
For instance, models like LLARVA~\cite{niu2024llarva} combine pre-trained visual processing modules with VLA-specific training techniques, such as fine-tuning visual encoders and employing visual trace prediction, aiming at improving the mapping between visual observations and task-aware actions.
On the other hand, models like $\pi_0$~\cite{black2024pi0} and its efficient adaptation ($\pi_0$-FAST~\cite{pertsch2025fast}) leverage the vast amount of robot data~\cite{RTX} and introduce advanced tokenization methods (e.g., frequency-based actions) to achieve high-frequency control and superior generalization in complex tasks.
In addition, LAPA~\cite{ye2024lapa} showcases a growing trend toward using unlabeled video data to pre-train latent action representations.

Also, exemplary LLM/MLLM-based robotic systems have been proposed recently.
Palm-E~\cite{Palm-e} is a pre-trained large language model, designed for multiple embodied tasks including sequential robotic manipulation planning and visual question answering.
SayCan~\cite{ahn2022can} is a system that leverages an LLM to ground language commands in robotic affordances.
The system evaluates the feasibility of actions based on language input and visual perception, allowing robots to execute tasks based on natural language instructions.
Code-as-Policies~\cite{liang2023code} employs LLMs to generate Python code that serves as a control policy for robots, demonstrating the potential for programmatically controlling robot behavior using human language.
Moreover, CLIPort \cite{shridhar2022cliport} uses a VLM (Vision-Language Model) for feature extraction in low-level perception to guide pick-and-place actions, and  ProgPrompt~\cite{singh2023progprompt} and REFLECT~\cite{pmlr-v229-liu23g} utilize VLMs and LLMs for scene recognition, failure explanation, and corrective task planning.
Recent work such as MedRescue-RT demonstrates MLLM-guided humanoids that detect a collapsed patient, request an AED, and verbally guide by standers. Table \ref{tab:llm-robotics-combined} summarizes the related studies about LLM/MLLM systems for robotics.

\begin{table*}[tbh!]
\caption{LLM/MLLM for robotics.}
\centering
\scriptsize
\renewcommand{\arraystretch}{0.4}
\setstretch{1.1}
\resizebox{\textwidth}{!}{
\begin{tabular}{c c c p{11.5cm}}
\toprule
\textbf{Topic} & \textbf{Method} & \textbf{Ref.} & \textbf{Key Contribution} \\
\midrule
\multirow{4}{*}{\rotatebox[origin=c]{90}{\hspace{-3.0cm}Multi-modal Sensing}}
& MUTEX & \cite{shah2023mutex} &
 LMM equipped with a Transformer-based cross-modal fusion module capable of processing six modalities. Introduces a two-stage training procedure and a multi-modal training dataset collected from both simulated and real-world environments.
\\ \cmidrule(lr){2-4}
& Matcha & \cite{zhao2023matcha} &
Robotic planning system that actively collects information from multi-modal sensors (vision, sound, haptics) by performing epistemic actions. The LMM reasons over the information obtained from such actions and plans the main task execution.
\\ \cmidrule(lr){2-4}
& REFLECT & \cite{pmlr-v229-liu23g} &
Robot failure explanation system based on the summarization of the robot experiences, represented as hierarchically structured multi-sensory data. The summarization is used to develop a natural language behavior explanation system and to improve the model's correction ability.
\\ \cmidrule(lr){2-4}
& MultiPLY & \cite{hong2024multiply} &
Embodied LMM that utilizes encoded sensor representations obtained from diverse interactive 3D environments. Sensors include visual, audio, tactile, and thermal information. A large-scale multi-sensor dataset, \textit{Multisensory Universe}, is also introduced.
\\
\midrule
\multirow{5}{*}{\rotatebox[origin=c]{90}{\hspace{-3.5cm}Language-based Task \& Env. Underst.}}
& PaLM-E & \cite{Palm-e} &
Embodied LMM that takes text-based task instructions for various robot tasks, including perception, visually-grounded question-answering, and planning. PaLM-E leverages a pre-trained LLM as the initial state and adapts the model using robot-aware training data.
\\ \cmidrule(lr){2-4}
& CLIPORT & \cite{shridhar2022cliport} &
CLIP-based language conditioning model for vision-based robot control. The model utilizes semantic (RGB) and spatial (RGB-D) visual data using separate submodules, both conditioned on the CLIP text embedding.
\\ \cmidrule(lr){2-4}
& Inner Monologue & \cite{huang2023inner} &
Providing text-based feedback to enhance LLM-based robot planning. Feedback types include simple success detection, passive scene description, and human-provided scene descriptions given only when requested by the LMM.
\\ \cmidrule(lr){2-4}
& SayCan & \cite{ahn2022can} &
Grounding LLM responses to the actual environment by providing value functions or pre-trained skills (tools). As LLMs cannot assess the correctness of their answers, external signals help them understand the robot's capabilities and environment.
\\ \cmidrule(lr){2-4}
& ScaleUp-DistilDown & \cite{ha2023scaling} &
Scaling up the language-labeled robot planning by decomposing tasks into hierarchical plans, and distilling the generated large-scale data into a multi-task, language-conditioned robot policy model.
\\
\midrule
\multirow{3}{*}{\rotatebox[origin=c]{90}{\hspace{-2.3cm}Direct Action Generation}}
& RT Family & \cite{RT1,RT2,RTX} &
Unified vision-language-action model that generates robot actions as output “sentences” from the LMM. The model leverages the successful LLM architecture and is trained with Internet-scale vision-language task datasets. This simple approach enables the use of a highly optimized LLM training pipeline.
\\ \cmidrule(lr){2-4}
& OpenVLA & \cite{kim2024openvla} &
Open-source vision-language-action model that demonstrates strong generalization and language grounding abilities in various robot manipulation tasks. Specifically, OpenVLA employs Llama-2-7B as LLM core, and DinoV2 and SigLIP as vision encoders.
\\ \cmidrule(lr){2-4}
& LLARVA & \cite{niu2024llarva} &
Advanced instruction tuning method designed for various robotic learning tasks and environments. In particular, providing 2D coordinate traces of an end-effector (e.g., robot hand) jointly with RGB images improves robot policy alignment.
\\
\midrule
\multirow{3}{*}{\rotatebox[origin=c]{90}{\hspace{-2.2cm}Policy Generation}}
& Code-as-Policies & \cite{liang2023code} &
Repurposing code-writing LLMs to generate robot policy codes based on natural language commands. Given a textual prompt and few-shot examples, LLM can write robot policies that generalize to new instructions and decide precise action values.
\\ \cmidrule(lr){2-4}
& ProgPrompt & \cite{singh2023progprompt} &
Writing robot plans in Python programs by leveraging LLMs to complete the code template. For example, step-by-step planning is represented as comments in Python code. Environment and action types are provided in the LLM prompt.
\\ \cmidrule(lr){2-4}
& Eureka & \cite{ma2024eureka} &
Human-level reward design algorithm powered by code-writing LLMs, such as GPT-4, to generate reward functions for learning complex robotic skills through evolutionary exploration and reward reflection. This reduces the need for manually designed rewards for each task.
\\
\midrule
\multirow{3}{*}{\rotatebox[origin=c]{90}{\hspace{-2.0cm}Planning}}
& DoReMi & \cite{guo2023doremi} &
Utilizing LMM not only for planning but also for failure detection. When vision information indicates a constraint violation, LMM immediately aborts the previous plan and performs re-planning to achieve the goal.
\\ \cmidrule(lr){2-4}
& LLM-Planner & \cite{song2023llmplanner} &
Few-shot robot planning based on off-the-shelf LLM (e.g., GPT). Few-shot examples are adaptively selected from the prebuilt database using the K-nearest neighbor retriever that measures the similarity between training examples and the target query.
\\ \cmidrule(lr){2-4}
& AutoTAMP & \cite{chen2024autotamp} &
Translate-and-Check approach, where LLM generates a set of timed waypoints to visit. LLM first translates the goal into trajectories (i.e., state sequence), and an external checker verifies the validity of the plan. This iterative process ensures the successful plan execution.
\\
\bottomrule
\end{tabular}
}
\label{tab:llm-robotics-combined}
\end{table*}

\subsubsection{Sensing Feature Extraction for LLM/MLLM-based Robots}
Sensing feature extraction is a critical process in LLM/MLLM-based robotic systems to perceive and understand their environment through various sensors such as cameras, depth sensors, and tactile sensors. 
The goal of feature extraction is to transform raw multimodal sensing data into a compact, meaningful representation that captures essential information while reducing redundancy.
In the low-level perception stage, raw sensing data, such as images or point clouds, are processed to extract relevant features for tasks.
This process involves several steps:

\begin{itemize}
    \item \textbf{Preprocessing:} raw data from sensors often contains noise or irrelevant information, so preprocessing techniques like normalization, filtering, and denoising are applied to clean the data and prepare it for feature extraction.
    
    \item \textbf{Feature Extraction:} this step is crucial in processing sensor data, where algorithms are used to identify the key attributes such as edges, corners, textures, or extracting points of interest from images and point clouds. Deep learning-based methods, such as CNN or ViT, are popularly used for their ability to automatically learn complex patterns from large datasets.

    \item \textbf{Compression and Transmission:} to transmit the feature vector to the central processing unit containing LLM/MLLM, compression of raw sensing data is indispensable. Compression reduces the precision of the feature vector elements, resulting in a more compact representation that facilitates faster and more efficient transmission over wireless channels. This is particularly crucial in robotic communication systems with limited bandwidth and stringent delay requirements. 
    By applying advanced quantization techniques, such as vector quantization (VQ), or deep learning-based quantization for the foreground object, we can ensure effective communication with minimal loss of sensing data. For example, VQ technique compresses the high-dimensional feature vectors by mapping them into a finite set of representative codewords. Let  $\mathbf{x} \in \mathbb{R}^d$ be the feature vector to be transmitted, and $\mathcal{C} = \{\mathbf{c}_1, \mathbf{c}_2, \ldots, \mathbf{c}_K\} \subset \mathbb{R}^d$ be the codebook comprising $K$ codewords. The VQ encoder maps $\mathbf{x}$ to the nearest codeword in $\mathcal{C}$ based on the squared Euclidean distance: $Q(\mathbf{x}) = \arg\min_{\mathbf{c}_k \in \mathcal{C}} \|\mathbf{x} - \mathbf{c}_k\|^2$.
This mapping does \emph{not} reduce the vector dimensionality (the reconstruction still lies in $\mathbb{R}^d$); but, it reduces the \emph{communication payload/bitrate} by transmitting only the \emph{codeword index}. For a codebook of size $K$, the index requires $\lceil \log_2 K \rceil$ bits per symbol (before any entropy coding or protocol overhead).
The decoder reconstructs the quantized vector as $\hat{\mathbf{x}} = \mathbf{c}_{Q(\mathbf{x})}$, yielding a deterministic, bounded-complexity reconstruction rule.
\end{itemize}

\subsection{Robot Operating Systems for R2X Networks}

A robot operating system (ROS) is a flexible framework designed to build robotic applications with minimal redundancy \cite{ROS1, ROS4}. 
Basically, it is a meta-operating system environment where developers can integrate multiple components (e.g., sensor, control, and planning components) into a cohesive system. 
Each node in ROS is responsible for executing a specific task, and nodes communicate with each other by sending and receiving messages.

ROS uses a publish/subscribe communication model, enabling the exchange of data between nodes. 
Nodes can either publish data to a topic or subscribe to a topic to receive data. 
Input data typically comes from sensors, which are handled by subscriber nodes that process the incoming information, while output data is sent to actuators through publisher nodes that control robotic actions.
This communication model is particularly effective in distributed robotic systems, as it supports asynchronous message passing, allowing for flexible, real-time data exchange across different system components.

The strengths of ROS lie in its robustness for hardware interfacing and real-world deployments.
Its modular design supports system-level innovation by allowing developers to work with reusable and scalable components. 
While ROS is easy to use, it cannot handle high-level tasks, such as natural language understanding and advanced perception.
This limitation can be addressed by employing a VLA model, which significantly enhances the robot's ability to understand and execute complex tasks that combine visual and linguistic inputs.
While ROS manages low-level communication and device control, VLA models provide high-level reasoning, making them complementary.

The output of the VLA model  $[\Delta \mathbf{x}, \Delta \boldsymbol{\theta}, \Delta \text{grip}]$ consists of key components that direct the robot's actions. $\Delta \mathbf{x}$ represents the change in the robot's position, specifying the distance the robot needs to move in a particular direction. 
$\Delta \boldsymbol{\theta}$ denotes the change in the robot's orientation, or the angular adjustment required for the robot to align itself correctly with a target or goal. 
$\Delta grip$ indicates the necessary change in the robot's grasping force, dictating how tightly it should grasp an object.

In latency-sensitive applications such as Multi-Arms Assembly (in representative examples in Fig. 1.), VLA models can be used without ROS.
For example, camera images of robots are transmitted via manufacturer SDKs directly to the VLA model, which predicts actions that are converted into robot commands (e.g., URScript, RAPID, libfranka) and sent straight to controllers. This direct approach simplifies the control pipeline and minimizes delays.
A typical example is bi-manual manipulation using models such as ACT or Pi0 \cite{black2024pi0}. 
Wrist- and waist-mounted cameras (640 × 480 RGB) stream images directly to the VLA, which must run on a cloud-hosted A100 GPU because its inference time and memory footprint exceed the limits of on-board hardware. 
Each 10 Hz control cycle therefore includes (i) network round-trip to the server, (ii) the ~70–80 ms forward pass of the VLA, and (iii) local image‐transfer overhead. With ROS publish/subscribe this overhead is typically 6–18 ms per cycle \cite{dugas2016ros,maruyama2016ros,ronnau2024ros2}; using the camera vendors' SDKs it falls to 1.5–3 ms \cite{intelrealsense2023performance}. 
Eliminating the extra 5–15 ms is essential because that time must be re-allocated to cloud communication latency and still leave the system within the 100 ms budget required for control rates above 10 Hz. 
For this reason, the recommended architecture bypasses ROS, relies on SDK streaming, and reserves the saved margin for the unavoidable server-side computation and network delays. 
Table \ref{tab:ROS_vs_RT} summarizes the complementary roles, strengths, and limitations of ROS and VLA-based control.

\begin{table*}[tbh!]
\caption{Comparison of ROS and VLA.}
\centering
\small
\renewcommand{\arraystretch}{1.2}
\begin{tabular}{p{1.9cm}p{5.9cm}p{5.9cm}}
\hline
 & \textbf{\hspace{0.1cm} Robot Operating System (ROS)} & \textbf{\hspace{0.1cm} Vision Language Action Model (VLA)} \\ \hline
\textbf{Purpose} & \begin{itemize}
    \item Middleware framework for robotic system development.
    \item System-level integration of hardware and software for robots.\vspace{-0.15cm} 
    \end{itemize} & \begin{itemize}
    \item AI-driven reasoning and perception.
    \item High-level reasoning, task abstraction, and perception-to-action mappings. \vspace{-0.35cm}
\end{itemize} \\ \hline
\textbf{Example} & \begin{itemize}
    \item ROS1\cite{ROS1}, ROS2\cite{ROS3} \vspace{-0.35cm}
\end{itemize} & \begin{itemize}
    \item SayCan\cite{ahn2022can}, PaLM-E\cite{Palm-e}, RT-1\cite{RT1}, RT-2\cite{RT2}, OpenVLA\cite{kim2024openvla}, LLARVA\cite{niu2024llarva}, $\pi_0$\cite{black2024pi0} \vspace{-0.15cm}
\end{itemize} \\ \hline
\textbf{Communication \& Computing Aspect} & \begin{itemize}
    \item Publish-subscribe model, standardized messaging, widely used.
    \item Distributed processes, moderate hardware requirements. \vspace{-0.35cm} \end{itemize} 
    & \begin{itemize}
    \item High-bandwidth data handling for large AI models, integrates advanced inference frameworks.
    \item Transformer-based deep learning, requires substantial computational resources. \vspace{-0.15cm}
\end{itemize} \\ \hline
\textbf{Strength} & \begin{itemize}
    \item Robustness in hardware interfacing and real-world deployment.
    \item Modular design and large ecosystem of tools/libraries. \vspace{-0.35cm}
\end{itemize} & \begin{itemize}
    \item Superior performance in high-level tasks and generalization across different task domains.
    \item Leverages state-of-the-art AI advancements. \vspace{-0.15cm}
\end{itemize} \\ \hline
\textbf{Limitation} & \begin{itemize}
    \item Communication overhead in large-scale robot systems.
    \item Limited AI/ML capabilities. \vspace{-0.15cm}
\end{itemize} & \begin{itemize}
    \item Extremely high computational cost.
    \item Dependency on data and compute infrastructure for training and inference. \vspace{-0.15cm}
\end{itemize} \\ \hline
\textbf{Application} & \begin{itemize}
    \item Autonomous navigation, robotic arm control, sensor fusion, multi-robot systems. \vspace{-0.35cm}
\end{itemize} & \begin{itemize}
    \item Task planning using natural language, multi-modal reasoning, human-robot interaction. \vspace{-0.15cm}
\end{itemize} \\ \hline
\end{tabular}
\label{tab:ROS_vs_RT}
\end{table*}

\subsection{Task-Oriented Collaboration for R2X Networks}

In multi-robot systems operating in complex physical environments, tight cooperation among sensing, communication, and computation with MLLM is essential for efficient and reliable task execution. Individual robots are inherently limited in sensing coverage and onboard computational capability, which makes it difficult to achieve comprehensive environmental understanding or high-level decision-making in isolation using small-sized LLM. By sharing sensed information, coordinating actions through communication, and leveraging distributed or centralized computation resources, robots can jointly perform MLLM-based perception, reasoning, and control at the system level. Such cooperation goes beyond raw data exchange and forms an integrated framework that enables collective intelligence.
Moreover, the manner in which sensing, communication, and computation are coordinated is strongly task-dependent. Different tasks impose diverse requirements on latency, reliability, semantic abstraction, and resource utilization. For example, tasks requiring fast reactions may prioritize local processing with minimal communication and small-sized LLM, whereas tasks involving complex reasoning or long-term planning may benefit from edge-assisted or centralized computation with large-sized MLLM. This motivates the concept of task-oriented collaboration, where sensing strategies, communication mechanisms, and computation architectures are jointly designed and dynamically adapted according to task objectives. In this paradigm, robots determine what information to sense, how to represent and transmit it, and where to process it, all from a task-centric perspective.

A representative example of task-oriented collaboration might be the smart farm application, where multiple robots cooperate to perform large-scale agricultural operations (see Fig. \ref{fig:overview2}). Such systems typically include heterogeneous robots, such as ground vehicles for soil sampling, autonomous tractors for planting and harvesting, and aerial drones for field monitoring. These robots employ diverse sensing modalities, including cameras, LiDAR, acoustic sensors, and radar-based soil moisture measurements, to collect complementary information. Lightweight, locally distilled models running on individual robots enable rapid analysis of immediate sensor inputs, while MLLM-based computation at an edge server performs higher-level semantic reasoning by fusing multimodal data and natural-language task descriptions. On top of this global understanding, the edge server can issue coordinated instructions, such as scheduling harvesting times or adjusting irrigation levels, to optimize overall farm productivity.

Communication in this setting is inherently task-oriented, emphasizing the exchange of semantically meaningful and context-aware information rather than raw sensor data. Instead of flooding the network with unprocessed measurements, robots and the edge server transmit concise task-relevant cues, which reduces bandwidth consumption, lowers latency, and improves reliability. This enables near-real-time adaptation to changing environmental conditions and ensures that MLLM-enabled perception, language-driven reasoning, and coordinated actuation operate in a tightly coupled manner. Recent studies have explored related aspects of robot collaboration, including dialogue-based multi-robot coordination and LLM-informed planning \cite{mandi2024roco}, as well as GNN-based collaborative perception frameworks \cite{zhou2022multi}. While these works highlight the benefits of cooperation, a unified task-oriented sensing–communication–computation co-design perspective remains an open and promising research direction.

In addition, the language-level task intent naturally defines the objective and constraints of orchestration (e.g., maximize task success under latency/energy budgets), making intent-aware co-design a core requirement rather than an optional feature.

\section{Integrated Sensing, Communication, and Computation Systems for R2X Networks} \label{sec:core_components}

In this section, we explore the fundamental building blocks of robotic networks, which include sensing, communications, and computing, and how these components are integrated to perform the given task. We also discuss the advantages of centralized architectures for effective multi-robot coordination under MLLM guidance. 

\subsection{Sensing for R2X Networks}\label{sec:multimodal}

\begin{table*}[h!]
\caption{Comparison of multimodal sensors.}
    \centering
    
    \resizebox{0.85\textwidth}{!}{
    \begin{tabular}{p{3cm}p{4cm}p{4cm}p{4cm}}

        \hline
        \textbf{} & \textbf{Principle} & \textbf{Advantages} & \textbf{Disadvantages} \\ 
        \hline
        \textbf{Camera} & Recognizes surrounding objects using photosensitive sensor & Cost-effectiveness, measurements of texture, color, and contrast & Sensitive to weather conditions, short detection range \\ 
        \hline
        \textbf{Radar} & Measures distance and speed using electromagnetic wave & Long-distance object detection, weather resistance, and material penetration detection & Difficult to identify the shape of objects, and low resolution \\ 
        \hline
        \textbf{LiDAR} & Measures distance, shape, and speed using laser & High resolution detection compared to radar & High cost, cannot penetrate objects, shorter detection range compared to radar \\ 
        \hline

        \textbf{IMU} & Measures motion and orientation using accelerometers and gyroscopes & Compact, lightweight, and capable of high-frequency measurements & Accumulates drift over time, requiring correction from other sensors \\ 
        \hline
        \textbf{Microphone} & Converts sound waves into electrical signals & Enables voice recognition, sound localization, and interaction with humans & Sensitive to background noise, limited to auditory applications \\ 
        \hline
        \textbf{Ultrasonic sensor} & Measures distance using high-frequency sound waves & Extremely low cost & Very limited range \\ 
        \hline
        \textbf{Tactile sensor} & Detects physical interactions through pressure, force, and texture sensing & Provides fine control for manipulation tasks, enabling robots to handle delicate objects & Limited to surface interactions, can be sensitive to wear and tear \\ 
        \hline
    \end{tabular}} \label{tbl:multimodal_sensor}
\end{table*}

\begin{table*}[t]
\caption{Sensor characteristics and applications in robotics.}
\centering
\resizebox{\textwidth}{!}{
\begin{tabular}
{p{1.8cm}p{1.8cm}p{1.8cm}p{1.8cm}p{1.8cm}p{6cm}}
\hline
\textbf{Sensor \newline type} & \textbf{Sensing \newline frequency} & \textbf{Resolution} & \textbf{Sensitivity} & \textbf{Data volume} & \textbf{Applications by robot type} \\ \hline

\textbf{Camera} & 30–60 fps & High \newline (e.g., 4K) & High & Very large & 
Mobile robots: visual SLAM \cite{thrun2005multi,garulli2005mobile} \newline
Autonomous vehicles: lane detection \cite{assidiq2008real}, object recognition \cite{du2018unmanned} \newline
Humanoid robots: Human interaction / medical triage / security alert\\ \hline

\textbf{Radar} & 20–100 Hz & Medium & Moderate & Moderate & 
Autonomous vehicles: Obstacle detection in adverse weather \cite{cheng2022novel, coluccia2020detection} \newline
Industrial robots: Machinery monitoring \cite{stetco2020radar}\\ \hline

\textbf{LiDAR} & 10 Hz & High \newline (mm-level) & High & Large & 
Autonomous vehicles: navigation and obstacle avoidance \cite{ hutabarat2019lidar}\newline
Agriculture robots: Plant detection and mapping \cite{weiss2011plant,malavazi2018lidar}\\ \hline

\textbf{IMU} & 50-1000 Hz & Medium & Low & Low & 
Drones: Flight stabilization \cite{yang2017multi}\newline
Humanoid: Balance and motion tracking \cite{semwal2022pattern}\newline
Autonomous vehicles: Dead reckoning \cite{brossard2020ai}\\ \hline

\textbf{Microphone} & 16–48 kHz & Medium & Moderate & Moderate & 
Service robots: Voice interaction \cite{norberto2005robot}\newline
Surveillance robots: Sound localization  \cite{wu2009surveillance}\\ \hline

\textbf{Ultrasonic Sensor} & 10–100 Hz & Low to \newline medium & Moderate & Low & 
Mobile robots: Obstacle avoidance \newline
Drones: Altitude sensing \newline
Warehouse robots: Proximity detection \\ \hline

\textbf{Tactile \newline Sensor} & 100–1000 Hz & High & High & Moderate & 
Robotic arm: precision grasping \cite{van2020large}\newline
Medical robots: surgical instrument handling \cite{jin2021development}\\ \hline

\end{tabular}
}
\label{tab:sensor_comparison}
\end{table*}

In recent years, sensing technologies for robotic systems have significantly improved in terms of resolution, sensitivity, and miniaturization \cite{alatise2020review,kam1997sensor,Kim2024RoleOfSensingCV6G}. Conventional sensing technologies often classify sensors by modality such as camera, LiDAR, IMUs, microphones, ultrasonic sensors, and tactile sensors. However, in R2X networks, sensing should be viewed not as isolated hardware modules but as cooperative, multimodal perception resources that jointly support robust understanding of dynamic environments. The core goal is to construct a comprehensive perception pipeline by integrating complementary modalities, allowing robots to share and utilize sensory information beyond their on-board view.

Table \ref{tbl:multimodal_sensor} describes comparison of various multimodal sensors. Vision sensors, including cameras, serve as the ``eye" of robots, capturing high-resolution imagery for object recognition, 3D perception, navigation, and SLAM (Simultaneous Localization-And-Mapping) \cite{biswas2012depth, Sumikura19}. Their versatility and low cost make them widely adopted. Meanwhile, radar measures distance, velocity, and shape using radio reflections, typically in the 24 or 77-81 GHz bands. Unlike cameras that rely on light, radar is resilient to rain, fog, and low-illumination environments, making it suitable for safety-critical autonomous operation \cite{cheng2022novel}. LiDAR, using laser beams to generate dense 3D maps, offers real-time geometric depth information and stable performance under extreme lighting conditions \cite{xuexi2019slam}, a key requirement for obstacle avoidance and precision navigation.

Complementary inertial and modalities also contribute to the multimodal stack. IMUs (Inertial Measurement Unit), consisting of accelerometers and gyroscopes, track linear acceleration and angular velocity, enabling fast state estimation for balance control and pose prediction \cite{semwal2022pattern}. Microphones serve as auditory sensors to capture spoken commands, environmental sounds, and non-verbal cues. With directional arrays and noise suppression, robots can localize sound sources and interact with humans via voice, forming a natural interface for MLLM-driven communication. In addition, ultrasonic sensors use echo timing (typically around 40 kHz) to provide short-range distance estimation. Although lower-resolution than LiDAR or cameras, they offer lightweight proximity awareness for underwater robots or confined navigation. Tactile sensors \cite{yuan2017gelsight}, mimicking human touch, detect pressure, force, and texture, enabling precise manipulation of fragile or irregular objects.

In R2X networks, these sensors are not merely independent components, they form a multimodal sensing fabric where information is fused spatially and temporally. A camera may capture visual context while LiDAR contributes spatial geometry; radar ensures reliability in adverse weather; IMU stabilizes motion estimation; microphones enable interactive command exchange; tactile sensors validate grasp stability. Depending on the task, robots selectively activate or combine sensing modalities, prioritizing radar-LiDAR in foggy outdoor tasks, camera-microphone for human guidance, tactile feedback for assembly operations. Moreover, for the purpose of robots can share perception data with nearby agents or offload sensory streams to edge servers for global scene understanding and multi-robot exploration.

Therefore, in R2X networks, sensing evolves a local capability into a distributed, task-adaptive perception resource. Rather than transmitting raw signals, robots may exchange compressed semantic features, share depth maps, or collaboratively build unified world models. This multimodal sensing approach provides a foundation upon which task-oriented communication and MLLM-based computation can operate efficiently at scale.

\subsection{Communications for R2X Networks}

\subsubsection{Taxonomy of R2X Communications}

\begin{figure}[tbh!]
\centering
\includegraphics[width=1\columnwidth]{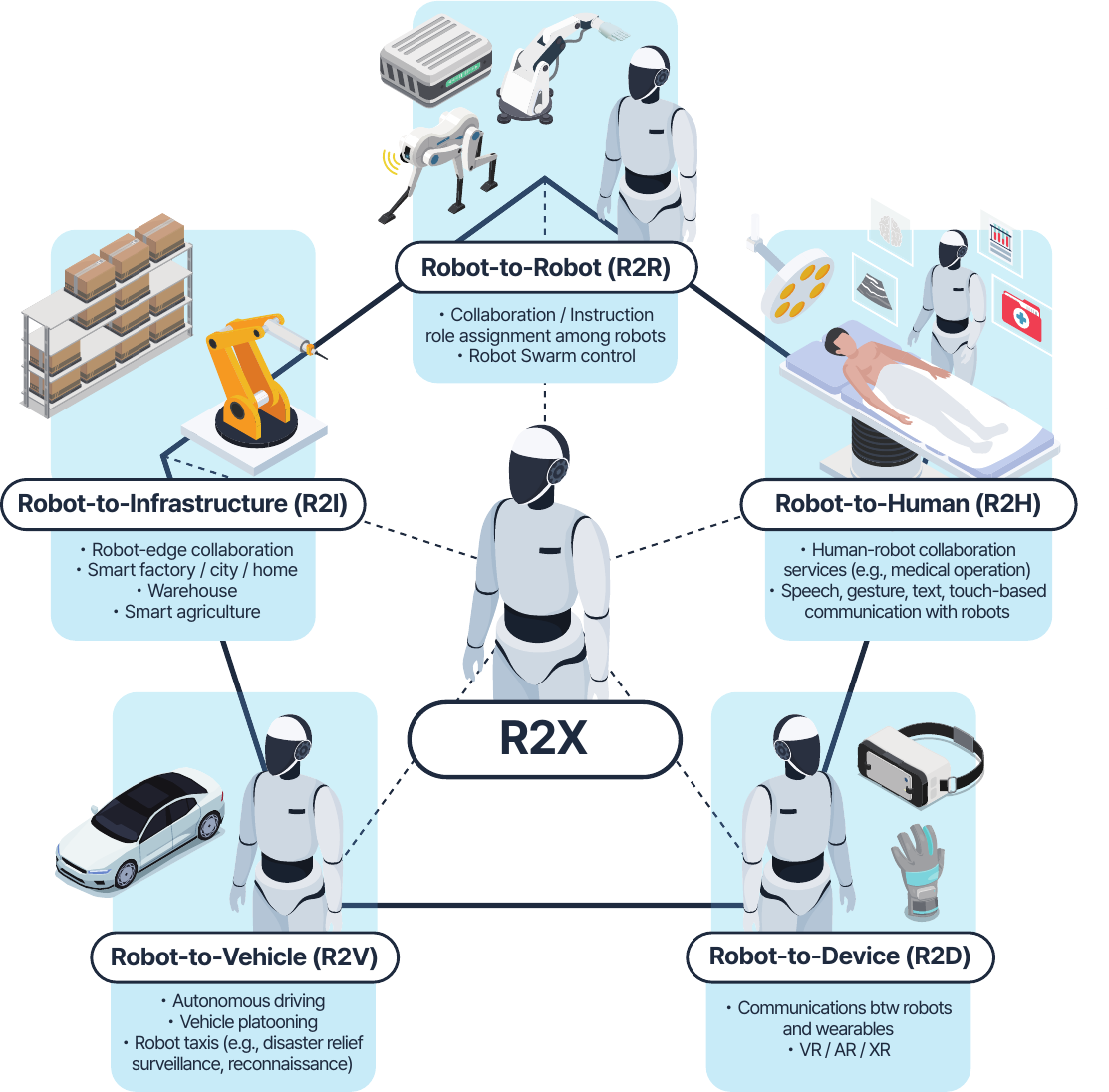}
\caption{A taxonomy of R2X communications.}
\label{fig_R2X_concept}
\end{figure}

The R2X communication paradigm envisions seamless interaction between robots and any entity within their operational environment. R2X encompasses not only direct robot-to-robot (R2R) communication but also interaction with humans (R2H), infrastructure (R2I), vehicles (R2V), and diverse devices (R2D), as illustrated in Figure \ref{fig_R2X_concept}.
R2X facilitates collaborative action, autonomous operation, and intelligent decision-making by connecting robots to a richer communication-based information ecosystem.

\begin{itemize}
    \item \textbf{Robot-to-robot (R2R) Communications}
    
R2R communications focus on the exchange of information, data, or commands among robots, which enables robots to collaborate, share sensor data, coordinate tasks, and make collective decisions without human intervention. The primary communication protocols for R2R communication include specialized mesh networks providing ultra-low latency, high reliability, and ultra-fast data rates. R2R communication is essential in multi-robot systems, swarm robotics, and collaborative robotic tasks where cooperation and synchronized operations are required. 

\item \textbf{Robot-to-human (R2H) Communications}

Safe and efficient human-robot interaction—spanning courteous assistance and rapid rescue/first-aid collaboration—is of paramount importance in shared work-spaces where robots interact closely with humans.
These environments include smart homes, smart factories, agricultural, healthcare, social, and military contexts~\cite{R2H01, R2H02, R2H03, R2H04}. In a military example, manned-unmanned teaming (MUM-T) refers to the collaboration between manned systems (e.g., piloted aircraft or vehicles) and unmanned systems (e.g., drones or autonomous robots)~\cite{MUMT01, MUMT02, MUMT03, MUMT04, MUMT05}. Common modalities for R2H communication include voice commands, gesture recognition, touch interfaces, and even AR interactions.

\item \textbf{Robot-to-infrastructure (R2I) Communications}

Major purpose of R2I communications is to allow robots to interact with and leverage infrastructure such as sensors, servers, networks, and smart environments~\cite{R2I01} to improve their performance, situational awareness, and decision-making. Examples include robots receiving navigation data from GPS systems, interacting with IoT devices in a smart city, or accessing cloud-based servers for computational tasks~\cite{R2I02}. Moreover, mobile edge computing can further improve R2I by supporting ultra-high speed and ultra-low latency connections for the latency-sensitive applications~\cite{TSN_R2I} like autonomous vehicles and drones.

\item \textbf{Robot-to-vehicle (R2V) Communications}

In R2V communications, robots interact with vehicles for coordination, control, and information sharing. Popular examples include robots sending real-time data to autonomous vehicles for obstacle avoidance, vehicles transmitting their status to robots in logistics settings, and robots interacting with connected vehicles in smart transportation systems~\cite{R2V01, R2V02, R2V03}.

\item \textbf{Robot-to-device (R2D) Communications}

R2D communication enables robots to interact with a wide range of devices such as actuators, smartphones, or IoT-enabled equipment~\cite{R2D01}. In healthcare, for example, robots can use wearable data, such as gait analysis or heart rate monitoring, to assist in rehabilitation processes. In manufacturing, AR devices can guide workers in assembly tasks, while robots collaborate with employees wearing safety and motion sensors.
\end{itemize}

\subsubsection{Comparison among V2X, M2M, and R2X}

V2X communications is a paradigm to facilitate data exchange between vehicles and their surroundings \cite{Garcia21}. 
V2X can be divided into multiple forms: infrastructure (V2I), other vehicles (V2V), pedestrians (V2P), and the cloud (V2N).
For example, cellular V2X (C-V2X)~\cite{CV2X}, standardized by 3GPP, supports these modes through two communication interfaces: the Uu interface for V2I and V2N via BSs, and the PC5 sidelink interface for direct V2V and V2P communications without network coverage.
Machine-to-machine (M2M) communication~\cite{Kim13} is a paradigm in the Internet of Things (IoT) to support automated data exchange between interconnected devices without human intervention.
For example, NB-IoT~\cite{NBIoT} and LTE-M~\cite{LTEM} are optimized for static low-power sensors (e.g., smart metering) and moderate mobility devices (e.g., asset tracking and wearables), respectively.

\begin{itemize}
    \item \textbf{Mobility and Environmental Adaptability} 
    
    V2X primarily focuses on structured road environments (e.g., urban streets and highways), where wireless channels vary rapidly due to high mobility, but the underlying physical layout and communication infrastructure are highly structured.
    M2M operates predominantly in semi-static settings (e.g., smart homes and factories), where wireless channels are more stable and predictable due to low or no mobility. R2X, however, extends mobility to diverse and unstructured locations, such as indoor facilities, industrial sites, agricultural fields, and off-road environments. Robots can navigate uneven surfaces, stairs, disaster sites, and natural environments. \tcb{To operate effectively in such complex and unpredictable environments, R2X systems require robust and flexible communications. They also rely on advanced AI functionalities for perception, decision-making, and adaptive control.}

    \item \textbf{Flexible Network Requirements} 
    
    V2X prioritizes URLLC for safety-critical applications.
  On the other hand, M2M focuses on low-power, low-bandwidth communication for massive deployments.
R2X uniquely combines and extends these requirements. \tcb{It demands URLLC-grade reliability/latency for real-time robot control in dynamic environments, and high-throughput data transmission for sensor data processing. Additionally, it requires energy-efficient communications for battery-operated platforms such as vehicles and drones.}
    It is clear that this multifaceted demand cannot be handled by the rule-based algorithms.
    
    \item \textbf{Diverse Sensor Integration} 
    
V2X relies on standard vehicular sensors such as cameras, LiDAR, and radar, for road-centric data. M2M uses simpler sensors, such as temperature, humidity, and pressure sensors, to monitor static environments.
R2X integrates a wider and more complex array of sensors, including tactile, thermal, chemical, and advanced visual sensors.
To efficiently deliver multimodal sensing data such as feature vectors or even text-based representations, a compression mechanism needs to be introduced.
For instance, transmitting a raw Full HD (1920×1080) image at 24-bit color requires roughly 6 MB per frame \tcb{(i.e., approximately 1.44~Gbps at 30~fps, or $\sim$192~Mbps even at a modest 4~fps; for reference, even a heavily compressed H.264/MP4 stream at the same resolution and 30~fps still requires $\sim$4--8~Mbps)}. In contrast, when DL-based compression methods such as SigLIP or ViT-based encoders are applied, the data can be compressed into a compact feature vector—typically in the range of 512 to 1024 dimensions—reducing the transmission cost to less than 10 KB.
To make this claim explicit, if a $D$-dimensional feature is quantized to $b$ bits per element, the payload is $Db$ bits $=\frac{Db}{8}$ bytes (excluding protocol headers). Thus, a 512--1024D feature vector corresponds to $0.5\sim 1.0$~KB for 8-bit quantization and $1.0 \sim 2.0$~KB for 16-bit quantization; even with modest metadata and transport headers, the payload remains well below 10~KB per update. \tcb{Assuming a typical control-cycle update frequency of 1--10~Hz, this translates to an uplink data rate of roughly 40--160~kbps, meaning the semantic feature payload is \emph{at least 13$\times$ smaller per update} than the H.264/MP4 baseline noted above, and over 100$\times$ smaller at 1~Hz update rates---more than three orders of magnitude below the raw streaming requirement.}

    \item \textbf{Energy and Size Constraints}
    
    Energy constraints are less critical for V2X systems, while M2M systems are more sensitive to power consumption.
    R2X has more stringent energy constraint. \tcb{This is because robots, especially mobile platforms, face significant energy limitations due to compact battery sizes. At the same time, they must support both high computational power and continuous mobility, creating a stringent energy budget.}     Thus, \tcb{MLLMs (and simpler LLMs where unimodal processing suffices)} need to seriously consider the balance between performance and energy.
    For example, by receiving lower resolution sensing information, power consumption and transmission latency can be reduced at the cost of marginal degradation in performance.
    
    \item \textbf{Control Time Scale}
    
    While V2X operates primarily in real-time for immediate safety decisions, M2M typically involves slower periodic or event-driven updates (seconds to minutes). R2X is distinct from V2X and M2M since it should handle both immediate and long-term control processes over varying timescales. Specifically, it demands real-time control for tasks like robotic manipulation and obstacle avoidance, and periodic control for higher-level task planning and environmental mapping. 
    To support this, the R2X should support flexible numerology, variable packet sizes, adaptive code rates, and dynamic resource reservation for time-critical tasks.
\end{itemize}

\subsection{Computation for R2X Networks}

\begin{table*}[tbh!]
    \centering
    \caption{Summary of resource management schemes in different network architectures.}
    \renewcommand{\arraystretch}{1.8}
    \setlength{\tabcolsep}{1pt}
    \scriptsize
    \begin{tabular}{ccccccc}
    \hline
    \textbf{Ref.} & \textbf{Resources} & \textbf{Architecture} & \textbf{Application} & \textbf{AI model} & \textbf{Control strategy} & \textbf{Features in robotics} \\
    \hline
    \cite{Kwak15} & Processing \& networking & Mobile-cloud & cloud storage, offloading & None & Distributed & Energy efficiency \\
    \hline
    \cite{Gao19} & Processing \& networking & Mobile-edge-cloud & IoT applications & None & Centralized \& distributed & Load balancing \\
    \hline
    \cite{Chen19} & Processing \& networking & Mobile-SBS-MBS & Hierarchical edge computing & None & Distributed & Hybrid control\\
    \hline
    \cite{Li23} & Storage \& networking & Mobile-edge & Contents/service caching & None & Centralized \& distributed & AI model caching \\
    \hline 
    \cite{Krolikowski18} & Storage \& networking & Mobile-BS & Video streaming & None & Centralized \& distributed & Robot-human interaction \\
    \hline 
    \cite{Ham23} & Storage \& networking & Mobile-edge-cloud & Cloud gaming & None & Centralized & Low latency  \\
    \hline 
    \cite{Liang23} & Processing \& networking & Mobile-cloud & Computer vision & DNN model & Centralized & Hybrid inference\\
    \hline
    \cite{Choi2024VisionScaling} & Processing \& networking & Mobile-edge & XR, autonomous vehicle & DNN model & Distributed & Energy efficiency \\
    \hline
    \end{tabular}
\label{tab:resource_management_summary}
\end{table*}

R2X networks operate under one of three computing paradigms, on-device, edge, or centralized (cloud), each with distinct trade-offs in latency, computational power, and connectivity.
On-device computing enables real-time response and autonomy by processing data locally on robots, but is limited by hardware constraints.
Edge computing enhances processing capabilities by the help of nearby servers (e.g., roadside units), balancing latency and model complexity.
Centralized computing in the cloud supports large-scale inference and coordination with abundant resources, though it incurs higher latency and depends on reliable connectivity.
Selecting an appropriate architecture is crucial to meet the diverse performance and resource demands of R2X applications.

In traditional resource allocation within hierarchical mobile-edge-cloud network architectures, network, storage, and computing resources are integrated by carefully considering the trade-offs associated with each layer for efficient robotic networks. For instance, mobile robots are characterized by limited computing and storage capabilities but excel in providing real-time responsiveness. On the other hand, cloud servers offer substantial computing and storage resources and also incur additional network latency, making it challenging to guarantee fixed deadlines for time-sensitive tasks. Keeping these factors in mind, various multiple resource management techniques in mobile systems have been proposed \cite{Kwak15, Kim23, Gao19, Li23, Liang23, Krolikowski18, Chen19, Choi2024VisionScaling}. 

First, dynamic code offloading has been extensively studied as a means to reduce energy consumption and computational load on mobile devices by offloading tasks to edge or cloud servers. For instance, Kwak {\em et al.} \cite{Kwak15} optimized offloading policies by considering dynamic factors such as network conditions, workload variability, and CPU clock frequency scaling. These systems demonstrate significant improvements in energy efficiency while meeting latency requirements. 
Second, caching strategies in hierarchical mobile-edge-cloud architectures have evolved to dynamically adapt to spatio-temporal variations in content demand. Hybrid edge and cloud caching \cite{Li23} leverages the complementary strengths of edge caching for the latency reduction and cloud caching for broader user coverage, optimizing caching decisions based on content popularity and network conditions. 
Third, recent advances in resource management have focused on integrating deep neural network (DNN) model to optimize the computing and networking resource allocation in edge-mobile network. For instance, dynamic model partitioning frameworks \cite{Liang23} split DNNs between mobile devices and edge servers, thereby improving latency and inference accuracy. 

\begin{itemize}
\item \textbf{On-Device (Fully Distributed)} 

In the fully distributed on-device layer, memory capacity and CPU/GPU performance are limited, and energy consumption is critical, especially for mobile or battery-operated robots. This layer becomes valuable under poor network conditions where the robot has limited or no connectivity. While smaller models can make fast and, localized inference, their accuracy is lower and may struggle with fine-grained perception. For instance, a personal assistant robot navigating at home can rely on a small on-device model to detect obstacles and adjust its path in real time, ensuring safe movement without depending on the edge.

\item \textbf{Edge Computing (Hybrid Centralized/Distributed)} 

The hybrid layer leverages edge computing to balance real-time processing with higher model complexity. Tasks are split according to latency requirements, with heavier inference stages offloaded to nearby edge servers. For example, a delivery robot may handle immediate obstacle avoidance locally while sending high-resolution visual data to the edge for route optimization or semantic scene understanding under traffic conditions. Likewise, a security robot patrolling a facility can detect motion or intrusions on-device, but offload facial recognition or anomaly analysis to the edge when deeper computation is needed.

\item \textbf{Cloud Computing (Fully Centralized)}

In the fully centralized cloud layer, memory, storage, and processing resources are abundant, allowing for the deployment of larger, highly complex AI models. 
For instance, the robot could offload extensive image analysis tasks, like facial recognition or multi-object tracking, to the cloud with the largest DNN models. 
However, the trade-off is that cloud processing incurs higher network latency and congestion, making it unsuitable for immediate decision-making but ideal for complex tasks that tolerate slight delays, especially when the network condition is good.
In a manufacturing plant, robots collect detailed performance data from machinery. For immediate diagnostics, the robots run basic anomaly detection on-device. However, they periodically offload this data to the cloud, where advanced generative AI models analyze patterns and predict maintenance needs with high accuracy, leveraging vast historical data and complex models that on-device resources could not handle.
\end{itemize}

As an example of trade-offs, we compare on-device LLM, edge computing LLM, and cloud LLM in Table \ref{tab:throughput_comparison}. On-device LLMs can respond quickly to simple queries, achieving a decoding throughput of approximately 3–10 tokens per second. However, as the complexity of the input increases, the response latency grows significantly, often making real-time interaction impractical. Furthermore, throttling events—automatic reductions in processor speed to prevent overheating—are frequently observed when on-device LLMs are executed on smartphones due to thermal constraints. To mitigate such events, it is essential to dynamically control CPU, GPU, and RAM clock frequencies, in conjunction with intelligent offloading strategies.

In contrast, cloud-based LLMs deliver the highest throughput for both prefill and decode stages.
Prefill refers to the initial stage where the entire input prompt is processed in a single forward pass to initialize attention states.
Decode refers to the autoregressive generation phase, where the model produces output tokens one by one, using previously generated tokens.
Cloud-based LLMs provide fast and accurate responses, but they incur substantial operational costs and potential privacy concerns. Edge-based LLMs serve as a middle ground, offering a trade-off between performance, latency, and resource demands, positioned between cloud and on-device solutions. 

\begin{table*}[tbh!] \label{fig:table_compare_llm}
\centering
\caption{Comparison of estimated throughput between cloud, edge, and on-device LLMs.}
\small
\label{tab:throughput_comparison}
\begin{tabular}{@{}lccc@{}}
\toprule
\textbf{Metric} & \textbf{Cloud LLM} & \textbf{Edge LLM} & \textbf{On-Device LLM} \\ \midrule
Processor & A100 GPU & A10 GPU & Octa-core ARM CPU \\
Model & ChatGPT-4o & LLaMA 2 & TinyLLaMA \\
Model size & 100B+ & 13--30B & 1--7B \\
Prefill throughput & 500--1000 tokens/s & 80--250 tokens/s & 10--30 tokens/s \\
Decode throughput & 200--400 tokens/s & 30--100 tokens/s & 3--10 tokens/s \\
Connectivity & Internet & Local & Not required \\
User-side power/heat & None & None & High \\
Operational cost & High (API-based) & Medium (server cost) & Low (Device only) \\ \bottomrule 
\end{tabular}
\end{table*} 

By structuring tasks across on-device, edge, and cloud layers, this architecture optimally distributes inference time, model complexity, and resource demand. Simple and time-critical tasks are handled locally to ensure responsiveness, while more complex and less time-sensitive tasks benefit from the extensive resources and high accuracy of cloud-based AI, making the R2X network both efficient and adaptable.

\subsection{Integration of Sensing, Communication, and Computation for R2X Networks}

In R2X networks, efficient robotic autonomy emerges from the tight integration of sensing, communication, and computation rather than from each component operating in isolation. Multimodal sensing captures visual, geometric, auditory, and tactile information, which is first processed through lightweight on-device models for low-latency perception and reflexive control. Task-relevant features or compressed representations are then shared through reliable communication links to edge servers, where more complex reasoning, such as semantic scene understanding or task planning, can be executed. When the large-scale mapping, cross-robot coordination, or long-horizon prediction is required, cloud computation further provides global knowledge and historical context. This hierarchical collaboration enables robots to perceive locally, think collectively, and act intelligently, ensuring scalable and robust operation even under dynamic environments and fluctuating network conditions.
\tcb{Viewed as the intent-conditioned joint optimization introduced in Section~\ref{sec:intro}, the orchestrator jointly selects sensing, communication, and computation strategies to maximize task success under goal-dependent constraints (latency, bandwidth, and energy budgets).}

\section{Experimental Demonstrations and Use Cases}
\label{sec:apps_intro} 

Sections~\ref{sec:intro}--\ref{sec:core_components} motivate R2X as a joint design of sensing,
communication, and computation for MLLM-driven robotic networks. In this section, we ground the discussion with end-to-end demonstrations that make the orchestration loop explicit:
\emph{sense} $\rightarrow$ \emph{communicate} $\rightarrow$ \emph{compute} $\rightarrow$ \emph{act}.
Our emphasis is mainly on system-level, measurable quantities---latency, reliability, and task completion time---rather than isolated algorithmic gains.
\tcb{Following the intent-to-resource orchestration principle, each demo is initiated by a high-level language instruction, and the orchestrator translates this intent into concrete sensing/communication/computation configurations.}
Accordingly, we evaluate not only link-layer metrics but also \emph{task-facing} outcomes that reflect how well the system executes the intended goal under constrained sensing/compute/wireless resources.

We present four end-to-end demonstrations that progressively increase realism and capability. First, we build a ray-tracing digital twin of an indoor warehouse and evaluate an LLM-based orchestrator that combines semantic visual
compression with predictive link-quality awareness to support collision-free multi-robot navigation. 
Second, we study proactive modulation-and-coding (MCS) control in a mobility scenario, where a central orchestrator uses compact semantic features to anticipate short-term channel variations and stabilize throughput, latency, and BLER under delayed feedback.
Third, we implement a real-hardware FollowMe robot prototype in which the orchestrator switches between conventional image streaming and
compact \emph{codeword-index} transmission based on measured WiFi conditions, demonstrating a practical semantic-sensing switch for low-latency operation. 
Fourth, we demonstrate real-hardware open-vocabulary trash sorting, where an edge-assisted MLLM grounds text instructions to unseen objects and out-of-FOV bins via multi-view sensing.

\subsection{Digital-Twin Demo I: Warehouse Navigation}
\label{sec:demo1_warehouse}

\subsubsection{Demo objective and orchestration knobs}
We first instantiate the R2X orchestration loop in a ray-tracing digital twin of an indoor warehouse,
where two mobile robots must reach their destinations while avoiding dynamically moving human workers.
The key challenge is that \emph{safe} and \emph{efficient} navigation requires (i) global reasoning
(e.g., multi-agent replanning and short-horizon human-trajectory anticipation) and (ii) reliable uplink
connectivity for timely perception offloading---both of which are difficult to guarantee with
standalone on-device autonomy.

\begin{figure*}[t]
    \centering
    \includegraphics[width=0.86\textwidth]{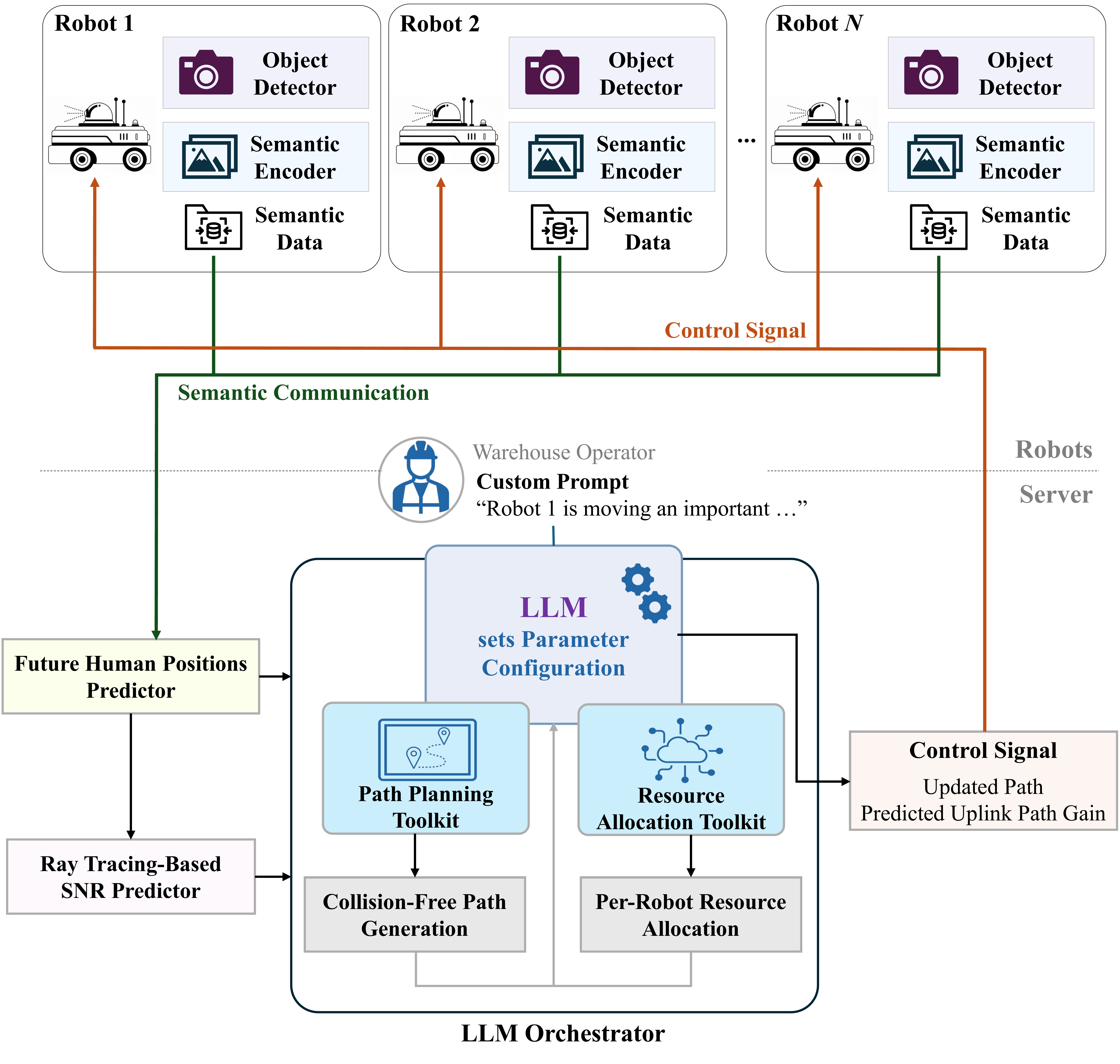}
    \caption{End-to-end architecture of Demo~I (warehouse digital twin): robots uplink semantic visual observations and positions; the server orchestrator fuses sensing and predicted link states to trigger global replanning and proactive uplink configuration, then downlinks updated commands. }
    \label{fig:demo1_system}
\end{figure*}

Figure~\ref{fig:demo1_system} summarizes the end-to-end pipeline. The server-side orchestrator
integrates three coupled control knobs:
(i) \textbf{sensing orchestration} (raw RGB-D streaming vs.\ compact semantic features),
(ii) \textbf{communication orchestration} (reactive vs.\ predictive link adaptation), and
(iii) \textbf{computation orchestration} (local stop-and-go vs.\ centralized global replanning).
In this demo, the orchestrator emits lightweight control messages that \emph{parameterize} existing toolkits
for collision-free path planning and proactive uplink power/resource configuration.
This design deliberately avoids end-to-end ``LLM-as-a-solver'' operation: the LLM specifies objectives, priorities,
and constraints in a schema-valid form, while deterministic solvers execute the actual optimization to ensure
reproducibility, auditability, and deadline-aware behavior.

\begin{tcolorbox}[colback=gray!5!white,
                  colframe=gray!55!black,
                  title=What is orchestrated in Demo~I (and why it is measurable),
                  fonttitle=\bfseries,
                  colbacktitle=gray!75!black,
                  coltitle=white,
                  breakable,
                  enhanced,
                  left=2mm, right=2mm, top=1mm, bottom=1mm]
\begin{itemize}
    \item \textbf{Sensing:} semantic encoding compresses visual observations by \(\sim\!1024{:}1\) (vs.\ 8-bit image transmission),
    reducing uplink payload and enabling faster offloading for server-side prediction and planning.
    \item \textbf{Communication:} ray-tracing-based future path-gain prediction provides a \emph{link-context signal} for proactive
    uplink power control (target SNR: \(15\)~dB), mitigating stop events caused by link drops.
    \item \textbf{Computation:} centralized multi-agent replanning avoids purely reactive behavior (e.g., repeated halting) and improves task completion time at the system level.
\end{itemize}
\end{tcolorbox}

\subsubsection{Digital-twin construction and sensing context}
\label{sec:demo1_dt_construction}
We construct a realistic 3D warehouse map in NVIDIA Isaac Sim using the \emph{Warehouse Creator} presets and built-in assets
(e.g., racks and equipment). The scene is saved as a USD file and used as the digital twin for both dataset generation and
communication-aware evaluation. To make the scene tractable for wireless simulation and channel extraction, we apply an
offline mesh optimization pipeline that (i) merges meshes with common characteristics to reduce scene complexity and (ii)
triangulates meshes and tags material properties for accurate propagation modeling.

To generate the perception dataset used by the semantic encoder and predictor, we render synchronized \emph{above-view} and
\emph{robot-view} RGB-D images from the digital twin. Images are sampled every \(0.5\)~s. Robot mobility is later constrained to a
\(40\times 40\) grid with \(2\)~m spacing; with a nominal speed of \(\sim 1.5\)~m/s, traversing one grid cell takes \(\approx 1.4\)--\(1.5\)~s
(\(\approx 3\) image frames), which motivates short-horizon prediction and low-latency orchestration.

\begin{figure*}[t]
    \centering
    \begin{subfigure}[b]{0.24\textwidth}
        \centering
        \includegraphics[width=\linewidth]{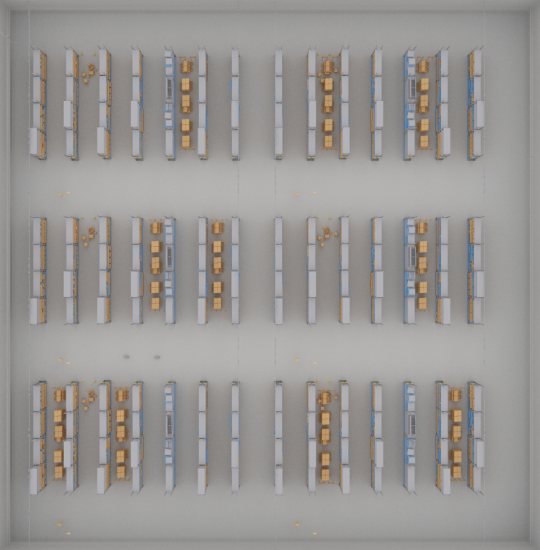}
        \caption{Top view (geometry)}
        \label{fig:demo1_topview}
    \end{subfigure}
    \hfill
    \begin{subfigure}[b]{0.24\textwidth}
        \centering
        \includegraphics[width=\linewidth]{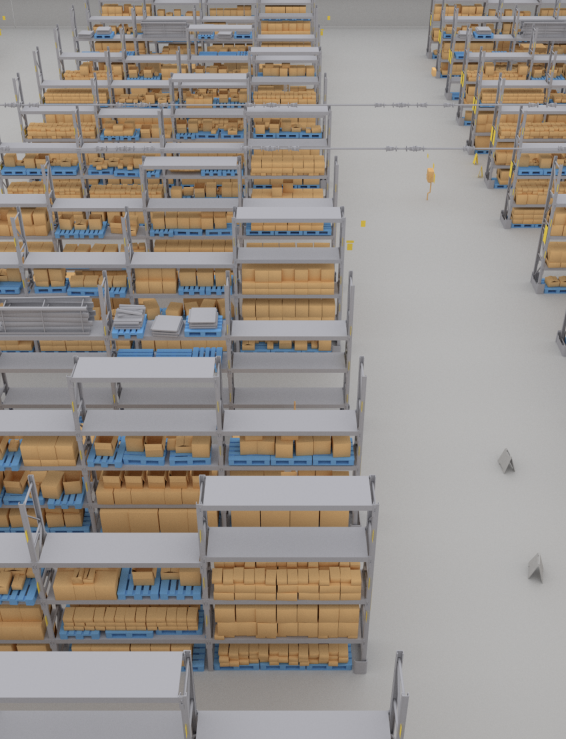}
        \caption{Above view (3D)}
        \label{fig:demo1_aboveview}
    \end{subfigure}
    \hfill
    \begin{subfigure}[b]{0.24\textwidth}
        \centering
        \includegraphics[width=\linewidth]{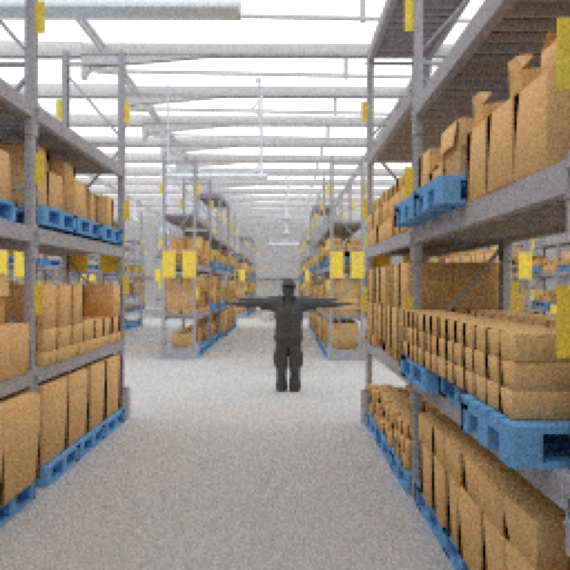}
        \caption{Robot-eye snapshot}
        \label{fig:demo1_roboteye}
    \end{subfigure}
    \hfill
    \begin{subfigure}[b]{0.24\textwidth}
        \centering
        \includegraphics[width=\linewidth]{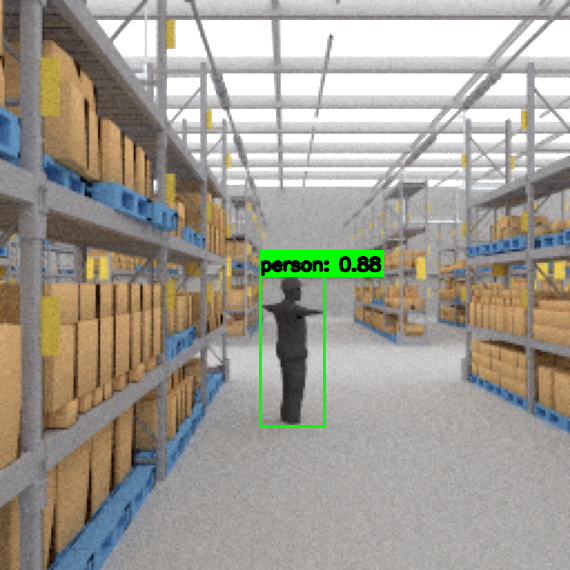}
        \caption{On-device detection}
        \label{fig:demo1_detection}
    \end{subfigure}
    \caption{Demo~I digital-twin and perception snapshots: (a)--(c) multiple viewpoints of the Isaac Sim warehouse scene and (d) an example robot-view frame with the on-device detection trigger used to request server-side replanning.}
    \label{fig:demo1_scene_and_detection}
\end{figure*}

Figure~\ref{fig:demo1_scene_and_detection} shows representative snapshots of the digital twin and the on-device detection trigger: a robot-eye view and the corresponding human detection output produced by a lightweight, single-stage object detection algorithm based on the YOLO architecture, which performs real-time bounding-box regression and classification in a unified forward pass.

\subsubsection{Mobility abstraction and link-context construction}
\label{sec:demo1_link_context}
For navigation and multi-agent planning, the warehouse floor is discretized into a \(2\)~m grid; invalid cells (e.g., racks/walls)
are marked as non-traversable. At each decision instant, the server updates a ray-tracing scene using the predicted near-future
human locations (from the semantic features) and computes prospective uplink path-gain values over candidate robot positions.
This yields a \emph{link-context signal} used for proactive link adaptation (e.g., power control to maintain a target SNR of \(15\)~dB).

\begin{figure*}[t]
    \centering
    \begin{subfigure}[b]{0.32\textwidth}
        \centering
        \includegraphics[width=\linewidth]{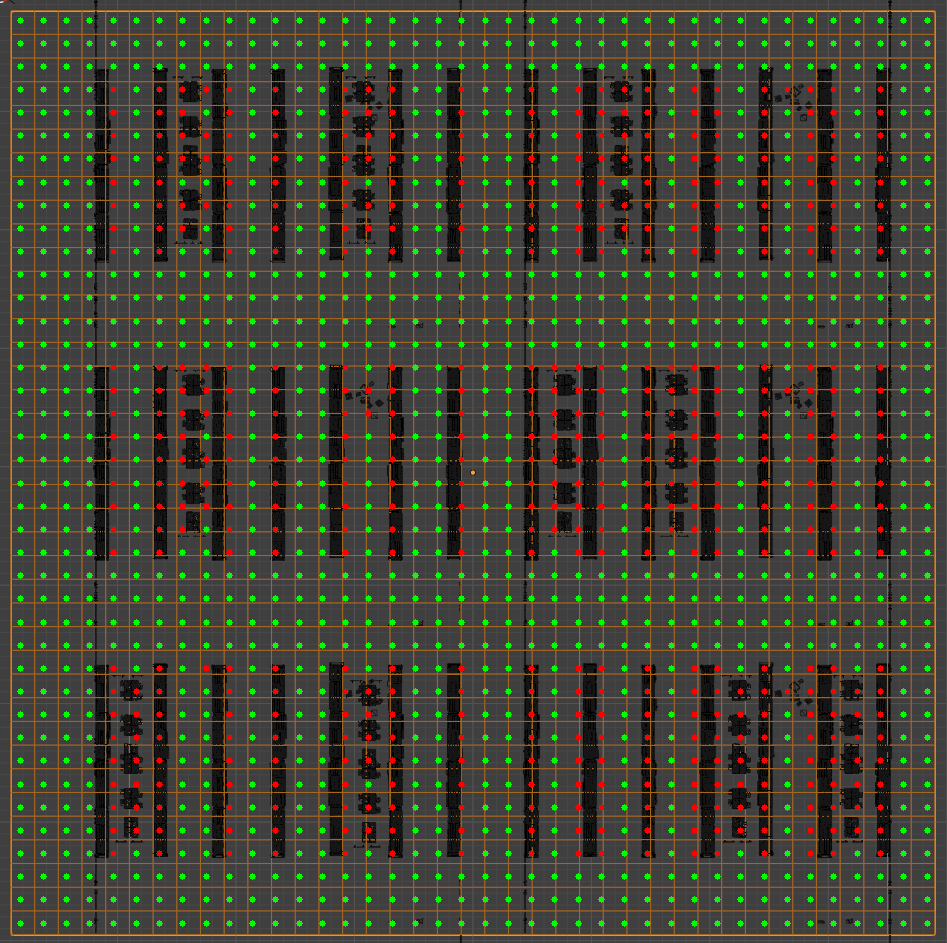}
        \caption{Valid/invalid navigation grid}
        \label{fig:demo1_grid}
    \end{subfigure}
    \hfill
    \begin{subfigure}[b]{0.32\textwidth}
        \centering
        \includegraphics[width=\linewidth]{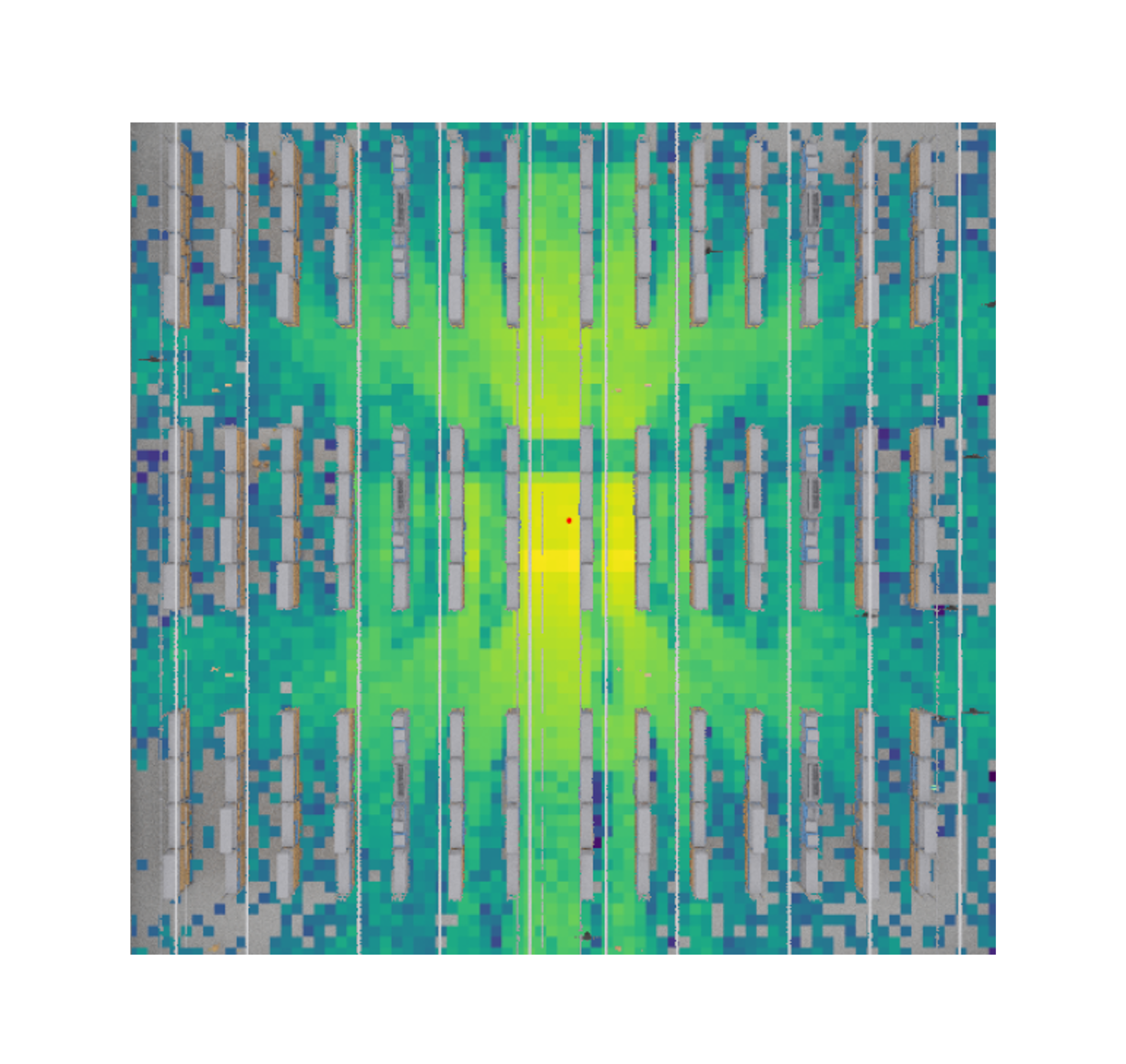}
        \caption{Example radio map}
        \label{fig:demo1_radiomap}
    \end{subfigure}
    \hfill
    \begin{subfigure}[b]{0.32\textwidth}
        \centering
        \includegraphics[width=\linewidth]{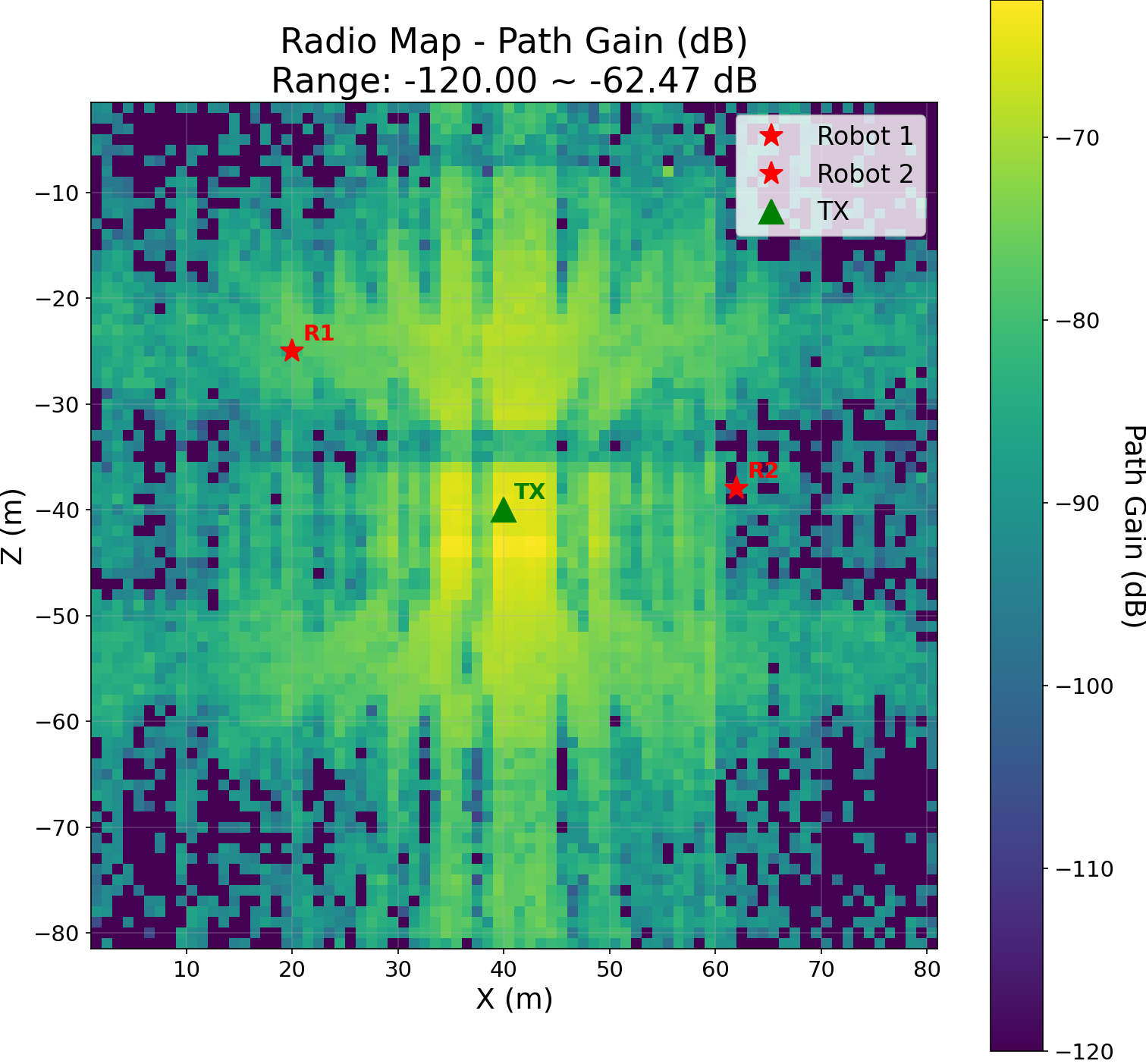}
        \caption{Path-gain from radio map}
        \label{fig:demo1_pathgain}
    \end{subfigure}
    \caption{Demo~I link-aware planning context: (a) grid-based mobility abstraction and (b)--(c) ray-tracing-derived radio context consumed by the orchestrator for proactive link-aware control.}
    \label{fig:demo1_grid_and_radio}
\end{figure*}

Figure~\ref{fig:demo1_grid_and_radio} illustrates (i) the valid/invalid grid abstraction and (ii) representative radio-map and
path-gain outputs generated by the ray-tracing engine.

\subsubsection{Semantic sensing and future-human prediction}
\label{sec:demo1_semantic}
To reduce uplink overhead while preserving the information needed for server-side reasoning, each robot transmits
compact semantic features instead of raw RGB-D frames. The semantic encoder is an end-to-end CNN that compresses
RGB-D observations (and robot state) by \(\sim 1024{:}1\), and is trained under a target SNR of \(15\)~dB (noise injected to match
the operating point). The predictor consumes the past \(5\) frames of semantic features and forecasts near-future human
locations over a horizon that covers (at least) one grid transition (\(\approx 1.4\)--\(1.5\)~s, \(\approx 3\) frames).

For later comparison with bit-based streaming (LORC-P), we summarize the sensing payload in Table~\ref{tab:demo1_payload}, where we compare the raw uplink size used in our bitstream baseline and the semantic-feature payload size used in the proposed design.

\begin{table*}[t]
\centering
    \caption{Demo~I sensing payload comparison (uplink).}
    \label{tab:demo1_payload}
    \small
    \renewcommand{\arraystretch}{1.15}
    \begin{tabular}{lcc}
        \toprule
        \textbf{Uplink sensing type} & \textbf{Payload (per update)} & \textbf{Notes} \\
        \midrule
        Raw image streaming (bit-based) & \(\approx 1.31\)~MB & 8-bit quantized images (LORC-P) \\
        Semantic features (proposed) & \(\approx 1280\) real values & e.g., \(256\times 5\) features (LORC-SC/LORC-SC-P) \\
        \bottomrule
    \end{tabular}
\end{table*}

\subsubsection{LLM orchestrator: language-to-solver configuration (inputs/outputs and controllable knobs)}
\label{sec:demo1_llm}

\paragraph{Role of the orchestrator.}
The server-side LLM orchestrator does \emph{not} numerically solve the underlying planning/optimization problems.
Instead, it acts as a \emph{configuration layer} that converts (i) an operator's natural-language intent and
(ii) the current multi-modal system context into a \textbf{schema-valid JSON control message}.
This JSON message parameterizes two deterministic solvers running on the server:
(i) a \textbf{multi-robot collision-free path planner} and (ii) a \textbf{wireless resource/power configurator}.
This design makes the orchestration loop explicit and measurable: the LLM changes \emph{objectives, priorities, and constraints}
without changing the solvers themselves.
In this sense, the operator intent defines the optimization goal (e.g., makespan vs.\ safety) and constraint strictness, while the system chooses the most resource-efficient sensing/communication/computation configuration that satisfies the intent.

\paragraph{Why parameterization instead of end-to-end LLM solving?}
In safety-critical robotic networking, end-to-end decisions must be \emph{reproducible} and \emph{verifiable} under strict timing
constraints.
Treating the LLM as a high-level orchestrator (rather than a numerical solver) provides a clear connection:
(i) the LLM expresses task intent in a constrained schema (objectives/constraints/priorities), while (ii) established solvers execute
the optimization with deterministic behavior and explicit feasibility checks.
This separation reduces failure modes such as invalid plans or schema drift, enables auditable decision logs via JSON, and allows the
system to enforce conservative fallbacks when deadline or feasibility checks fail.

\vspace{0.3em}
\paragraph{What exactly goes into the LLM (input contract).}
At each decision instant, the orchestrator packages the following inputs:
\begin{itemize}
    \item \textbf{Operator instruction (text):} a short free-form sentence specifying mission intent (e.g., urgency, safety, fairness).
    \item \textbf{Robot state (structured):} current grid cell, goal cell, and current motion status for each robot.
    \item \textbf{Predicted human mobility (structured):} near-future human locations over the short horizon used in this demo
    (covering at least one grid transition, \(\sim\!1.4\)~s).
    \item \textbf{Link context (visual + scalar):} ray-tracing outputs consisting of a spatial radio-map associated path-gain values, together with per-robot predicted path-gain at candidate next cells. This information is used to reason about link reliability and to support proactive uplink configuration.
\end{itemize}

\noindent Importantly, the link context is structured in a form suitable for LLM-based spatial reasoning,
where the radio-map provides spatial context and the path-gain values encode link quality
(e.g., the radio-map and corresponding path-gain visualizations in Fig.~\ref{fig:demo1_grid_and_radio}).

\paragraph{What comes out of the LLM (output contract).}
The LLM outputs a \textbf{single JSON object} that specifies:
(i) \texttt{pp\_config} for path planning, and (ii) \texttt{ra\_config} for resource/power configuration.
The downstream solvers consume only this JSON object, which makes all orchestration decisions explicit and auditable.

\begin{tcolorbox}[colback=gray!3!white,
                  colframe=gray!55!black,
                  title=LLM I/O contract used in Demo~I,
                  fonttitle=\bfseries,
                  colbacktitle=gray!75!black,
                  coltitle=white,
                  breakable,
                  enhanced,
                  left=2mm, right=2mm, top=1mm, bottom=1mm]
\noindent\textbf{LLM input:}
\begin{itemize}
    \item operator text instruction (e.g., "Robot 1 is moving an important cargo...")
    \item robot states (start/goal/current cell, motion status)
    \item predicted human locations (short horizon)
    \item ray-tracing link context (radio-map/path-gain map + predicted per-robot path-gain)
\end{itemize}
\noindent\textbf{LLM output:} one schema-valid JSON with keys
\(\{\)\texttt{pp\_config}, \texttt{ra\_config}\(\}\). 
\end{tcolorbox}

\vspace{0.3em}
\paragraph{Path planning: objectives, constraints, and parameters controlled by the LLM.}
The path planner operates on the \(2\)~m grid and produces time-indexed robot trajectories.
Let robot~$i$ follow a discrete-time path
$\pi_i = \{v_i(t)\}_{t=0}^{T_i}$,
where $v_i(t)$ denotes the grid cell occupied by robot~$i$ at discrete time step
$t \in \{0,1,\ldots,T_i\}$, and $T_i$ is the time horizon required for robot~$i$
to reach its destination.
The planner enforces the following \textbf{hard constraints}:

\begin{itemize}
    \item \textbf{Map feasibility:} \(v_i(t)\) must be a valid (traversable) cell; moves occur only between 4-neighbor adjacent cells
    (or \emph{wait} actions).
    \item \textbf{Robot--robot collision avoidance:}
    (i) \emph{vertex conflict} \(v_i(t) \neq v_j(t)\) for all \(i\neq j\),
    and (ii) \emph{edge conflict} \((v_i(t),v_i(t{+}1)) \neq (v_j(t{+}1),v_j(t))\) (no swap).
    \item \textbf{Human-aware safety constraints (from prediction):} cells/time-steps deemed unsafe by the near-future human
    prediction are treated as forbidden (or equivalently, imposed as constraints in the planner) to prevent near-collisions.
\end{itemize}

\noindent On top of these hard constraints, the LLM configures the planning objective.
In our implementation, the LLM selects between:
\begin{itemize}
    \item \textbf{Makespan-oriented operation:} minimize \(\max_i T_i\) to complete the multi-robot task quickly.
    \item \textbf{Safety-oriented operation:} prioritize safer motion by tightening temporal/spatial constraints around predicted human
    locations and by enforcing stronger separation at shared/conflicting cells.
\end{itemize}
\noindent The planner is implemented using a \textbf{prioritized Conflict-Based Search (CBS)} framework:
conflicts are detected on candidate paths and resolved by adding constraints; the LLM controls which robot receives priority
when conflicts occur and how conservative the temporal separation should be.

\vspace{0.3em}
\paragraph{Resource/power configuration: objectives, constraints, and parameters controlled by the LLM.}
Using predicted path-gain values from ray tracing, the wireless configurator sets per-robot uplink parameters to ensure timely
perception offloading. In this demo, we set the key QoS target as \textbf{the target uplink SNR of \(15\)~dB}.
Given predicted path-gain \(g_i\) for robot \(i\) at the next location, the configurator computes a required transmit power
to satisfy
\(
\mathrm{SNR}_i \ge \gamma_{\mathrm{tar}}
\)
with \(\gamma_{\mathrm{tar}}=15\)~dB, subject to practical power limits.
When multiple robots contend for wireless resources or when stricter safety priorities are requested, the LLM configures the following:
\begin{itemize}
    \item \textbf{Fairness objective:} \emph{max--min fairness} (protect the worst robot) vs.\ \emph{proportional fairness}
    (balanced throughput/latency trade-off).
    \item \textbf{Per-robot priority weights:} weights reflecting mission criticality (e.g., medical supplies robot gets higher priority),
    used by the configurator to bias the chosen fairness policy.
\end{itemize}

\noindent The output of this module is a per-robot communication configuration (e.g., transmit power and/or scheduling priority) used at the next decision instant.

\paragraph{LLM-configurable knobs (explicit schema).}
Table~\ref{tab:demo1_knobs} summarizes the exact knobs exposed to the LLM in Demo~I, with each knob mapped to its corresponding solver.
These fields are \emph{the only degrees of freedom} the LLM is allowed to modify.

\begin{table*}[t]
    \centering
    \caption{Demo~I: LLM-configurable knobs for path planning and resource/power configuration (JSON fields and meanings).}
    \label{tab:demo1_knobs}
    \small
    \renewcommand{\arraystretch}{1.15}
    \setlength{\tabcolsep}{4pt}
    \begin{tabular}{p{0.38\linewidth} p{0.58\linewidth}}
    \toprule
    \textbf{JSON field} & \textbf{Meaning (objective/constraint/parameter)} \\
    \midrule
    {\footnotesize\ttfamily pp\_config.objective} &
    Planning objective type:
    \texttt{makespan} (minimize \(\max_i T_i\)) or
    \texttt{safety\_first} (tighten safety constraints around predicted humans and conflicts). \\
    {\footnotesize\ttfamily pp\_config.priority\_robot} &
    Conflict priority policy:
    \texttt{robot\_1} / \texttt{robot\_2} / \texttt{none}.
    When a conflict is detected, the lower-priority robot is constrained (rerouted / delayed) first. \\
    {\footnotesize\ttfamily pp\_config.min\_time\_gap\_at\_conflict} &
    Temporal separation constraint (integer time steps) imposed at shared/conflicting cells to reduce close encounters,
    especially under safety-first operation. \\
    \midrule
    {\footnotesize\ttfamily ra\_config.fairness} &
    Wireless fairness policy:
    \texttt{max\_min} (protect worst-case robot) or
    \texttt{proportional} (maximize weighted \(\sum_i \log(\cdot)\) style utility). \\
    {\footnotesize\ttfamily ra\_config.priority\_weights} &
    Per-robot weights (e.g., \([w_1,w_2]\), \(w_1+w_2=1\)) used to reflect task criticality in the chosen fairness policy. \\
    \bottomrule
    \end{tabular}
\end{table*}

\paragraph{Meta-prompting for reliable JSON (schema enforcement + self-correction).}
\textit{LLM implementation.}
In our implementation, the server-side orchestrator is instantiated using the Llama~3 8B model.
\tcb{Because all inputs fed to this orchestrator—operator text, robot state vectors, predicted human locations, and link-context scalars—are serialized as structured text (JSON/natural language), this module is a \emph{unimodal LLM} rather than a full MLLM; no raw visual or sensor stream is passed directly to the model.}
The LLM is used solely as a high-level configuration engine that maps structured system context and
operator intent to a schema-valid JSON message, and does not directly solve numerical planning or
optimization problems.

We wrap the LLM with a \textbf{constrained meta-prompt} that:
(i) enumerates the allowed JSON keys/values in Table~\ref{tab:demo1_knobs},
(ii) requires outputting \emph{only} one JSON object (no extra text),
and (iii) provides an error-driven \textbf{self-correction} step if the output violates the schema.
Concretely, the orchestrator validates the LLM output; if parsing fails or a field is out-of-domain,
the LLM is re-prompted with the validation error message and asked to return a corrected JSON.
This mechanism yields schema-valid JSON in the vast majority of runs (with residual failures handled by the same correction loop).

\begin{tcolorbox}[colback=blue!2!white,
                  colframe=blue!60!black,
                  title=Example: operator prompt $\rightarrow$ schema-valid JSON configuration,
                  fonttitle=\bfseries,
                  colbacktitle=blue!70!black,
                  coltitle=white,
                  breakable,
                  enhanced,
                  left=2mm, right=2mm, top=1mm, bottom=1mm]
\noindent\textbf{Operator prompt (text):}
\begin{quote}\small
``Robot~2 is carrying high-priority medical supplies. The minimum quality of communication for each robot has to be guaranteed. The communication of Robot 2 is much more important than Robot 1. They have to move very safely.''
\end{quote}

\noindent\textbf{LLM output (single JSON object; directly consumed by both solvers):}
{\footnotesize\ttfamily
\{\\
\ \ "pp\_config": \{\\
\ \ \ "objective": "safety\_first",\\
\ \ \ "priority\_robot": "robot\_2",\\
\ \ \ "min\_time\_gap\_at\_conflict": 3\\
\ \ \},\\
\ \ "ra\_config": \{\\
\ \ \ "fairness": "max\_min",\\
\ \ \ "priority\_weights": [0.3, 0.7],\\
\ \ \}\\
\}
}
\end{tcolorbox}

\subsubsection{Baselines and evaluation protocol}
\label{sec:demo1_baselines}
To isolate the role of each orchestration knob, we compare four variants:
\emph{Stop-and-Go} (no communication), \emph{LORC-P} (bitstream uplink + predictive path-gain),
\emph{LORC-SC} (semantic uplink + reactive pilot-based power control), and
\emph{LORC-SC-P} (semantic uplink + predictive path-gain).
The primary metric is \textbf{task completion time} for two robots under four navigation scenarios:
short, medium, and long-path cases corresponding to Euclidean travel distances of 40, 70, and 100 m, respectively, as well as a long-path scenario with additional obstacles.

\begin{table*}[t]
    \centering
    \caption{Baseline variants in Demo~I (which orchestration components are enabled).}
    \label{tab:demo1_baselines}
    \small
    \renewcommand{\arraystretch}{1.15}
    \begin{tabular}{lccc}
        \toprule
        \textbf{Method} & \textbf{Semantic sensing} & \textbf{Predictive link context} & \textbf{Central replanning} \\
        \midrule
        Stop-and-Go  & -- & -- & -- \\
        LORC-P       & -- (raw images) & \checkmark & \checkmark \\
        LORC-SC      & \checkmark & -- (reactive pilots) & \checkmark \\
        LORC-SC-P    & \checkmark & \checkmark & \checkmark \\
        \bottomrule
    \end{tabular}
\end{table*}

\subsubsection{End-to-end timing breakdown}
\label{sec:demo1_timing}
A key feasibility criterion is whether the closed-loop orchestration finishes \emph{before} a robot traverses a grid cell
(\(\sim 1.4\)~s). Table~\ref{tab:demo1_latency} reports representative runtime measurements obtained in our implementation on a workstation with an AMD Ryzen 5950X CPU and two NVIDIA RTX~3090 GPUs, where LLM orchestration and perception pipelines are executed on separate GPUs.
The LLM step dominates the time budget (sub-second), while semantic preprocessing and ray-tracing are in the millisecond range.
Overall, the measured sensing-to-command round-trip time (RTT) is \(\sim 1.02\)~s for LORC-SC-P, which enables continuous motion.

\begin{table*}[t]
    \centering
    \caption{Demo~I experimental setup and latency breakdown (representative values).}
    \label{tab:demo1_latency}
    \small
    \renewcommand{\arraystretch}{1.15}
    \setlength{\tabcolsep}{6pt}
    \begin{tabular}{lcc}
        \toprule
        \multicolumn{3}{c}{\textbf{Experimental Setup}} \\
        \midrule
        CPU & \multicolumn{2}{c}{AMD Ryzen 5950X (16C/32T)} \\
        GPU & \multicolumn{2}{c}{2$\times$ NVIDIA RTX~3090 (24~GB VRAM)} \\
        GPU usage & \multicolumn{2}{c}{GPU 0: semantic encoder + YOLOv8n, GPU 1: LLaMA~3~8B } \\
        \midrule
        \multicolumn{3}{c}{\textbf{Latency Breakdown}} \\
        \midrule
        \textbf{Stage} & \textbf{Role in loop} & \textbf{Latency} \\
        \midrule
        Object detection (YOLOv8n) & on-robot trigger & \(96.17\)~ms \\
        Semantic encode + human prediction & uplink preprocessing & \(1.23\)~ms \\
        Ray-tracing path-gain prediction & link-context construction & \(46.91\)~ms \\
        LLM orchestration (LLaMA~3~8B) & JSON generation + solver selection & \(0.884\)~s \\
        \midrule
        Total (sensing-to-command RTT) & end-to-end closed loop & \(\approx 1.028\)~s \\
        \bottomrule
    \end{tabular}
\end{table*}

\subsubsection{Key results: completion time and qualitative trajectories}
Table~\ref{tab:demo1_perf} reports task completion times. The fully orchestrated configuration
(LORC-SC-P) consistently achieves the lowest completion time by reducing uplink overhead (semantic sensing)
while preventing link-induced halts (predictive link adaptation), thereby enabling continuous motion with timely server-side replanning.
In contrast, LORC-P incurs large uplink payloads due to raw streaming, which leads to increased RTT and enforced waiting.  
LORC-SC reduces the uplink payload size but still suffers from stop events under reactive link control.

\begin{table*}[t]
    \centering
    \caption{Demo~I: Task completion time (seconds) for two robots in four warehouse navigation scenarios.}
    \label{tab:demo1_perf}
    \small
    \renewcommand{\arraystretch}{1.15}
    \begin{tabular}{lcccc}
        \toprule
        \textbf{Method} &
        \shortstack{\textbf{Scenario 1}\\(Short)} &
        \shortstack{\textbf{Scenario 2}\\(Medium)} &
        \shortstack{\textbf{Scenario 3}\\(Long)} &
        \shortstack{\textbf{Scenario 4}\\(Long+Obstacles)} \\
        \midrule
        Stop-and-Go & 72.0 & 151.0 & 209.0 & 209.5 \\
        LORC-SC-P   & \textbf{66.0} & \textbf{77.5} & \textbf{108.0} & \textbf{117.0} \\
        LORC-P      & 110.0 & 127.0 & 179.5 & 189.5 \\
        LORC-SC     & 78.5 & 93.5 & 128.5 & 138.0 \\
        \bottomrule
    \end{tabular}
\end{table*}

\begin{figure*}[h!t]
    \centering
    \begin{subfigure}[b]{0.24\textwidth}
        \centering
        \includegraphics[width=\linewidth]{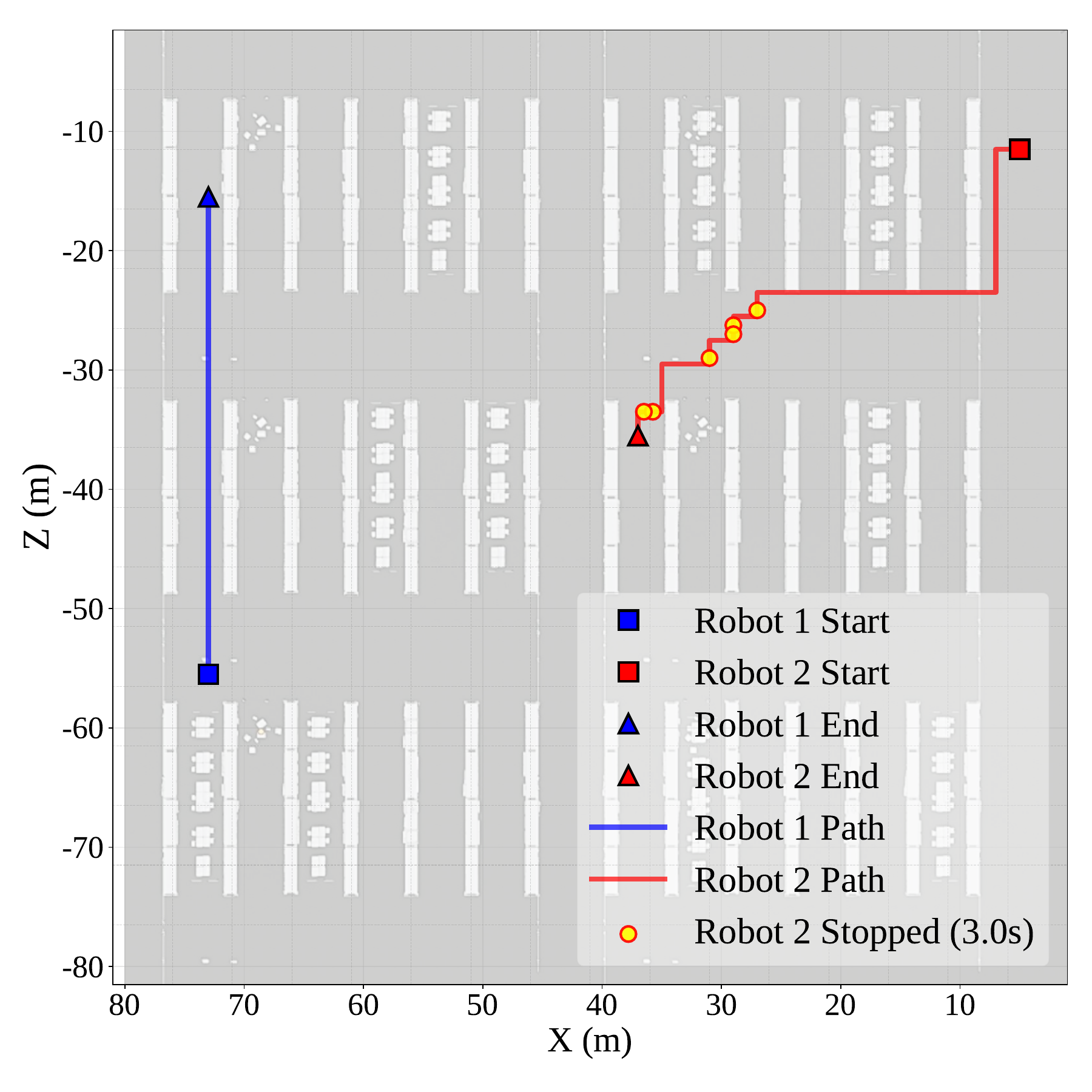}
        \caption{Stop-and-Go}
    \end{subfigure}
    \hfill
    \begin{subfigure}[b]{0.24\textwidth}
        \centering
        \includegraphics[width=\linewidth]{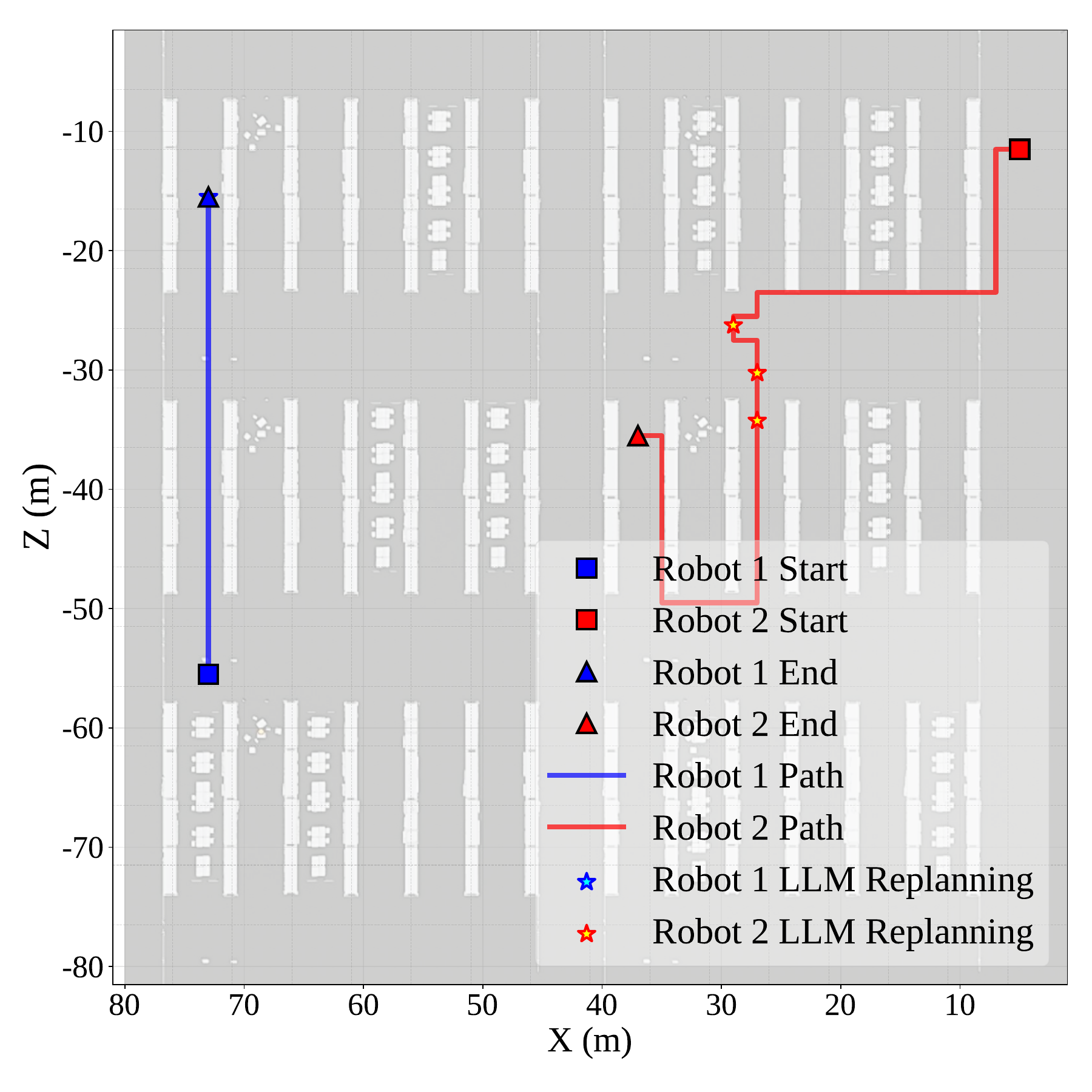}
        \caption{LORC-SC-P}
    \end{subfigure}
    \hfill
    \begin{subfigure}[b]{0.24\textwidth}
        \centering
        \includegraphics[width=\linewidth]{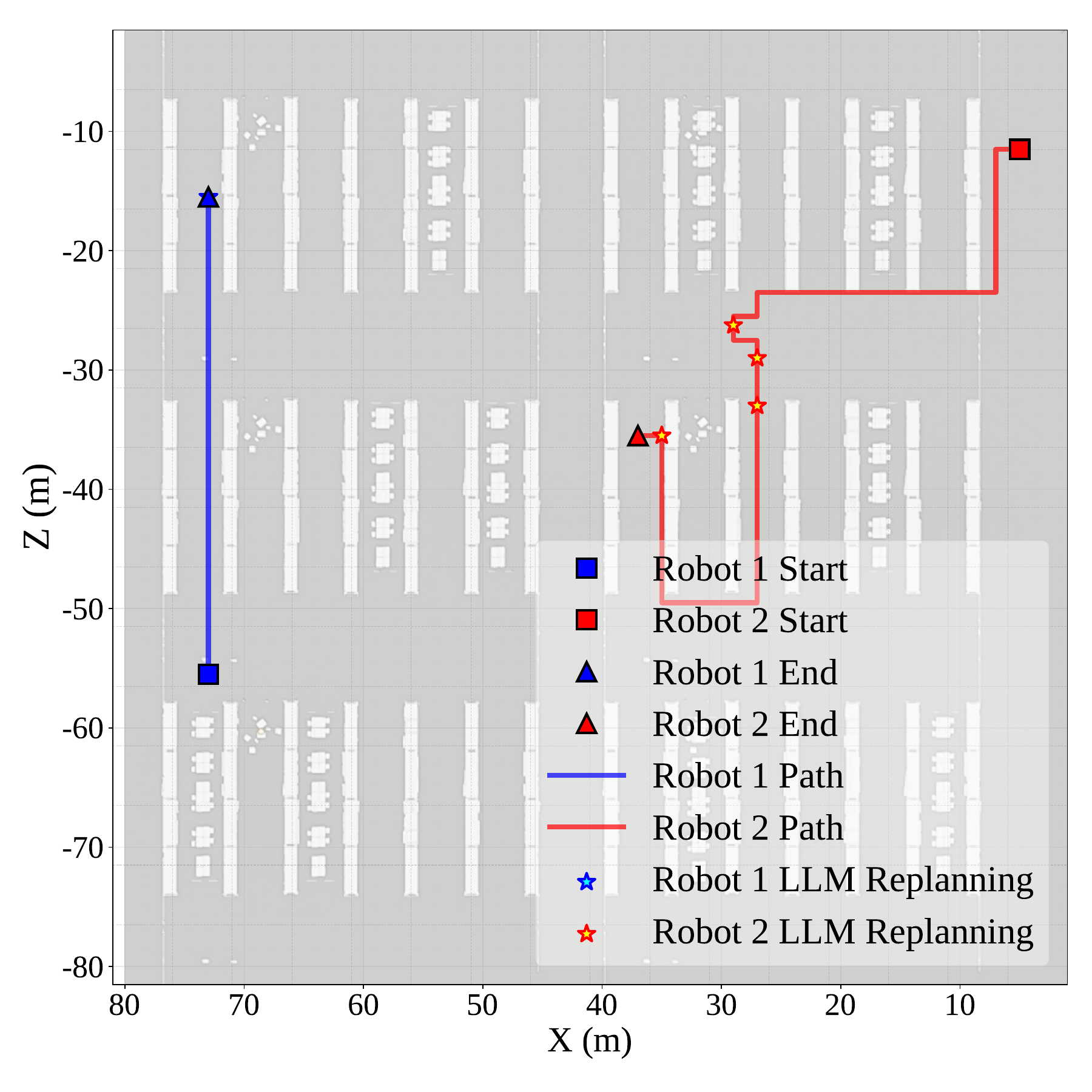}
        \caption{LORC-P}
    \end{subfigure}
    \hfill
    \begin{subfigure}[b]{0.24\textwidth}
        \centering
        \includegraphics[width=\linewidth]{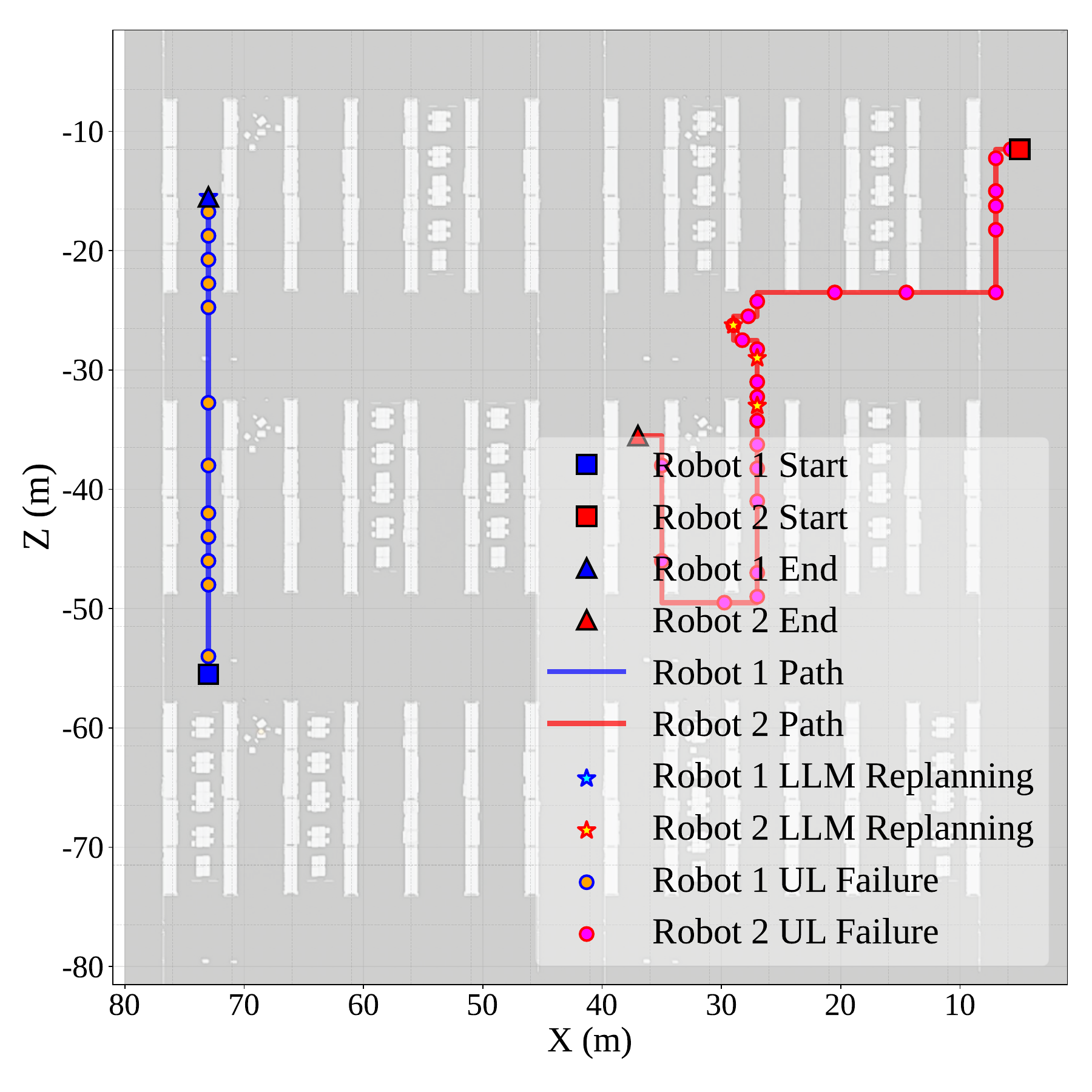}
        \caption{LORC-SC}
    \end{subfigure}
    \caption{Demo~I (Scenario~1): Qualitative trajectory comparison across baselines.}
    \label{fig:demo1_traj_s1}
\end{figure*}

\begin{figure*}[h!t]
    \centering
    \begin{subfigure}[b]{0.24\textwidth}
        \centering
        \includegraphics[width=\linewidth]{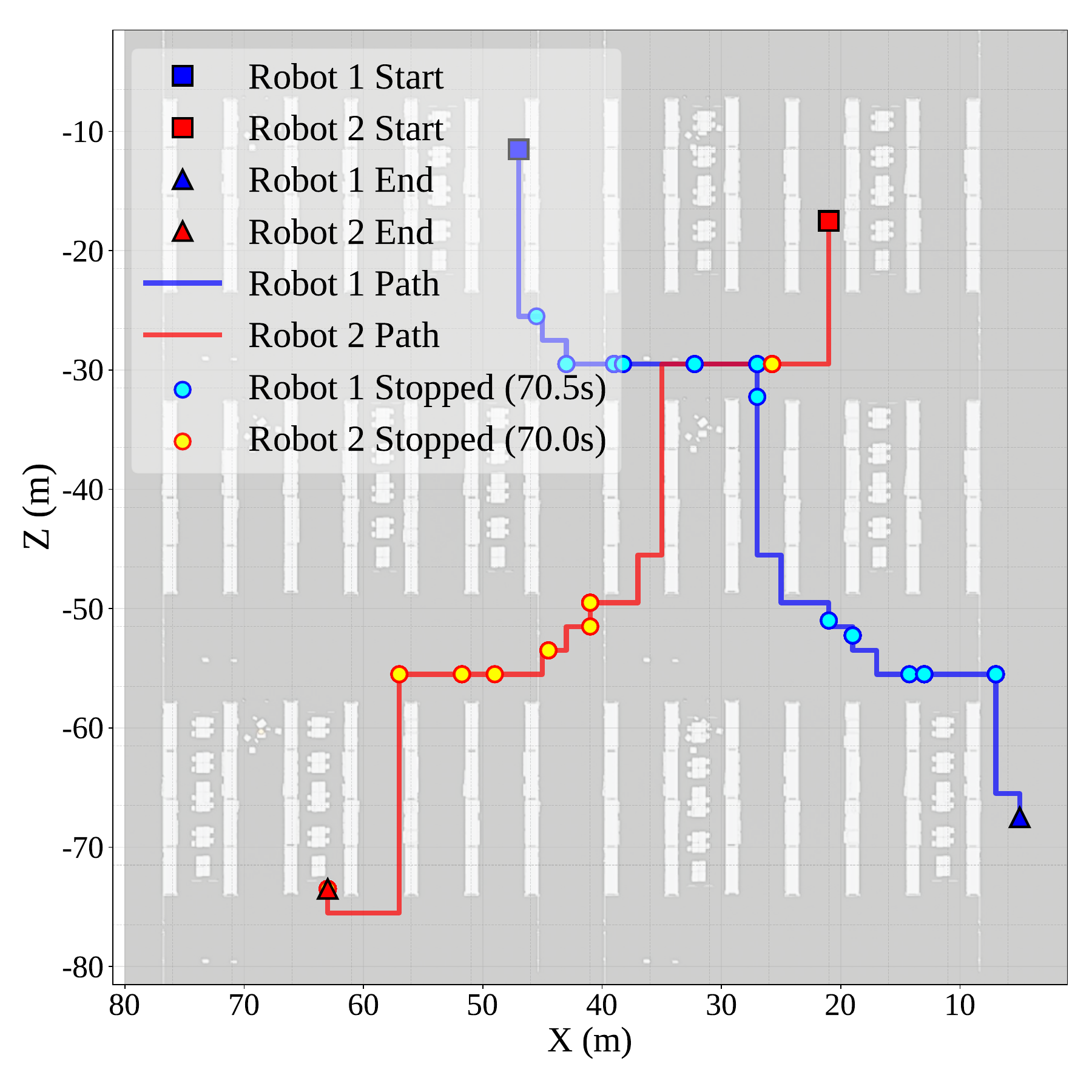}
        \caption{Stop-and-Go}
    \end{subfigure}
    \hfill
    \begin{subfigure}[b]{0.24\textwidth}
        \centering
        \includegraphics[width=\linewidth]{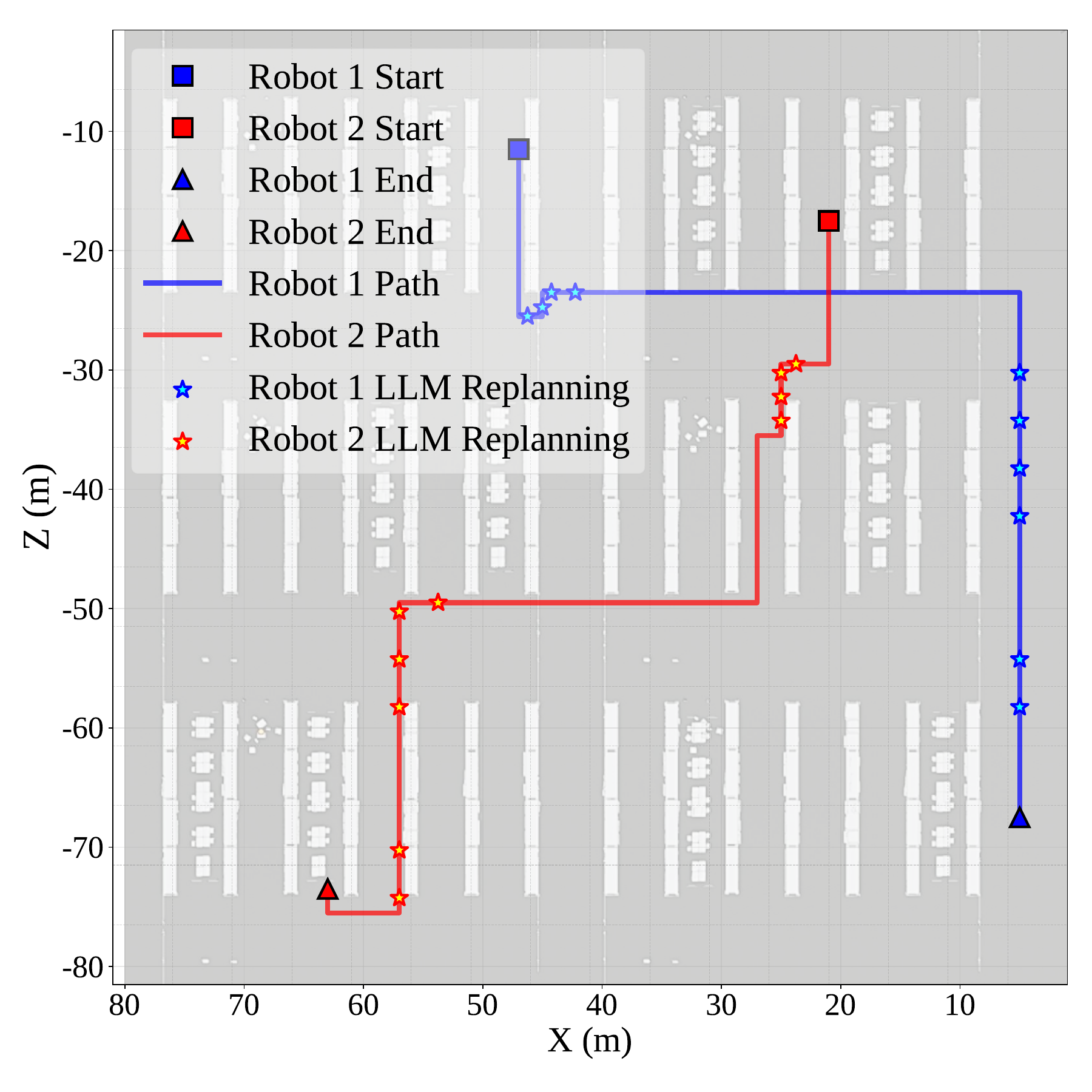}
        \caption{LORC-SC-P}
    \end{subfigure}
    \hfill
    \begin{subfigure}[b]{0.24\textwidth}
        \centering
        \includegraphics[width=\linewidth]{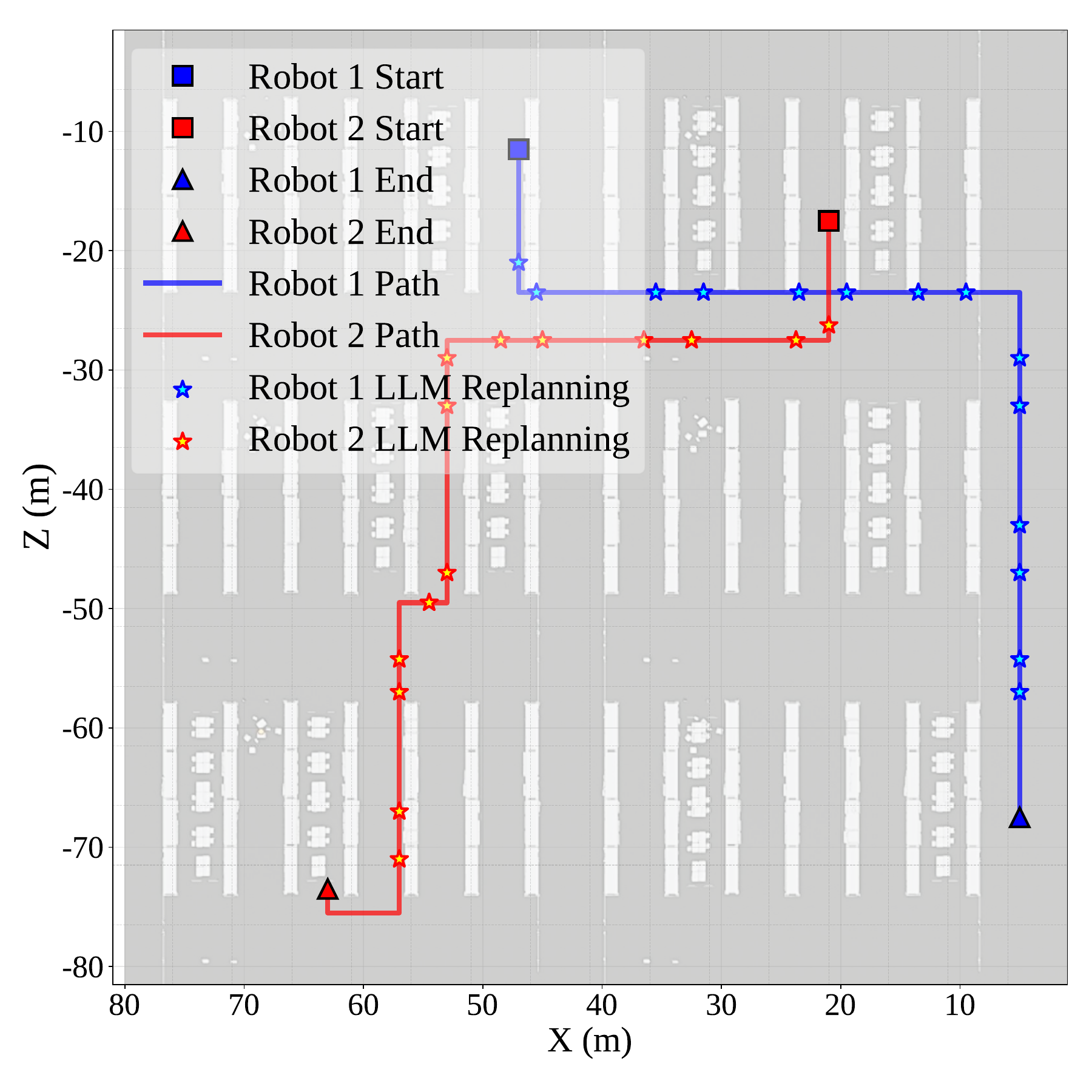}
        \caption{LORC-P}
    \end{subfigure}
    \hfill
    \begin{subfigure}[b]{0.24\textwidth}
        \centering
        \includegraphics[width=\linewidth]{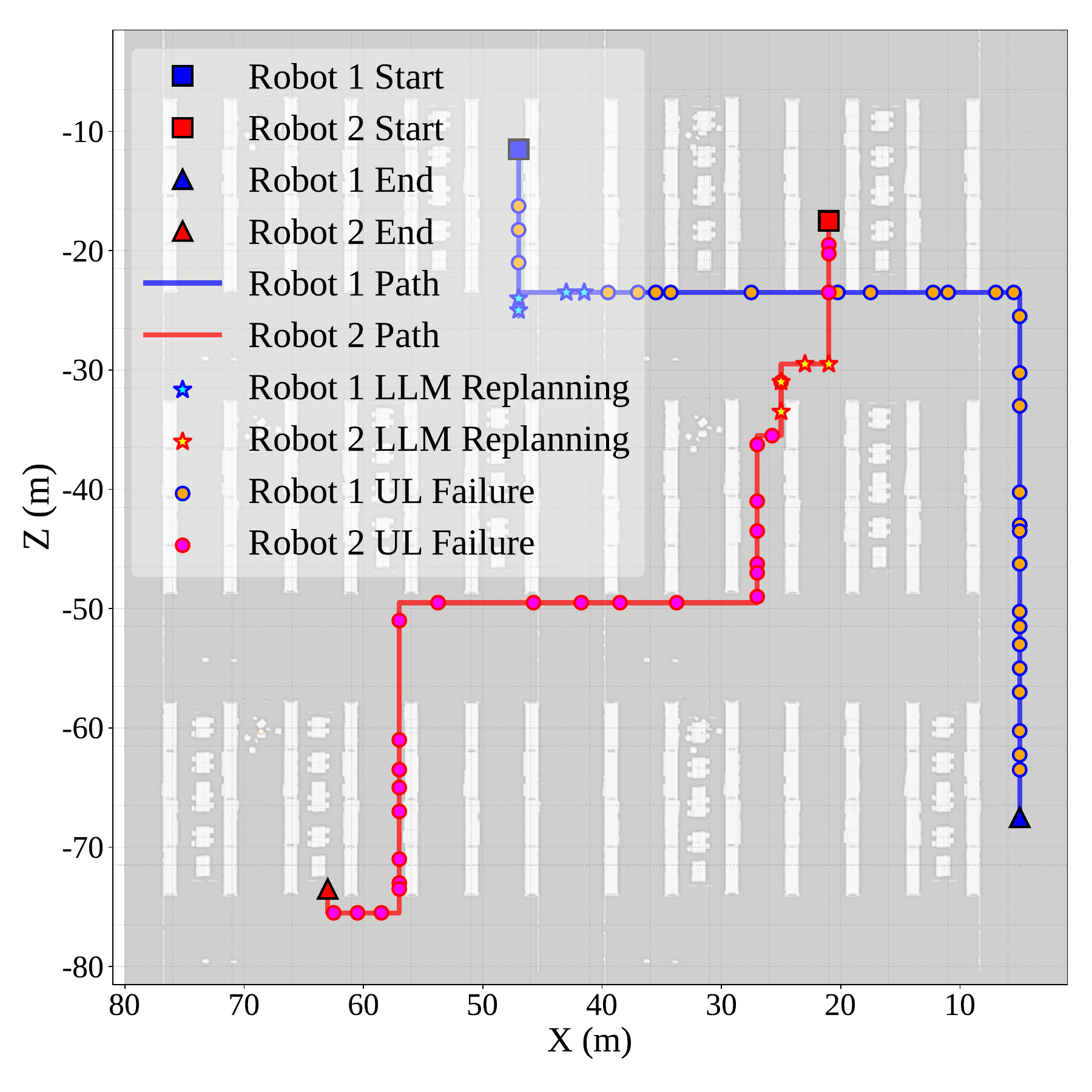}
        \caption{LORC-SC}
    \end{subfigure}
    \caption{Demo~I (Scenario~2): Qualitative trajectory comparison across baselines.}
    \label{fig:demo1_traj_s2}
\end{figure*}

\begin{figure*}[h!t]
    \centering
    \begin{subfigure}[b]{0.24\textwidth}
        \centering
        \includegraphics[width=\linewidth]{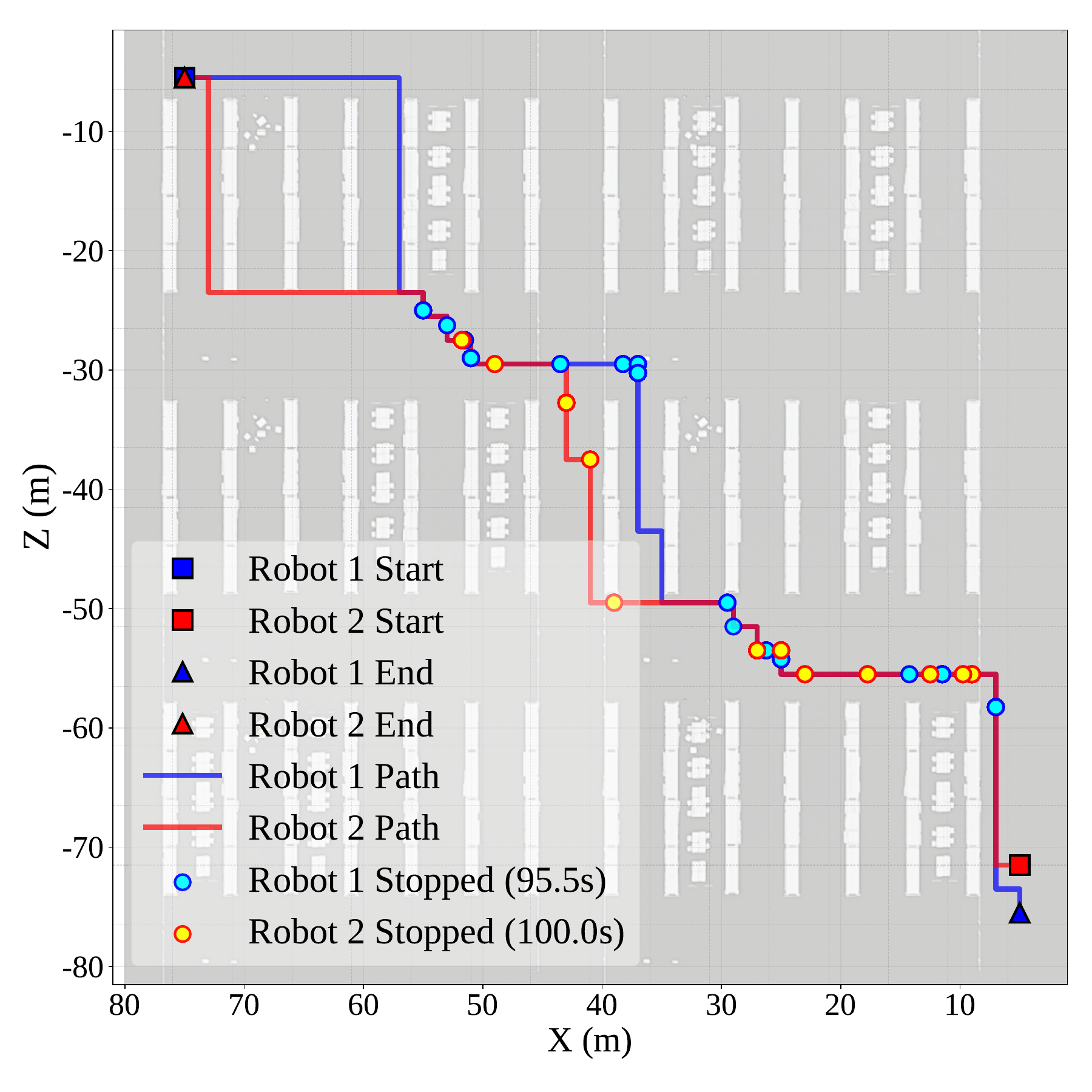}
        \caption{Stop-and-Go}
    \end{subfigure}
    \hfill
    \begin{subfigure}[b]{0.24\textwidth}
        \centering
        \includegraphics[width=\linewidth]{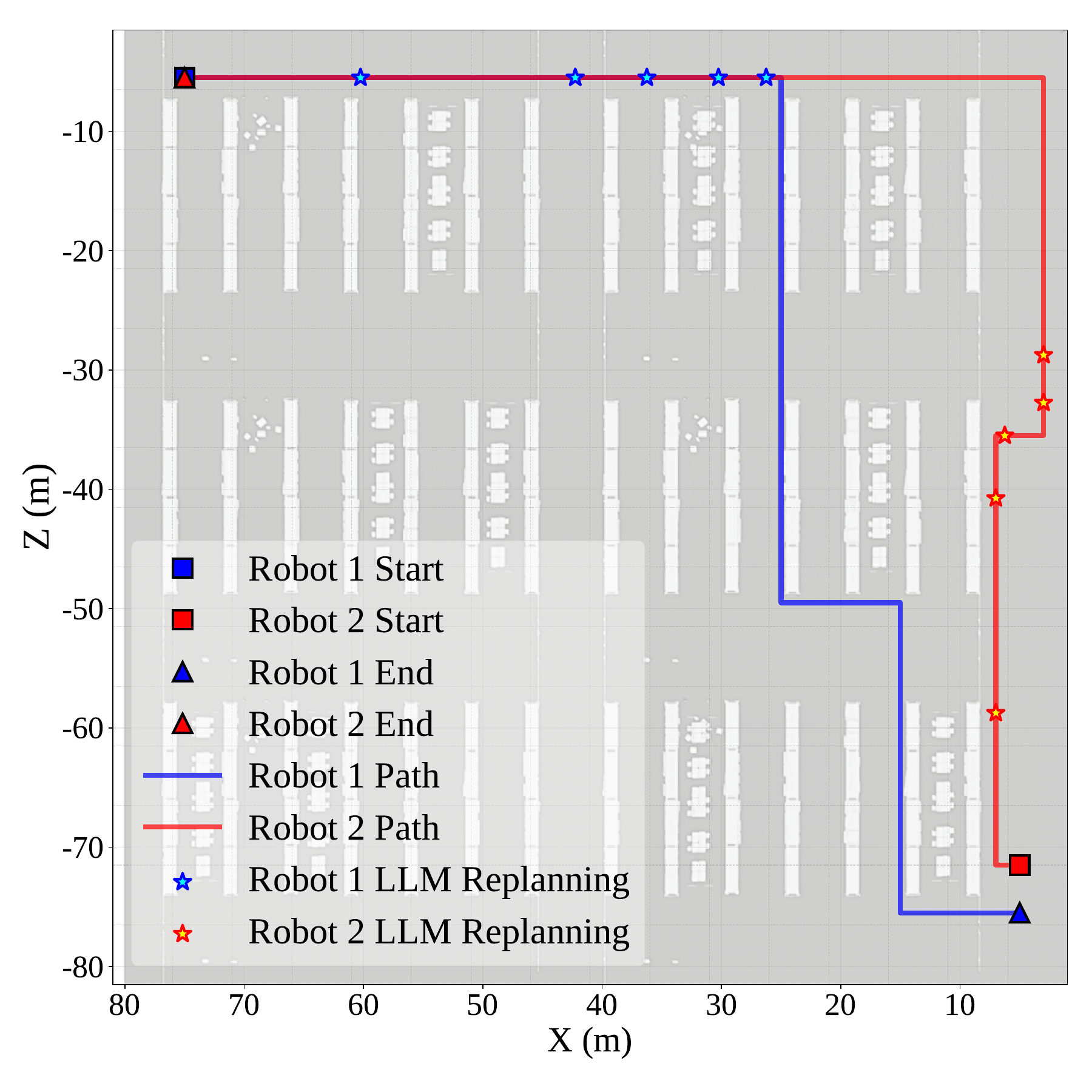}
        \caption{LORC-SC-P}
    \end{subfigure}
    \hfill
    \begin{subfigure}[b]{0.24\textwidth}
        \centering
        \includegraphics[width=\linewidth]{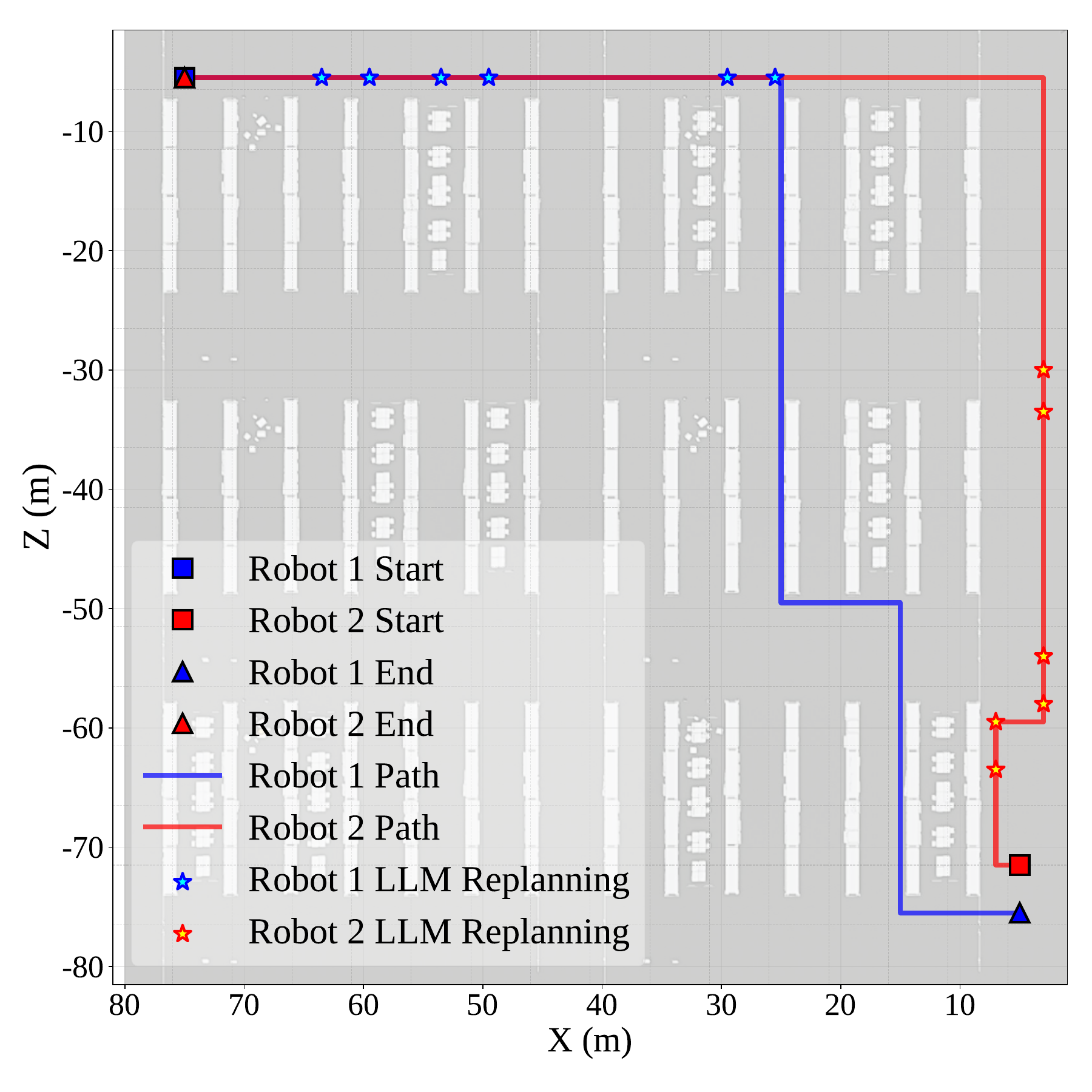}
        \caption{LORC-P}
    \end{subfigure}
    \hfill
    \begin{subfigure}[b]{0.24\textwidth}
        \centering
        \includegraphics[width=\linewidth]{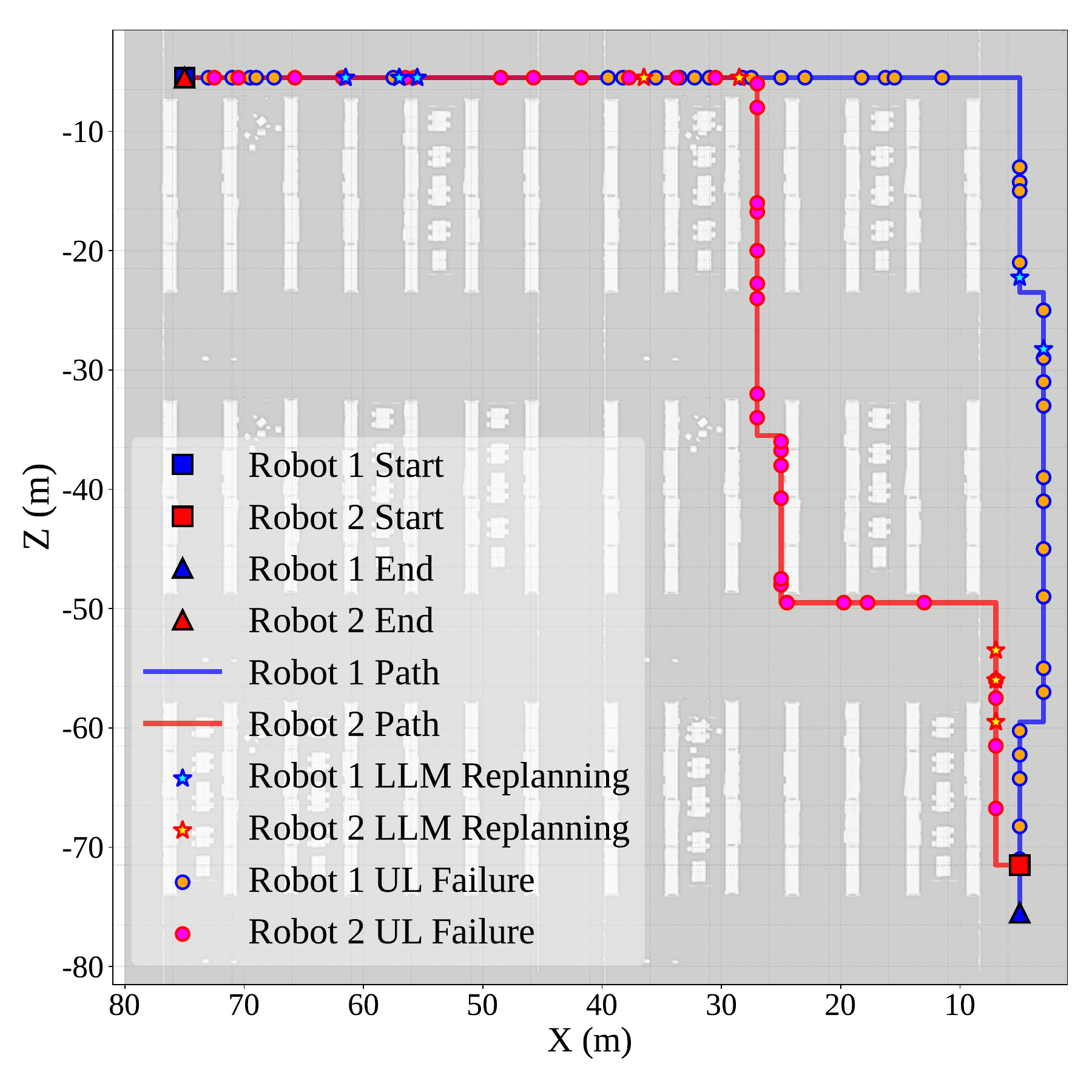}
        \caption{LORC-SC}
    \end{subfigure}
    \caption{Demo~I (Scenario~3): Qualitative trajectory comparison across baselines.}
    \label{fig:demo1_traj_s3}
\end{figure*}

We further visualize representative trajectories for Scenarios~1--3 in Figs.~\ref{fig:demo1_traj_s1}--\ref{fig:demo1_traj_s3}.
These qualitative comparisons highlight how different orchestration choices lead to distinct behaviors (e.g., frequent halts,
waiting for delayed uplink, or uplink failures) versus continuous progress under LORC-SC-P.

\begin{figure*}[h!t]
    \centering
    \begin{subfigure}[b]{0.24\textwidth}
        \centering
        \includegraphics[width=\linewidth]{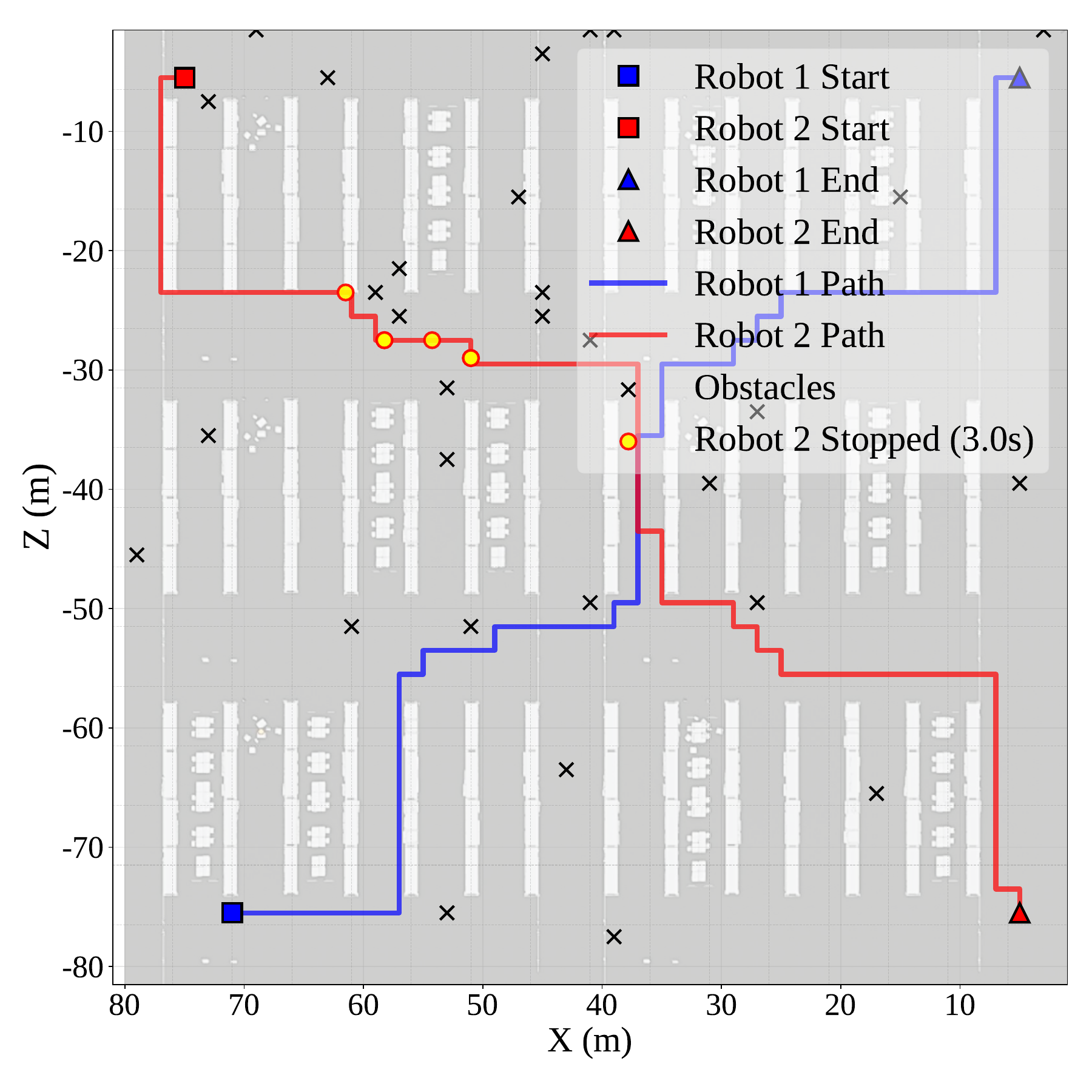}
        \caption{Stop-and-Go}
    \end{subfigure}
    \hfill
    \begin{subfigure}[b]{0.24\textwidth}
        \centering
        \includegraphics[width=\linewidth]{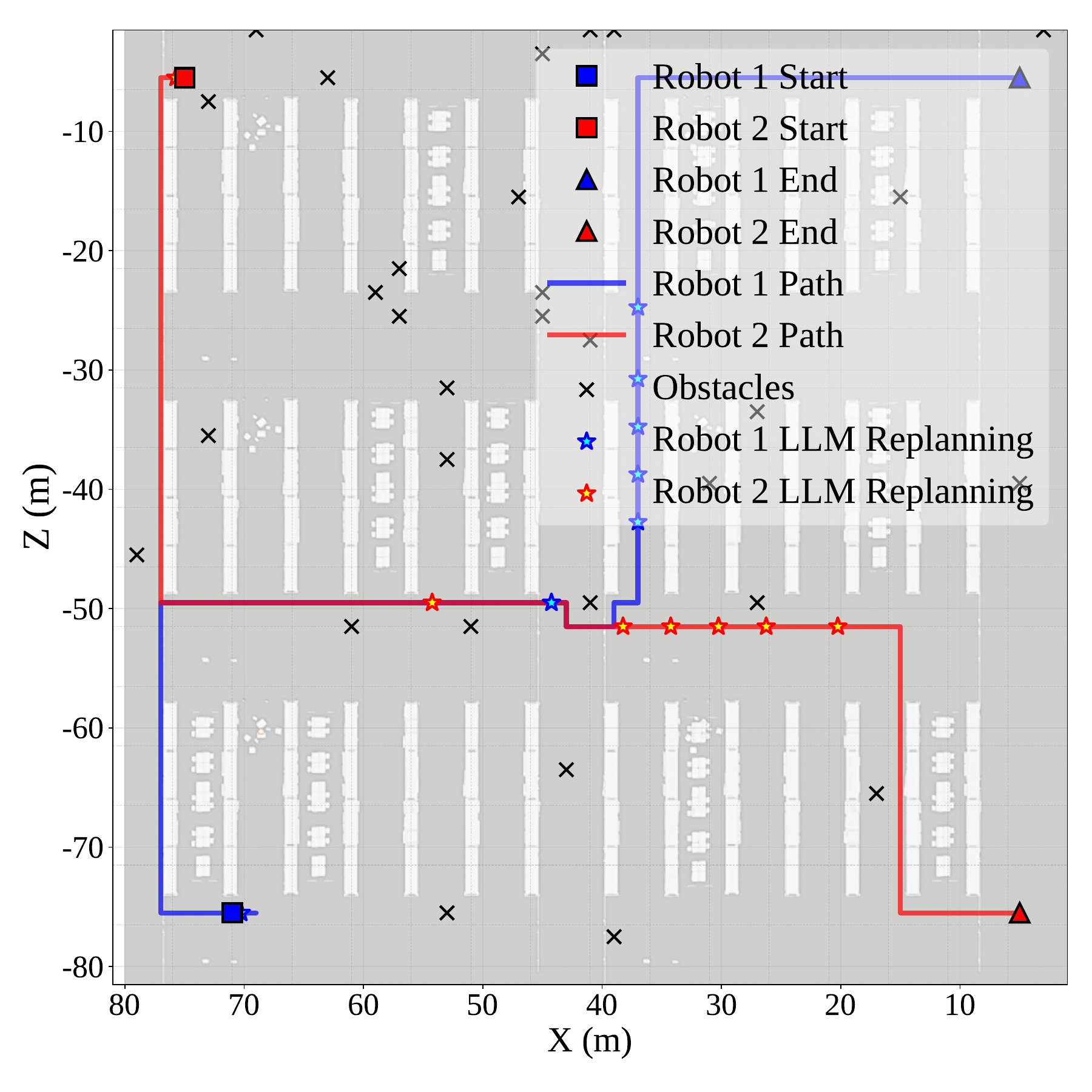}
        \caption{LORC-SC-P}
    \end{subfigure}
    \hfill
    \begin{subfigure}[b]{0.24\textwidth}
        \centering
        \includegraphics[width=\linewidth]{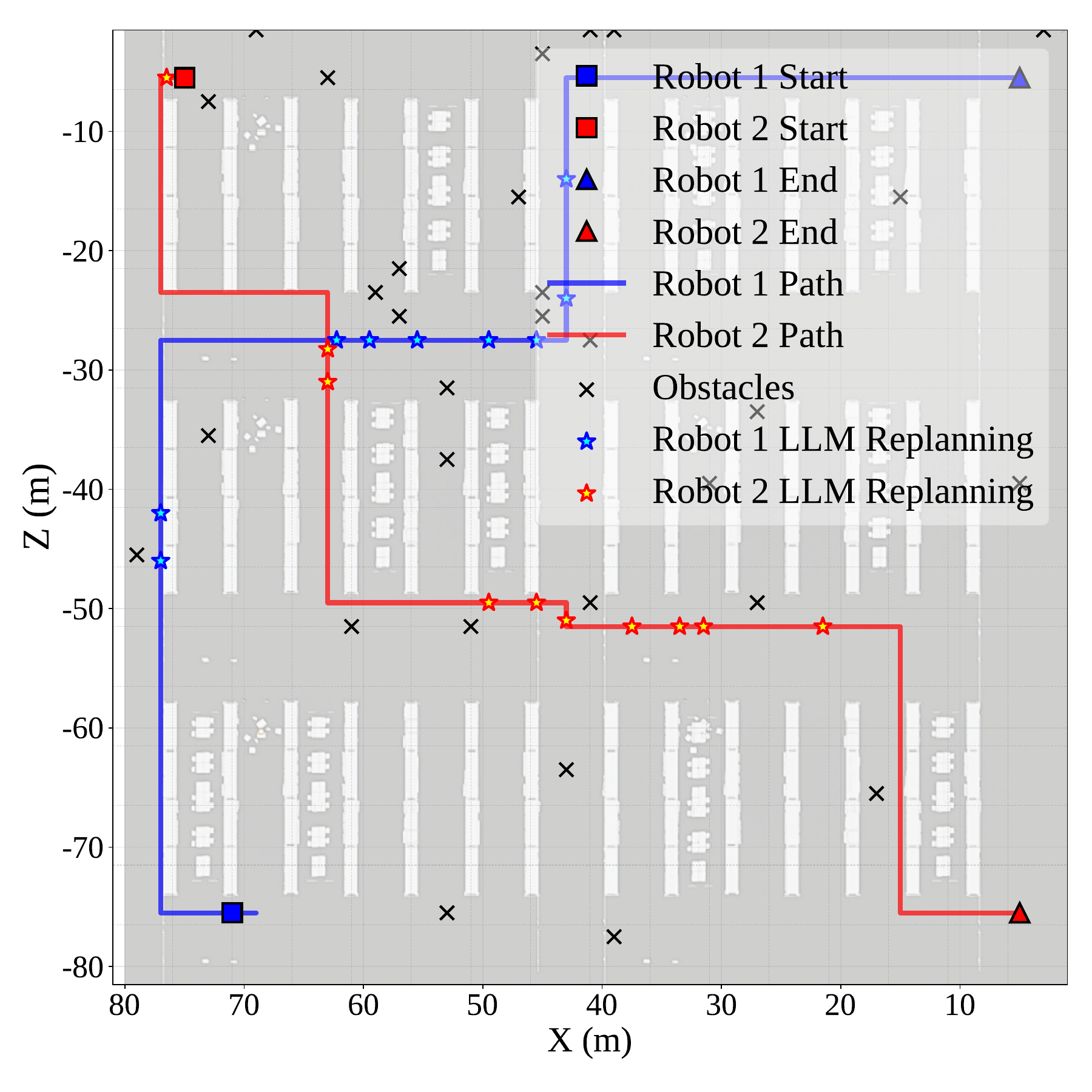}
        \caption{LORC-P}
    \end{subfigure}
    \hfill
    \begin{subfigure}[b]{0.24\textwidth}
        \centering
        \includegraphics[width=\linewidth]{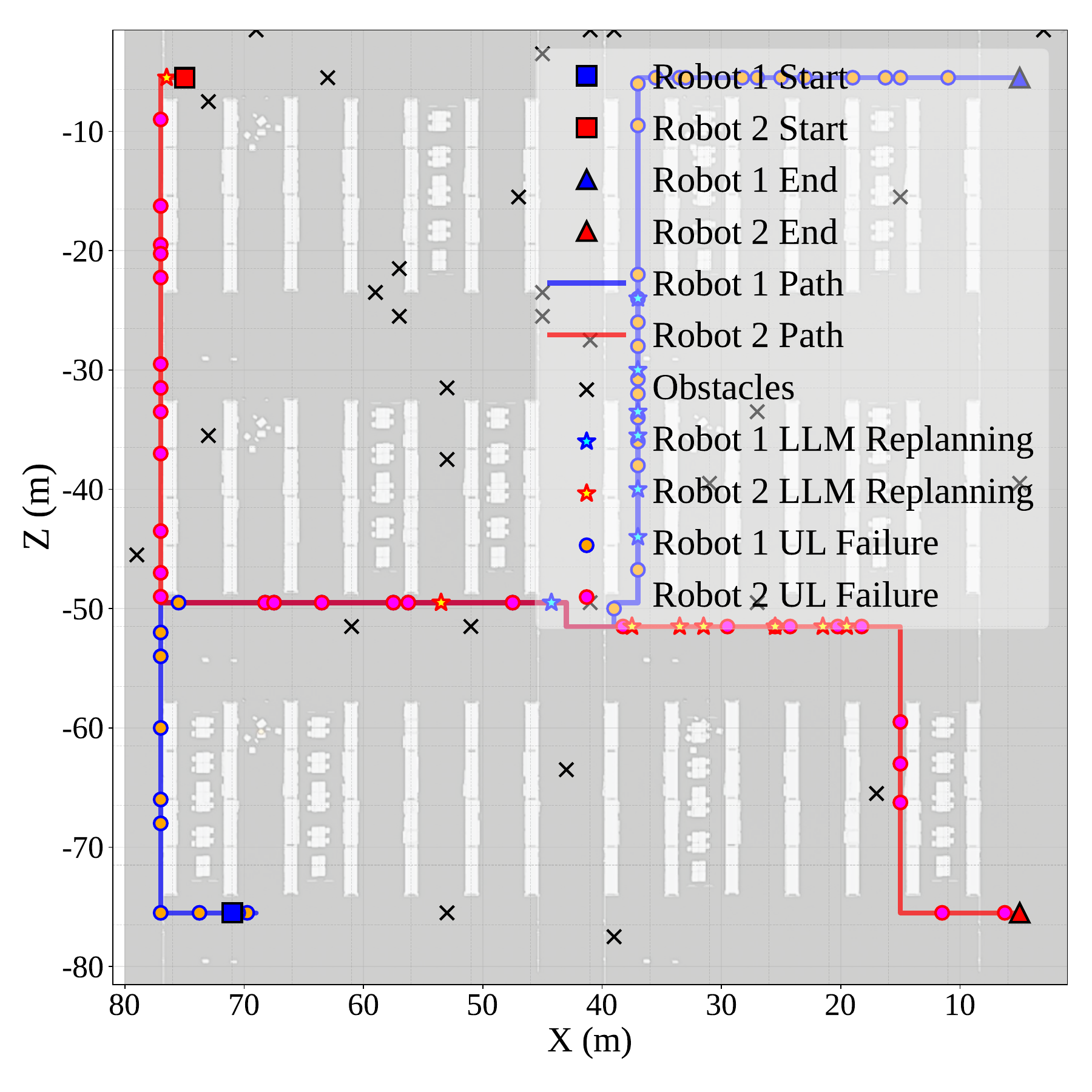}
        \caption{LORC-SC}
    \end{subfigure}
    \caption{Demo~I (Scenario~4): Qualitative trajectory comparison across baselines. The orchestrated design (LORC-SC-P) reduces waiting/halts by combining semantic sensing with predictive link-aware control.}
    \label{fig:demo1_traj_s4}
\end{figure*}

\subsubsection{Limitations and future research directions}
\label{sec:demo1_limits_future}
Demo~I relies on a high-fidelity digital twin and ray-tracing-derived link context, which may provide a level of channel awareness that is more idealized than what is achievable in real deployments with model mismatch, localization errors, and unmodeled dynamics.
In addition, mobility is discretized on a \(2\)~m grid and evaluated with a limited number of robots and humans, which may under-represent dense interactions, network contention, and edge-compute saturation.

A natural next step is twin-to-real calibration that continuously updates radio/material parameters and learns residual corrections so that link-context prediction remains robust under environment changes.
Another direction is scalable multi-robot orchestration that integrates uplink contention management, compute-aware replanning, and uncertainty-aware safety constraints with conservative fallback policies.

\subsection{Mobility Demo II: \tcb{MLLM}-Driven Proactive MCS Control}
\label{sec:demo2_mcs}

\subsubsection{Demo objective and orchestration knobs}
Demo~II focuses on \emph{proactive link adaptation} under mobility, where the channel quality varies rapidly and the
feedback available at the base station can be \emph{stale} due to sensing and processing delays.
The objective is to stabilize \textbf{throughput}, \textbf{latency}, and \textbf{BLER} for real-time robotic networking by
configuring the modulation-and-coding scheme (MCS) \emph{ahead of time}, rather than reacting to outdated CSI.
This demo can be initiated by a high-level operator intent expressed in text (e.g., ``prioritize reliability (BLER$\le 0.1$)'' vs.\ ``prioritize latency'' under mobility).
In our evaluation, we use a fixed intent---``maintain BLER$\le 0.1$ while reducing latency''---and report how the orchestrator translates this intent into proactive MCS decisions under delayed feedback.

\begin{figure*}[t]
    \centering
    \includegraphics[width=0.92\textwidth]{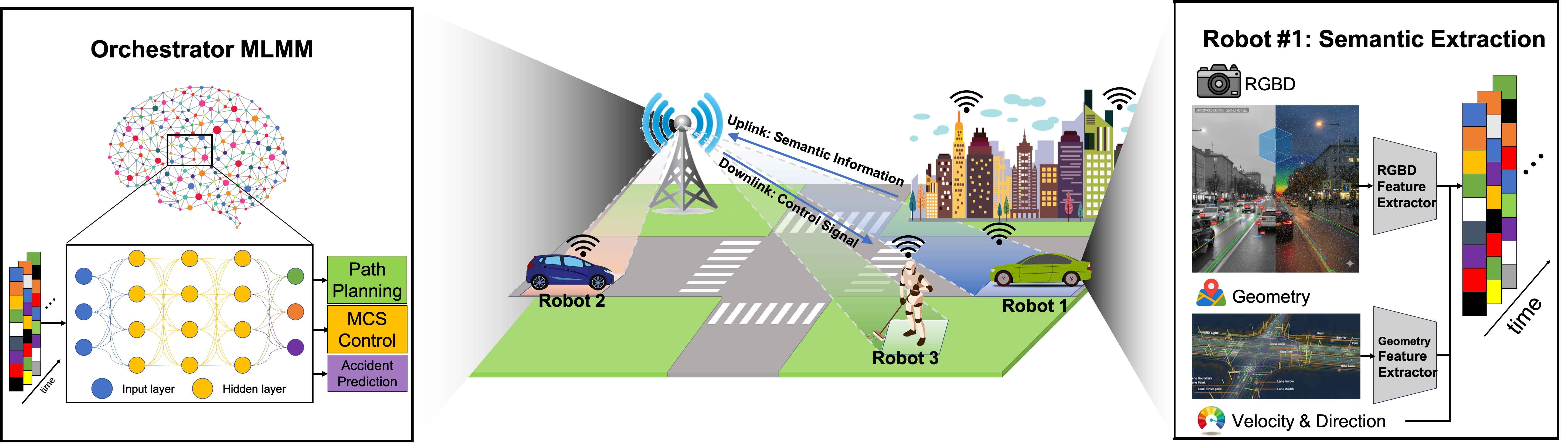}
\caption{End-to-end architecture of Demo~II (mobility scenario): local agents transmit compact semantic features; the \tcb{Orchestrator MLLM} predicts near-future link states and proactively configures MCS (and other network/robot control signals) to stabilize reliability and latency under mobility.}
    \label{fig:demo2_system}
\end{figure*}

Figure~\ref{fig:demo2_system} summarizes the end-to-end pipeline.
A centralized \tcb{\textbf{Orchestrator MLLM}} is co-located with the base station and receives \emph{compact semantic features}
delivered by mobile agents (e.g., autonomous vehicles and humanoid robots). \tcb{Because the orchestrator fuses multimodal inputs—RGBD-derived visual cues, geometric context, and motion embeddings—alongside channel feedback, it constitutes a multimodal model (MLLM) rather than a text-only LLM.} Using these semantics as context, the orchestrator
predicts short-horizon channel variations and selects an MCS that (i) maintains reliability (target BLER \(\le 0.1\)) and
(ii) maximizes effective rate, thereby suppressing retransmissions and latency spikes.

\begin{tcolorbox}[colback=gray!5!white,
                  colframe=gray!55!black,
                  title=What is orchestrated in Demo~II (and why it is measurable),
                  fonttitle=\bfseries,
                  colbacktitle=gray!75!black,
                  coltitle=white,
                  breakable,
                  enhanced,
                  left=2mm, right=2mm, top=1mm, bottom=1mm]
\begin{itemize}
    \item \textbf{Sensing:} instead of raw RGB/LiDAR streaming, each agent transmits \emph{semantic features}
    (RGBD cues, geometry, velocity/direction), reducing payload while preserving the context needed for link prediction.
    \item \textbf{Communication:} the orchestrator performs \emph{predictive} MCS selection to satisfy a reliability target (BLER \(\le 0.1\)) while improving throughput and reducing latency caused by HARQ retransmissions.
        \item \textbf{Computation:} a centralized \tcb{MLLM} fuses multimodal semantic context and delayed link feedback to anticipate
    short-term channel variations, mitigating the mismatch inherent to reactive link adaptation.

\end{itemize}
\end{tcolorbox}

\subsubsection{Simulation testbed and channel-context construction}
\label{sec:demo2_testbed}
We emulate a complex urban mobility scenario in which autonomous vehicles and humanoid robots coexist.
To model the wireless channel with geometry-aware blockage and multipath effects, we use the open-source ray-tracing library
\textit{Sionna~1.2.1} to generate a \emph{path-loss heatmap} over the region of interest.
This heatmap provides an interpretable \emph{link-context signal} that reflects location-dependent attenuation due to urban obstacles.

\begin{figure*}[t]
    \centering
    \includegraphics[width=0.92\textwidth]{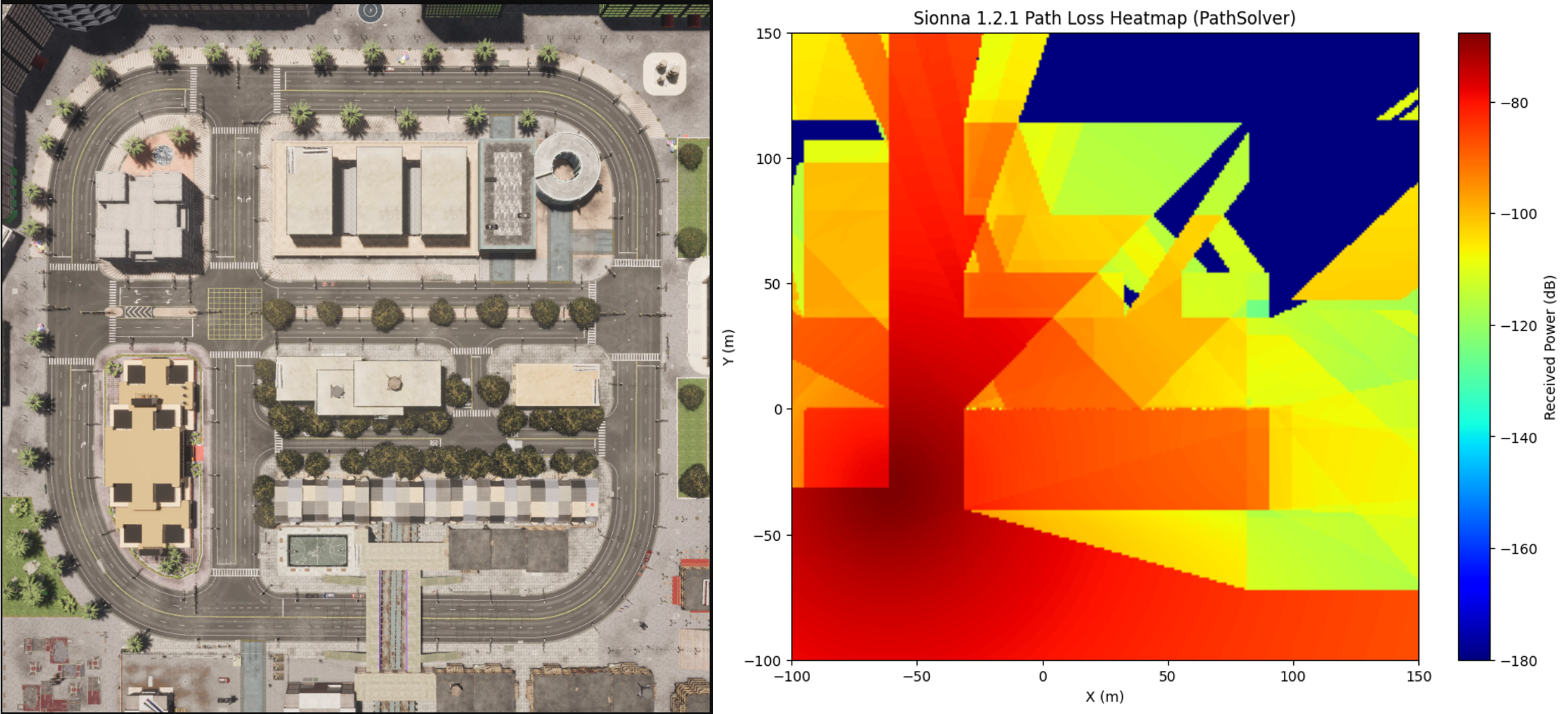}
    \caption{Demo~II testbed snapshots: urban simulation environment (left) and Sionna~1.2.1-based path-loss heatmap (right).}
    \label{fig:demo2_pathloss}
\end{figure*}

Figure~\ref{fig:demo2_pathloss} shows (left) the urban simulation environment and (right) the corresponding Sionna-based
path-loss heatmap used for communication evaluation.

\subsubsection{Semantic sensing for proactive link adaptation}
\label{sec:demo2_semantic}
Each local agent perceives the surrounding scene using onboard RGB cameras and LiDAR sensors.
Rather than transmitting raw sensor streams, the agent extracts compact semantic features that summarize
(i) RGBD-related visual cues, (ii) geometric structures, and (iii) motion context such as object velocity and direction.
These semantic representations are delivered to the base station, enabling the \tcb{Orchestrator MLLM} to reason about \emph{how mobility and geometry} 
may impact the near-future channel state (e.g., entering/exiting blockage regions), while keeping the uplink payload modest.

\subsubsection{\tcb{Orchestrator MLLM} for predictive MCS control}
\label{sec:demo2_orchestrator}
The \tcb{Orchestrator MLLM} delivers \textbf{predictive link adaptation} under delayed feedback.
Let \(\gamma(t)\) be the (instantaneous) SNR at time \(t\), and we assume that the available feedback satisfies the condition such that a reactive baseline uses
\(\gamma(t-d)\) for MCS selection with delay \(d\).
Under mobility, this mismatch can lead to either (i) overly aggressive MCS (high BLER \(\rightarrow\) retransmissions \(\rightarrow\) latency spikes)
or (ii) overly conservative MCS (lower throughput).

The proposed \tcb{MLLM-driven} policy uses transmitted semantic context to estimate a near-future SNR \(\widehat{\gamma}(t)\) (or an equivalent short-horizon link-quality proxy) and selects an MCS level that balances \emph{rate} and \emph{reliability}.
Specifically, MCS selection follows a reliability-constrained rule aligned with the target BLER requirement:
\begin{equation}
m^\star(t) \in \arg\max_{m \in \mathcal{M}} \; R(m)
\quad \text{s.t.} \quad \mathrm{BLER}\!\left(m,\widehat{\gamma}(t)\right) \le 0.1,
\label{eq:demo2_mcs_select}
\end{equation}
where \(\mathcal{M}\) is the discrete MCS set and \(R(m)\) is the nominal rate associated with \(m\).
This predictive selection is evaluated against ideal and delayed-feedback baselines in the next subsection.

\subsubsection{Baselines and evaluation protocol}
\label{sec:demo2_baselines}
We evaluate three techniques using three metrics: \textbf{BLER}, \textbf{throughput}, and \textbf{latency}.
The key reliability target is \(\mathrm{BLER} \le 0.1\), consistent with the dashed reference used in the plots.

\begin{itemize}
    \item \textbf{Oracle (upper bound):} assumes perfect knowledge of future channel states (theoretical upper bound).
    \item \textbf{Ideal MCS policy:} uses the \emph{actual} channel information but selects the highest MCS satisfying
    \(\mathrm{BLER} \le 0.1\) using instantaneous but non-predictive channel knowledge. 
    \item \textbf{Reactive baselines (t--1 to t--30):} select MCS using outdated channel measurements, modeling feedback delays
    induced by mobility and sensing/processing latency.
\end{itemize}

\subsubsection{Key results: temporal stability and SNR-dependent performance}
\label{sec:demo2_results}
\begin{figure*}[t]
    \centering
    \begin{subfigure}[b]{0.32\textwidth}
        \centering
        \includegraphics[width=\linewidth]{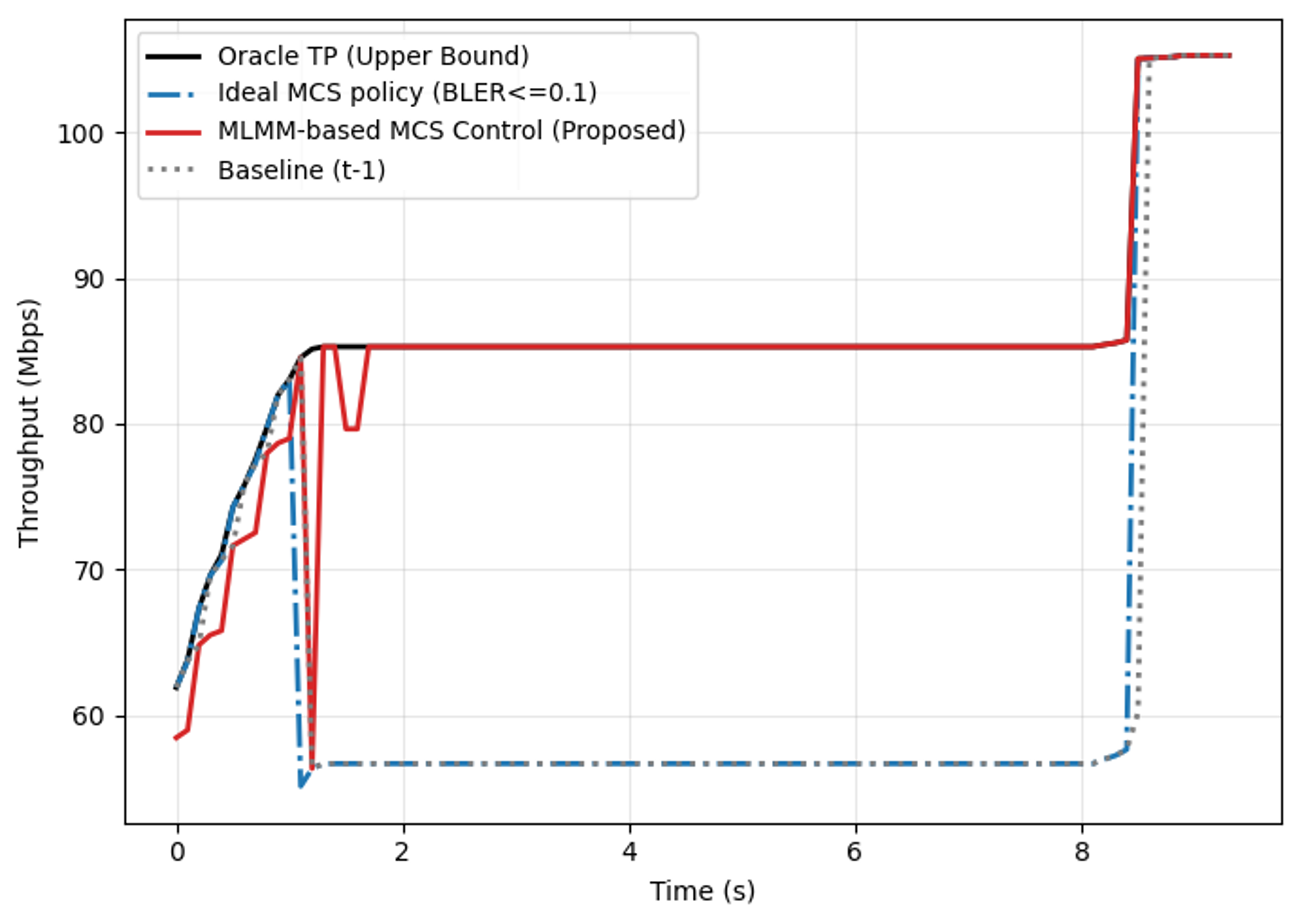}
        \caption{Throughput over time}
        \label{fig:demo2_tp_time}
    \end{subfigure}
    \hfill
    \begin{subfigure}[b]{0.32\textwidth}
        \centering
        \includegraphics[width=\linewidth]{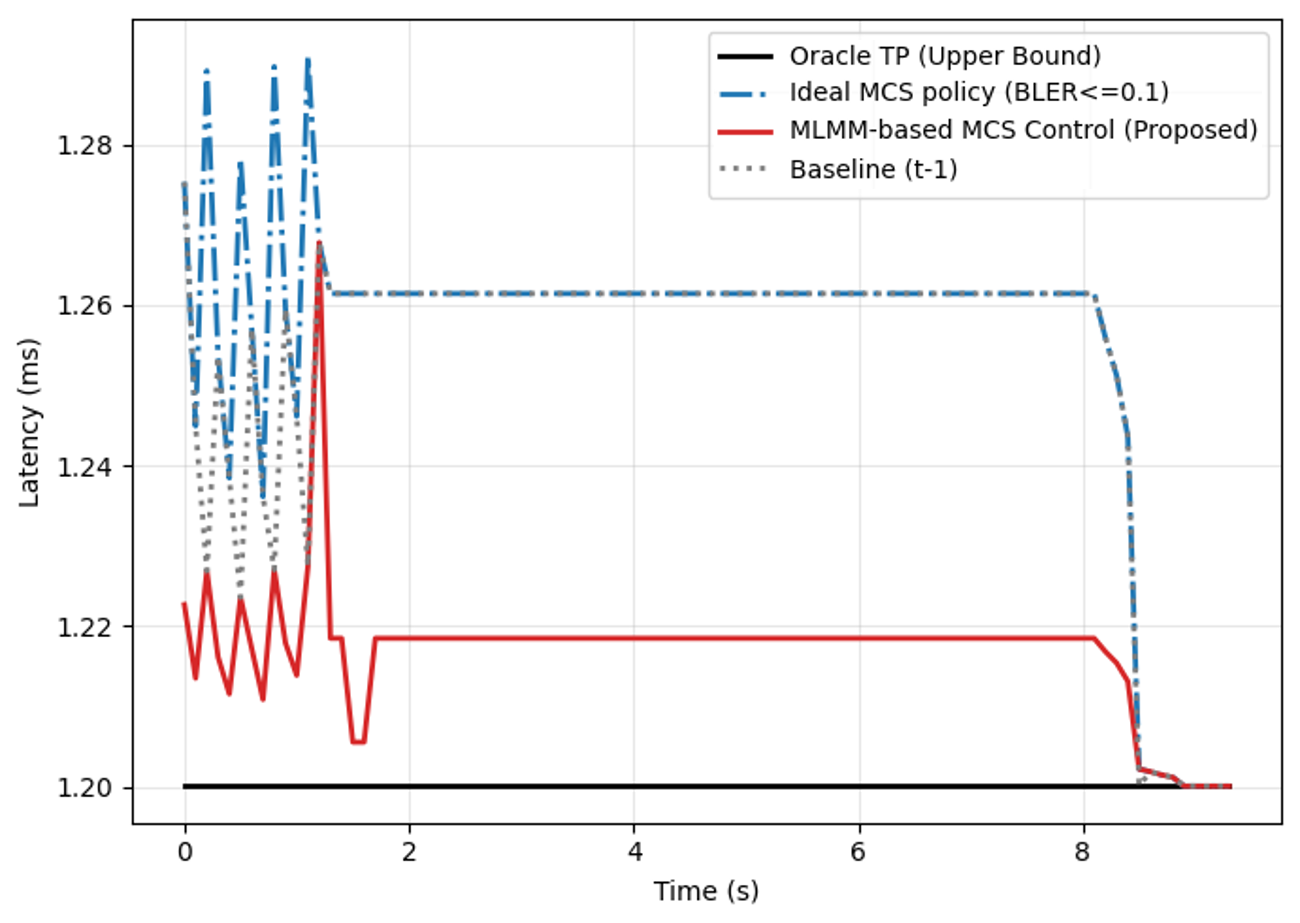}
        \caption{Latency over time}
        \label{fig:demo2_lat_time}
    \end{subfigure}
    \hfill
    \begin{subfigure}[b]{0.32\textwidth}
        \centering
        \includegraphics[width=\linewidth]{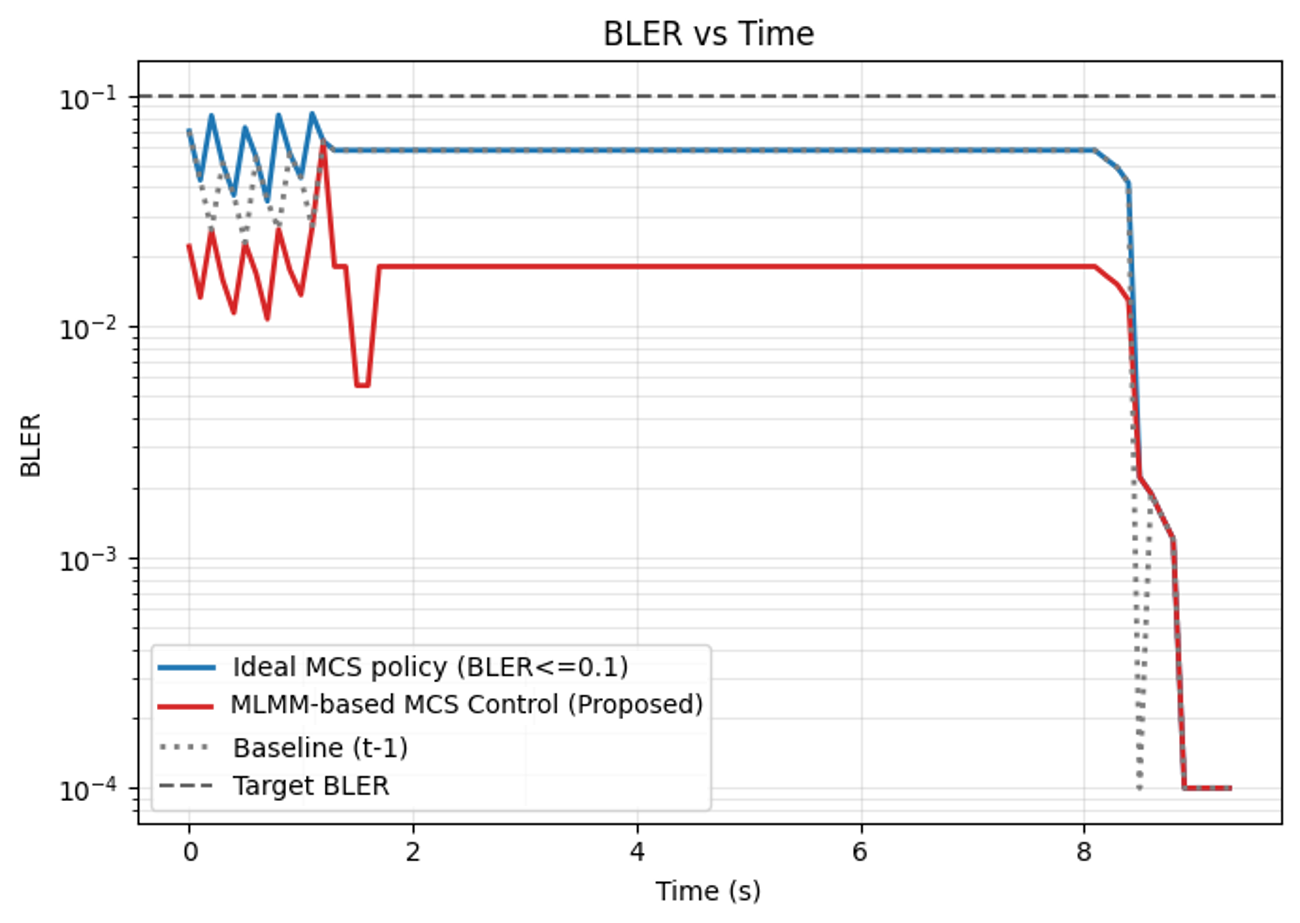}
        \caption{BLER over time}
        \label{fig:demo2_bler_time}
    \end{subfigure}
    \caption{Demo~II temporal performance: (a) throughput, (b) latency, and (c) BLER. The proposed \tcb{MLLM-driven} MCS control
    tracks near-ideal behavior and mitigates BLER spikes caused by delayed feedback.}

    \label{fig:demo2_temporal}
\end{figure*}

Figure~\ref{fig:demo2_temporal} compares temporal trajectories of throughput, latency, and BLER.
The proposed \tcb{MLLM-driven} controller closely follows the Oracle/Ideal policies over time, while the reactive baseline (e.g., t--1)
exhibits degraded stability when the channel changes abruptly (BLER spikes and corresponding latency penalties).

\begin{figure*}[t]
    \centering
    \begin{subfigure}[b]{0.32\textwidth}
        \centering
        \includegraphics[width=\linewidth]{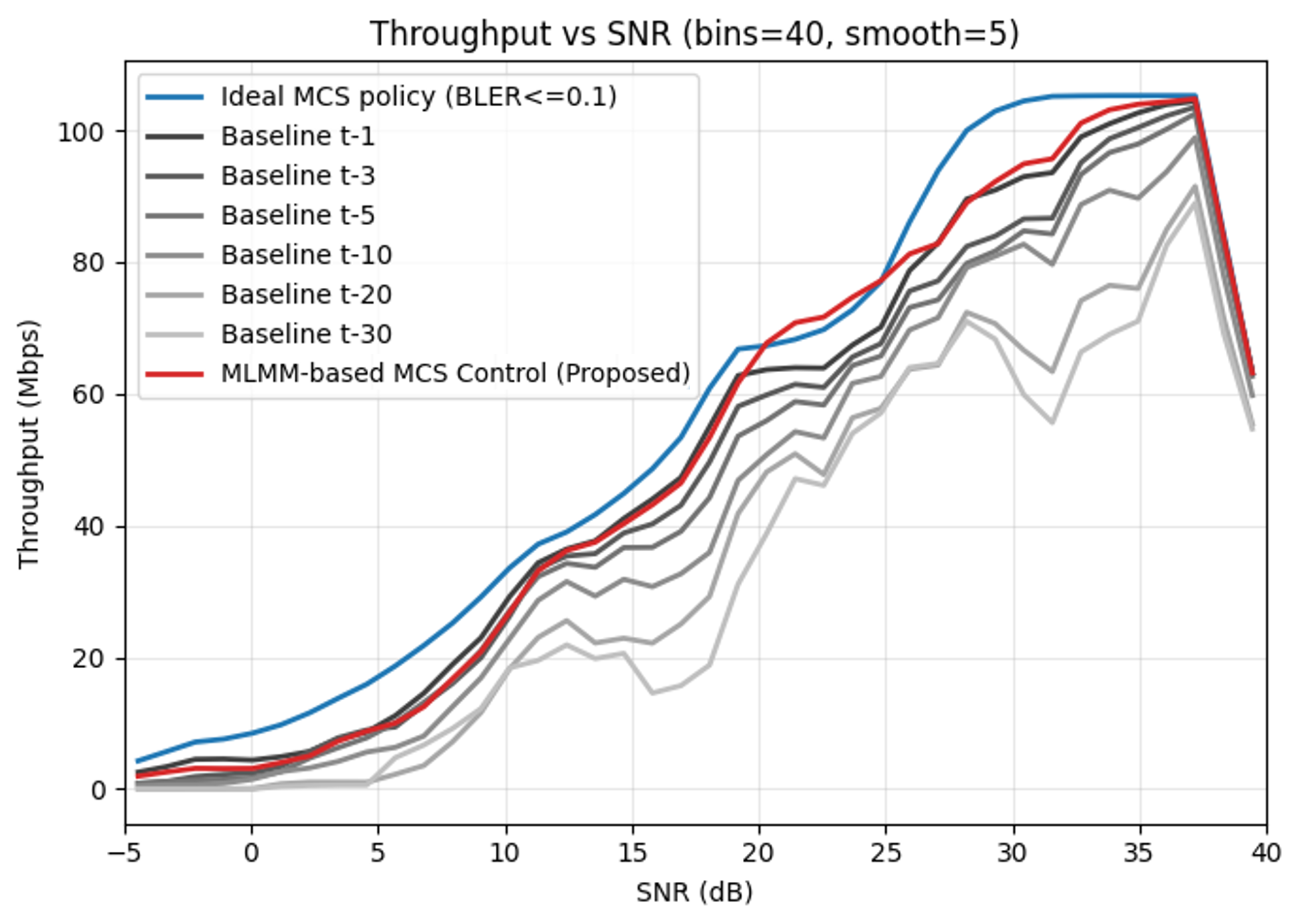}
        \caption{Throughput vs.\ SNR}
        \label{fig:demo2_tp_snr}
    \end{subfigure}
    \hfill
    \begin{subfigure}[b]{0.32\textwidth}
        \centering
        \includegraphics[width=\linewidth]{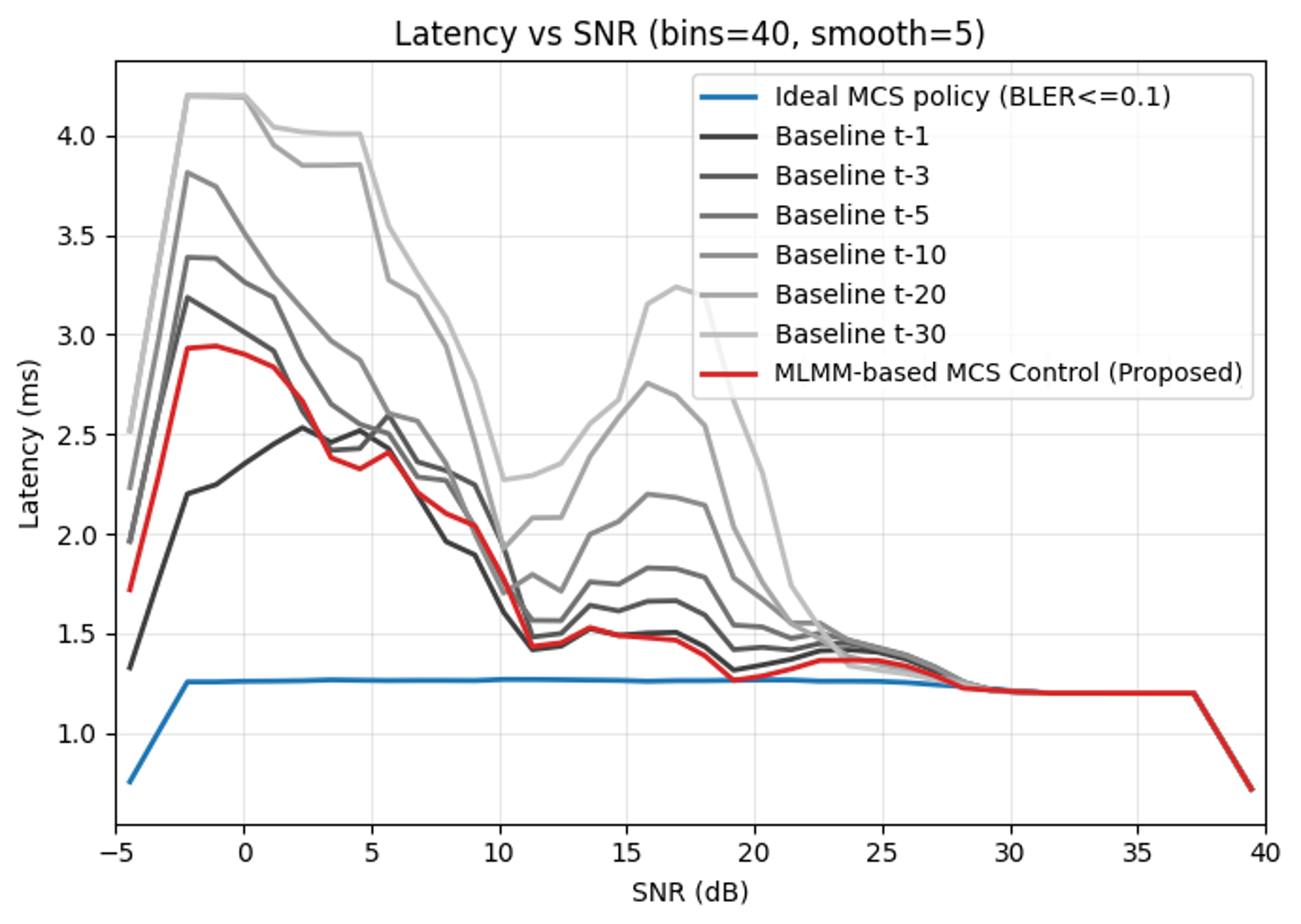}
        \caption{Latency vs.\ SNR}
        \label{fig:demo2_lat_snr}
    \end{subfigure}
    \hfill
    \begin{subfigure}[b]{0.32\textwidth}
        \centering
        \includegraphics[width=\linewidth]{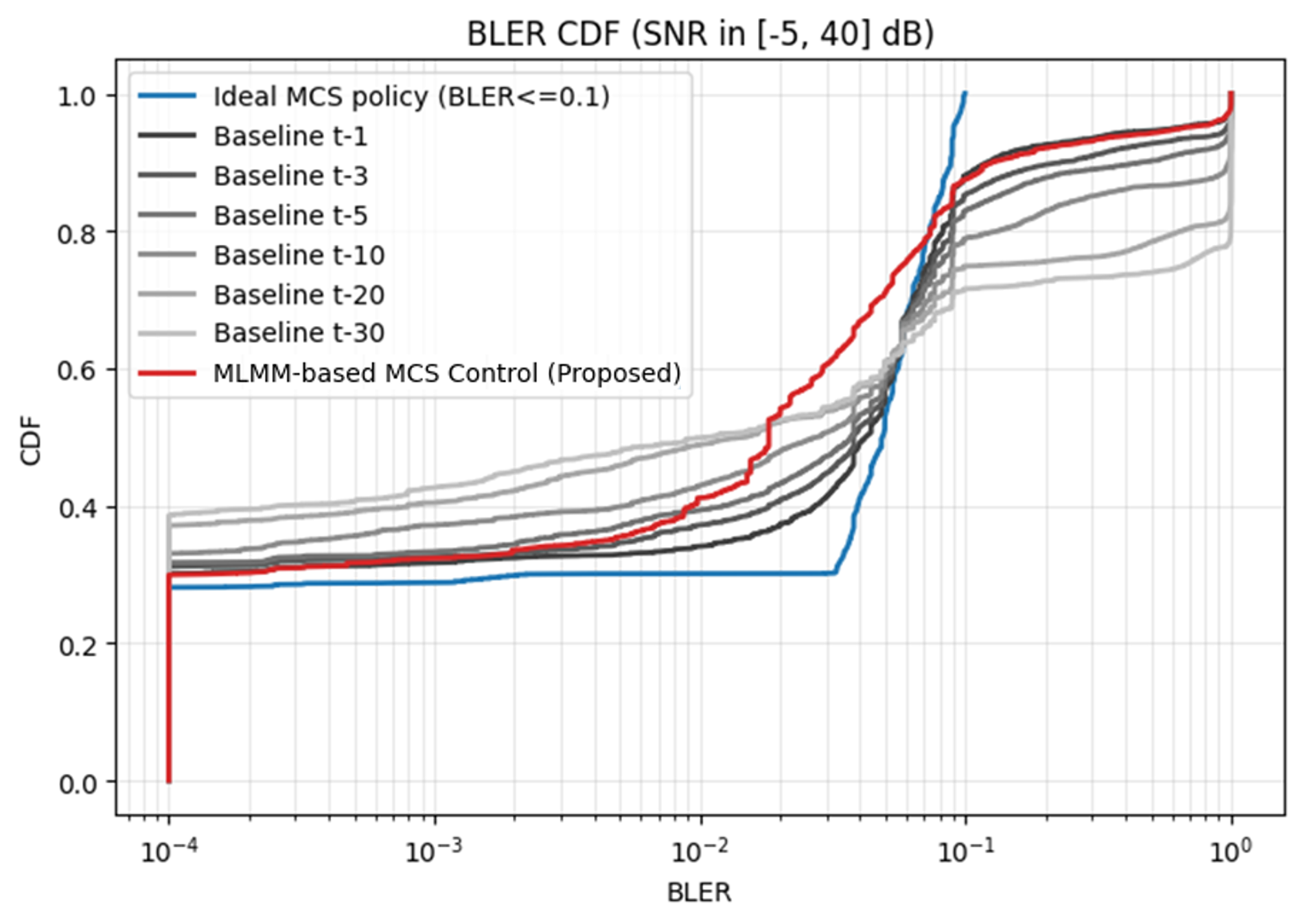}
        \caption{BLER distribution (CDF) over SNR range}
        \label{fig:demo2_bler_snr}
    \end{subfigure}
    \caption{Demo~II SNR-based performance: (a) throughput, (b) latency, and (c) BLER distribution.
    The proposed \tcb{MLLM-driven} controller consistently outperforms delayed-feedback baselines.}

    \label{fig:demo2_snr}
\end{figure*}

We further report SNR-conditioned statistics in Fig.~\ref{fig:demo2_snr}.
Across whole SNR regimes, the proposed technique outperforms delayed-feedback baselines (t--1 to t--30) in terms of throughput and latency. In addition, the BLER distribution is concentrated below the target BLER of 0.1, indicating that predictive MCS selection reduces reliability violations and therefore minimizes the packet retransmissions.

\subsubsection{Quantitative performance gains under delayed feedback}
\label{sec:demo2_gain}
Table~\ref{tab:demo2_gain} summarizes aggregate gains of the proposed method over delayed-feedback baselines.
We compute (i) throughput gain and (ii) latency reduction as
\(\Delta\mathrm{TP}(\%) = 100\cdot(\mathrm{TP}_{\mathrm{prop}}-\mathrm{TP}_{\mathrm{base}})/\mathrm{TP}_{\mathrm{base}}\) and
\(\Delta\mathrm{Lat}(\%) = 100\cdot(\mathrm{Lat}_{\mathrm{base}}-\mathrm{Lat}_{\mathrm{prop}})/\mathrm{Lat}_{\mathrm{base}}\).
Positive values indicate improvements by the proposed intelligent orchestration; negative values indicate the baseline is better under that metric.

\begin{table*}[t]
\centering
\caption{Demo~II: Aggregate performance gains of the proposed \tcb{MLLM-driven} MCS control over baselines.}

\label{tab:demo2_gain}
\small
\renewcommand{\arraystretch}{1.15}
\setlength{\tabcolsep}{7pt}
\begin{tabular}{lcc}
\toprule
\textbf{Baseline} & \multicolumn{1}{c}{\textbf{TP gain (\%)}} & \multicolumn{1}{c}{\textbf{Latency reduction (\%)}} \\
\midrule \midrule
Ideal MCS policy (\(\mathrm{BLER}\le 0.1\)) & -14.31 & -39.95 \\
Baseline t--1  &  2.83 & -0.03 \\
Baseline t--3  & 14.14 & 11.62 \\
Baseline t--5  & 26.43 & 20.23 \\
Baseline t--10 & 44.77 & 29.15 \\
Baseline t--20 & 102.68 & 40.89 \\
Baseline t--30 & 100.91 & 39.88 \\
\bottomrule
\end{tabular}
\end{table*}

Overall, the results demonstrate that proactive, \tcb{MLLM-driven,} semantic-aware MCS control can substantially reduce the performance loss caused by delayed feedback under mobility.

\subsubsection{Limitations and future research directions}
\label{sec:demo2_limits_future}
Demo~II is simulation-driven and exposes MCS as the primary communication knob, while real stacks also include scheduler dynamics, buffering, HARQ behavior, and multi-user interference that can affect latency and reliability.
Moreover, the semantic context used for prediction may depend on the specific environment, and further validation is required to assess how well the learned policy generalizes across different geometries and sensing modalities.
Despite these limitations, this demo provides a clean isolation of the impact of predictive link adaptation under mobility, making the benefits of semantic-aware prediction directly measurable.

An important extension is joint PHY/MAC orchestration that goes beyond MCS-only control to include scheduling and queue-aware reliability management, directly optimizing tail latency under mobility. Another promising direction is risk-aware predictive control that outputs calibrated uncertainty (e.g., distributions of SNR/BLER) and selects MCS conservatively when abrupt blockage is predicted, together with domain generalization strategies that maintain performance across diverse environments and sensing configurations.

\subsection{Real-Hardware Demo III: FollowMe Robot with a Practical Semantic-Sensing Switch}
\label{sec:demo3_followme}

\subsubsection{Demo objective and closed-loop requirement}
While Demos~I--II validate R2X orchestration in controlled (digital-twin / simulated-channel) environments, real deployments must additionally handle \emph{non-ideal wireless stacks}, \emph{unpredictable interference}, and \emph{robot-control sensitivity to delay and jitter}.
In this demo, we implement a real FollowMe prototype in which a mobile robot follows a walking human in an indoor corridor.
We view FollowMe as an intent-driven service that can start from a short language command such as ``Follow me smoothly and safely'' (or ``prioritize low latency under poor WiFi'').
In this demo, we assume a fixed intent---stable tracking with low CTA tail latency---and the orchestrator realizes it by switching sensing representations (JPEG vs.\ VQ) and communication QoS based on measured WiFi conditions.

In particular, the robot must transmit an image stream to the edge server to receive FollowMe instructions.
Images can be transmitted using the JPEG compression method to maintain a high visual quality.
We also test the vector quantization (VQ)-based communication technique to significantly reduce the bandwidth footprint~\cite{vitvqgan,vitvqgan_comm}.
In this scheme, visual features extracted from images are quantized into compact \emph{codeword indices} (e.g., $13$ bits per token for a $K=8192$ codebook).
The core objective of an orchestrator is to maintain stable tracking and smooth motion under time-varying WiFi conditions, by dynamically switching the uplink sensing representation between (i) conventional JPEG-compressed images and (ii) compact VQ codewords.

\begin{tcolorbox}[colback=gray!5!white,
                  colframe=gray!55!black,
                  title=What is orchestrated in Demo~III (and what we measure end-to-end),
                  fonttitle=\bfseries,
                  colbacktitle=gray!75!black,
                  coltitle=white,
                  breakable,
                  enhanced,
                  left=2mm, right=2mm, top=1mm, bottom=1mm]
\begin{itemize}
    \item \textbf{Sensing (semantic-sensing switch):} choose \emph{JPEG} (high fidelity, larger payload) vs.\ \emph{VQ-based technique} (low payload, stable uplink), and optionally adjust parameters (JPEG compression quality ratio, VQ fidelity granularity).
    
    \item \textbf{Communication (practical stack knobs):} configure ROS2 / DDS QoS (e.g., reliable vs.\ best-effort), and monitor link indicators (RSSI / RTT / loss) to trigger mode switching.
    
    \item \textbf{Computation and actuation:} edge-side human detection/tracking produces velocity commands; the robot executes commands under a Command-to-Action (CTA) constraint to avoid oscillation or overshoot.
    
\end{itemize}
\end{tcolorbox}

\subsubsection{System architecture and hardware prototype}
\begin{figure*}[t]
\centering
\includegraphics[width=0.92\textwidth]{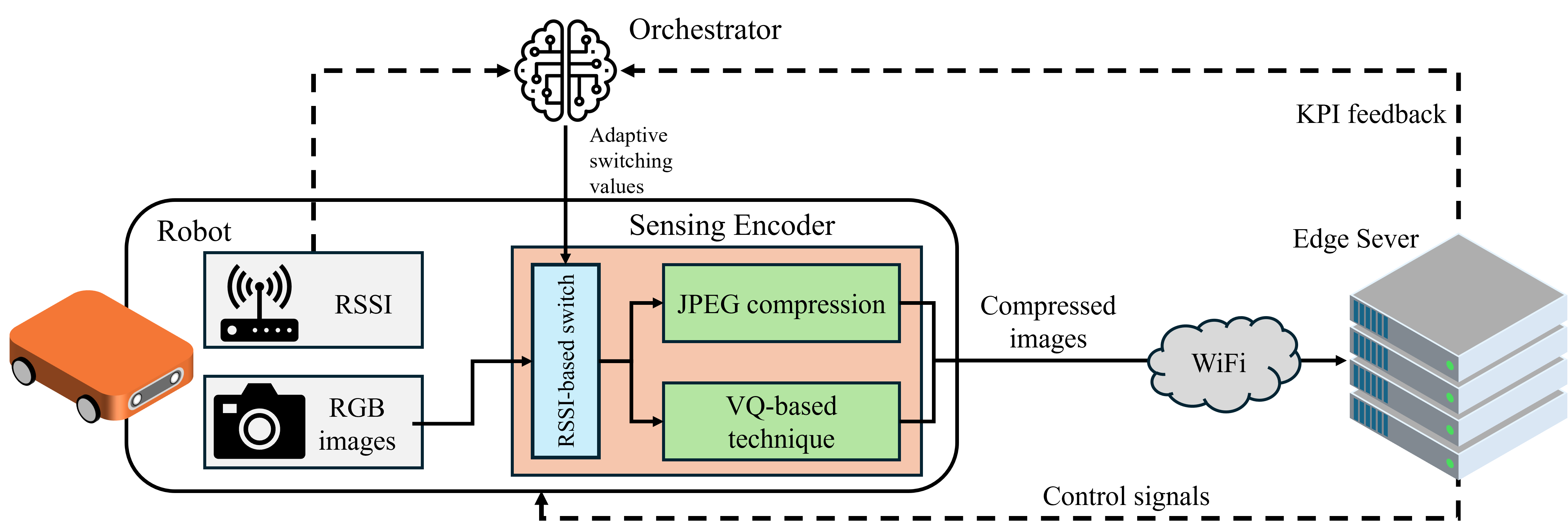}
\caption{Demo III real-hardware FollowMe pipeline. The orchestrator switches between JPEG image streaming and compact ViT embeddings based on WiFi state to stabilize closed-loop tracking and control.}
\label{fig:demo3_system}
\end{figure*}

\begin{figure*}[t]
    \centering
    \begin{subfigure}[b]{0.32\textwidth}
        \centering
        \includegraphics[width=\linewidth]{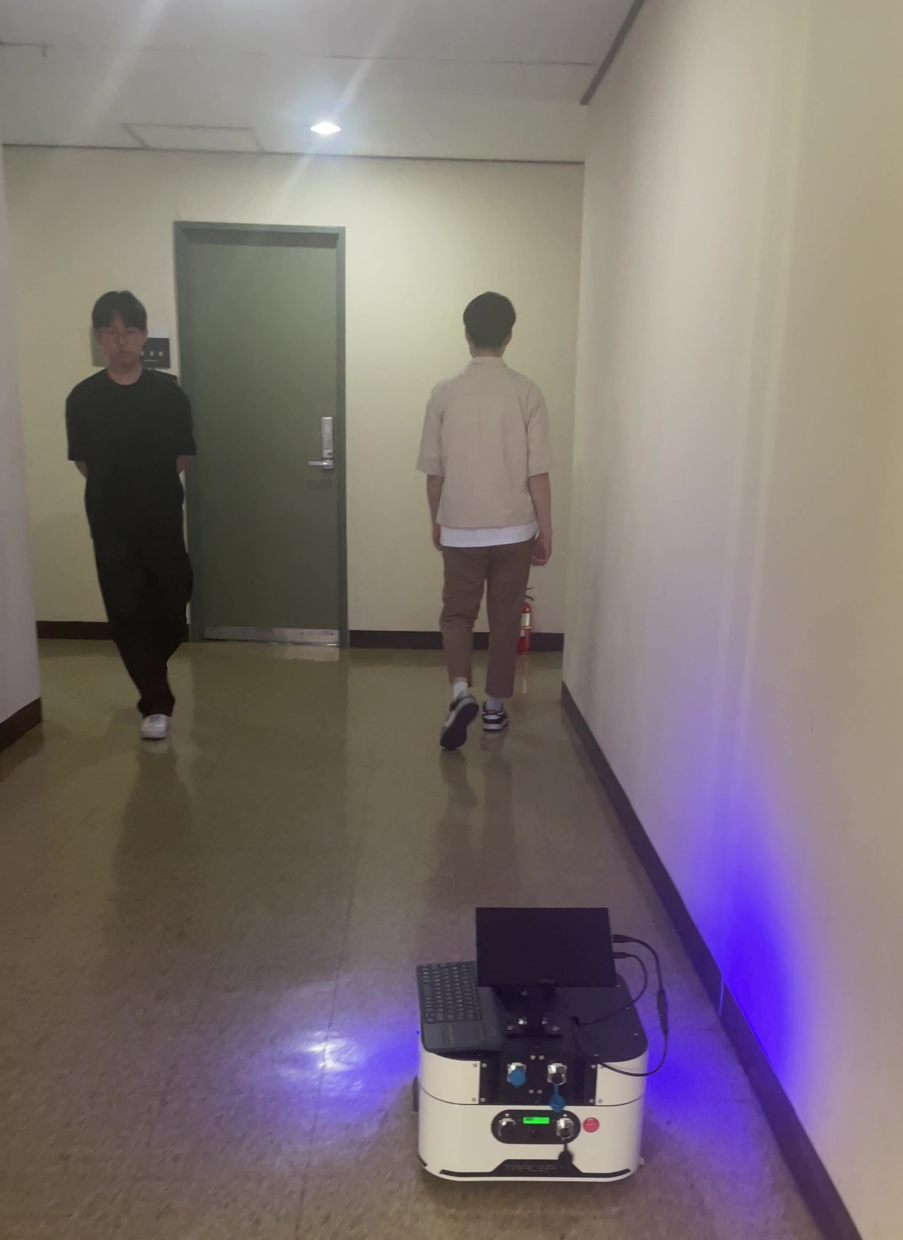}
        \caption{Corridor FollowMe run}
        \label{fig:demo3_corridor}
    \end{subfigure}
    \hfill
    \begin{subfigure}[b]{0.32\textwidth}
        \centering
        \includegraphics[width=\linewidth]{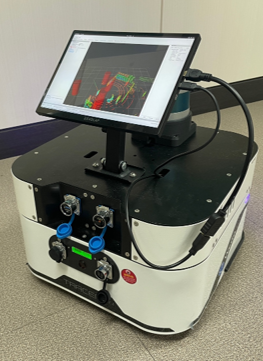}
        \caption{On-robot sensing \& UI}
        \label{fig:demo3_hw2}
    \end{subfigure}
    \hfill
    \begin{subfigure}[b]{0.32\textwidth}
        \centering
        \includegraphics[width=\linewidth]{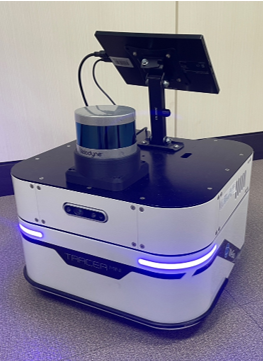}
        \caption{Mobile base (FollowMe prototype)}
        \label{fig:demo3_hw1}
    \end{subfigure}
    \caption{Demo~III prototype photos. The system performs human tracking at the edge and sends velocity commands to the robot; uplink sensing is adaptively compressed based on WiFi state.}
    \label{fig:demo3_photos}
\end{figure*}

Figure~\ref{fig:demo3_system} illustrates the closed-loop pipeline.
The robot reports the RSSI, which is varied due to distance and the WiFi environment, every $0.5$ second to the server.
Using the reported RSSI, an orchestrator commands the robot to utilize the appropriate encoding technique that ensures a steady stream of images, facilitating smooth robot responsiveness.
After the server receives and decompresses/decodes the camera observation, the sensing information (i.e., images) is used for the FollowMe stack, which consists of object detection and user tracking modules.
Finally, the FollowMe stack returns actuation commands to the robot.
Figure~\ref{fig:demo3_photos} shows the snapshot of the proposed FollowMe robot experiment. 

\subsubsection{Communication setup and sensing modes}
The WiFi link quality varies with distance, indoor multipath, occlusions, and more.
For instance, during the demonstration, the WiFi network is occupied with $15$ other WiFi networks.
To reduce packet delay variation and improve reproducibility, we configure the WiFi router to separate 2.4~GHz and 5~GHz bands and run the demo exclusively \textbf{in the} 5~GHz band (higher uplink capacity but shorter range).

\paragraph{Mode 1: JPEG compression.}
The FollowMe robot transmits JPEG-compressed RGB images to the edge server.
JPEG quality (e.g., Q=95/80/60) provides a direct bitrate--quality knob.

\paragraph{Mode 2: VQ-based communication technique.}
Instead of transmitting images, the FollowMe robot extracts visual features using a lightweight ViT encoder, which is then quantized into compact codewords.
For example, a $256\times 256$ is converted to $1024$ codewords, each of which is represented by $13$-bits.
To trade off granularity and latency, we optionally partition the image into an $a\times b$ grid and encode each patch independently (e.g., $1\times 3$), increasing semantic detail at the cost of a larger codeword payload and higher encoding latency.

\subsubsection{Codec-level microbenchmark (measured) and implication}
Table~\ref{tab:demo3_codec_microbench} reports a representative per-frame microbenchmark under varying WiFi conditions when the robot is $5\,$m away from the WiFi access point.
Specifically, JPEG transmission achieves a lower mean square error (MSE) than that achieved by the VQ-based technique.
Also, transmitting discretized codewords significantly reduces the payload size, showing potential when the link becomes unstable or when multiple robots contend for the uplink.
In particular, the jitter of transmitting codewords is $7.85$, achieving a $61.25\,\%$ reduction compared to the JPEG compression method.

\begin{table*}[t]
\centering
\caption{Demo~III codec-level microbenchmark (5~m LOS, 5~GHz WiFi; representative single-frame measurement).}
\label{tab:demo3_codec_microbench}
\small
\renewcommand{\arraystretch}{1.15}
\setlength{\tabcolsep}{5pt}
\begin{tabular}{lcccccc}
\toprule
\textbf{Mode} & \textbf{Enc.} & \textbf{Comm.} & \textbf{Dec.} & \textbf{Total} & \textbf{Payload} & \textbf{MSE} \\
 & \textbf{(ms)} & \textbf{(ms)} & \textbf{(ms)} & \textbf{(ms)} & \textbf{(KB)} & \textbf{diff} \\
\midrule
JPEG (Q=95) & 6.5 & 22.34 $\pm$ 20.26 & 1.6 & 30.44 & 80.00 & 0 \\
JPEG (Q=80) & 5.6 & 22.32 $\pm$ 19.90 & 1.5 & 29.42 & 33.38 & 0.00017 \\
JPEG (Q=60) & 5.4 & 21.68 $\pm$ 12.26 & 1.5 & 28.58 & 22.28 & 0.00025 \\
ViT-small (grid $1\times 3$) & 76.4 & 14.01 $\pm$ 7.85 & 3.6 & 94.1 & 5.16 & 0.00225 \\
ViT-small (grid $1\times 2$) & 54.5 & 12.28 $\pm$ 6.42 & 2.4 & 69.25 & 3.44 & 0.00279 \\
ViT-small (grid $1\times 1$) & 43.5 & 11.49 $\pm$ 4.39 & 1.2 & 56.17 & 1.72 & 0.00393 \\
\bottomrule
\end{tabular}
\end{table*}

\subsubsection{Orchestration policy (runtime mode switching)}
A key engineering observation is that \emph{reliable} ROS middleware can increase the control latency under packet loss due to retransmissions and queueing. Therefore, the orchestrator jointly configures: (i) sensing/codec mode and (ii) communication QoS.

In our prototype, the orchestrator emits a compact \texttt{sense\_config} message:
\begin{itemize}
    \item \texttt{mode} $\in \{ \text{\texttt{jpeg, vq}} \}$
    \item \texttt{jpeg\_quality} $\in \{ 95, 80, 60 \}$
    \item \texttt{vit\_grid} $\in \{ 1\times 1, 1\times 2, 1\times 3 \}$ (optional granularity knob)
    \item \texttt{qos} $\in \{ \text{\texttt{reliable, best\_effort}} \}$
\end{itemize}

To determine the proper RSSI values for switching rules, the FollowMe stack reports the KPIs (e.g., latency, standard deviation, and 95 percentile) to the orchestrator.
In particular, we position the robot at three pre-defined locations with distance of $2$, $5$, and $7\,$m respectively.
Then, the LLM collects the reported KPIs of $6$ compression settings.
Using the reported KPI, the orchestrator determines the switching values that balances the image quality, transmission latency, and connection stability.
During the run time, the LLM can continue collecting KPIs and re-evaluate the switching values and adapt to robot's changing environment.
For instance, a LLM-powered switching rule is:
\begin{itemize}
    \item if RSSI $\ge -39$~dBm ($1\,$m): use \texttt{jpeg} (Q=80) for high fidelity;
    \item if RSSI $\ge -41$~dBm ($2\,$m): use \texttt{jpeg} (Q=60) for sufficient fidelity;
    \item if RSSI $\ge -43$~dBm ($4\,$m): use \texttt{vq} with $1\times 3$ grid for the highest granularity;
    \item if RSSI $\ge -45$~dBm ($5\,$m): use \texttt{vq} with $1\times 2$ grid for acceptable granularity;
    \item if RSSI $\ge -48$~dBm ($6\,$m): use \texttt{vq} with $1\times 1$ grid for best effort in challenging wireless environment;
\end{itemize}
This setup ensures that the demo is aligned well with the R2X principle: adapt \emph{what to send} (semantics vs.\ pixels) and \emph{how to send} (QoS / rate) according to the task-level control requirement.

\subsubsection{Evaluation protocol and metrics}
We evaluate FollowMe stability over repeated corridor trials, each consisting of a $10\,$m walk in crowded WiFi environments.
We report (i) closed-loop control delay (including sensing data transmission and FollowMe stack), and (ii) tracking robustness, using two metrics: \textbf{Command-to-Action latency (CTA)} and \textbf{User Tracking Failure Rate (UTFR)}.
Following the definitions used in our earlier task-level latency analysis, CTA measures end-to-end loop delay from sensing to actuation, and UTFR measures the fraction of time the user is lost due to perception or communication failures.

\subsubsection{End-to-end results}
Table~\ref{tab:demo3_projected} summarizes end-to-end performance of various schemes.
We observe that the orchestrator maintains a low-latency and stable connection between the robot and the edge server.
In particular, the orchestrator does not consider the JPEG compression option with $95$ quality for transmission due to unfavourable characteristics (e.g., high payload, high standard deviation, and high $95$ percentile).
Furthermore, due to the steady stream of images, the edge server achieves the lowest UTFR, controlling the robot to closely follow the user.
Specifically, the orchestrated robot achieves the perfect ($100\,\%$) FollowMe success rate due to the frequent image transmission.
On the other hand, while transmitting VQ codewords at a granularity of $1\times1$ requires the lowest bandwidth footprint, the received images are of low quality (e.g., blurry), providing limited visual information for reliable user tracking.
Specifically, while transmitting VQ codeword embeddings achieves the lowest CTA at $17.93\,$ms, the orchestrated robot achieves real-time CTA at $32.62\,$ms.
Nonetheless, using the JPEG compression technique leads to high latency and unstable connections, indicated by the latency with high standard deviation and high $95$ percentile.
This is because WiFi quality is reduced (lower RSSI) when the robot is far away from the access point, limiting the overall bandwidth and preventing high payload transmission.

\begin{table*}[t]
\centering
\caption{Demo~III end-to-end performance.
Metrics averaged over repeated corridor runs with varying WiFi quality.}
\label{tab:demo3_projected}
\small
\renewcommand{\arraystretch}{1.3} 
\setlength{\tabcolsep}{3pt}      
\begin{tabularx}{\textwidth}{l X c c c c c}
\toprule
\textbf{Method} & \textbf{Sensing policy} &
\textbf{\makecell{Avg. payload \\ (KB)}} &
\textbf{\makecell{CTA mean \\ (ms)}} &
\textbf{\makecell{CTA 95\% \\ (ms)}} &
\textbf{\makecell{UTFR \\ (\%)}} &
\textbf{\makecell{Success ratio \\ (\%)}} \\
\midrule
JPEG-only (HQ) &
Fixed JPEG streaming (Q=95), reliable QoS &
420.0 & 118.75 $\pm$ 188.83 & 514.77 & 63.73 & 75 \\
\addlinespace
JPEG-only (LQ) &
Fixed JPEG streaming (Q=60), best-effort QoS &
120.0 & 108.10 $\pm$ 157.82 & 363.87 & 26.47 & 75 \\
\addlinespace
ViT-only &
Fixed ViT embedding ($1\times 1$), best-effort QoS &
1.72 & 7.99 $\pm$ 6.86 & 17.93 & 25.37 & 75 \\
\addlinespace
\textbf{Orchestrated} &
\textbf{LLM-powered switch:} JPEG(Q=80) $\rightarrow$ ViT($1\times 3$ or $1\times 1$) + QoS adaptation &
\textbf{15.04} & \textbf{19.60} $\pm$ \textbf{9.69} & \textbf{32.62} & \textbf{8.47} & \textbf{100} \\
\bottomrule
\end{tabularx}
\end{table*}

\subsubsection{Deployment notes}
This demo highlights a practical, implementation-level insight: in realistic robotic networks, \emph{semantic-sensing switches} provide an effective control knob to balance between sensing data quality and the connection stability.
By reducing uplink payloads and enabling QoS-aware adaptation, the orchestrator not only improves CTA tail latency but also enhances tracking robustness.
These gains directly translate into smoother and safer robot navigation in shared indoor environments.
This is because in FollowMe and general robotic applications, a safe navigation inherently requires steady streams of sensing data to issue commands in time.

\subsubsection{Limitations and future research directions}
\label{sec:demo3_limits_future}
In multi-robot deployments, the orchestrator must account for Wi-Fi contention among robots.
In such scenarios, simple RSSI measurements fail to capture the actual network conditions, as multiple robots share the same bandwidth.
Moreover, the orchestrator must dynamically balance image quality by frequently switching compression modes, even when the selected mode is suboptimal for an individual robot.

Beyond the multi-robot scenarios, an LLM-based orchestrator would definitely benefit from multi-model inputs.
For example, incorporating robots’ historical positions enables trajectory prediction, allowing the orchestrator to shift from reactive to proactive compression mode switching and maintain stable transmission in dynamic environments.
Furthermore, integrating sensing data provides fine-grained context for compression mode selection.
Through environment-aware communication, the orchestrator can better characterize multipath conditions and select more appropriate compression modes.

\subsection{Real-Hardware Demo IV: Open-Vocabulary Trash Sorting via Edge-Assisted MLLM Grounding}
\label{sec:demo4_trashsorting}

\subsubsection{Demo objective and why edge \& communication matter}
This demo targets a practical limitation of on-device (closed-set) perception: 
\emph{unseen} or \emph{non-predefined} objects cannot be reliably detected if they are outside the detector's label space, and 
objects outside the robot's field-of-view (FOV) require time-consuming exploration.
We consider a room-to-corridor scenario where (i) a target item (e.g., \emph{paper cup} or \emph{can}) is located \emph{inside} a room, and 
(ii) two bins (general trash vs.\ recycling) are located \emph{outside} the room in a corridor.

The key objective is to show that \textbf{edge robotics + communication} enables:
(i) \textbf{open-vocabulary} object grounding from \textbf{text instructions} (no class predefinition),
and (ii) \textbf{out-of-FOV} target discovery via \textbf{multi-view sensing} (robot camera + corridor camera),
while maintaining a bounded closed-loop delay for stable manipulation and navigation.

\begin{tcolorbox}[colback=gray!5!white,
                  colframe=gray!55!black,
                  title=What is orchestrated in Demo~IV (and what we measure end-to-end),
                  fonttitle=\bfseries,
                  colbacktitle=gray!75!black,
                  coltitle=white,
                  breakable,
                  enhanced,
                  left=2mm, right=2mm, top=1mm, bottom=1mm]
\begin{itemize}
    \item \textbf{Sensing:} multi-view perception (\emph{robot ego camera} inside the room + \emph{corridor camera} outside the room);
    optional \textbf{semantic-sensing switch} (pixels vs.\ compressed/reconstructed view) for uplink efficiency.
    \item \textbf{Communication:} uplink two video streams (or compressed semantic views) to an edge server; downlink bounding boxes and task goals.
    \item \textbf{Computation:} edge-hosted \textbf{MLLM grounding} produces bounding boxes for \emph{arbitrary} text queries (e.g., ``paper cup'', ``recycling bin'').
    \item \textbf{Actuation:} grasp the item with the arm, then navigate (SLAM) to the correct bin and dispose (place/drop).
\end{itemize}
\end{tcolorbox}

\subsubsection{System overview (sense--communicate--compute--act)}
\begin{figure*}[t]
    \centering
    \includegraphics[width=0.92\textwidth]{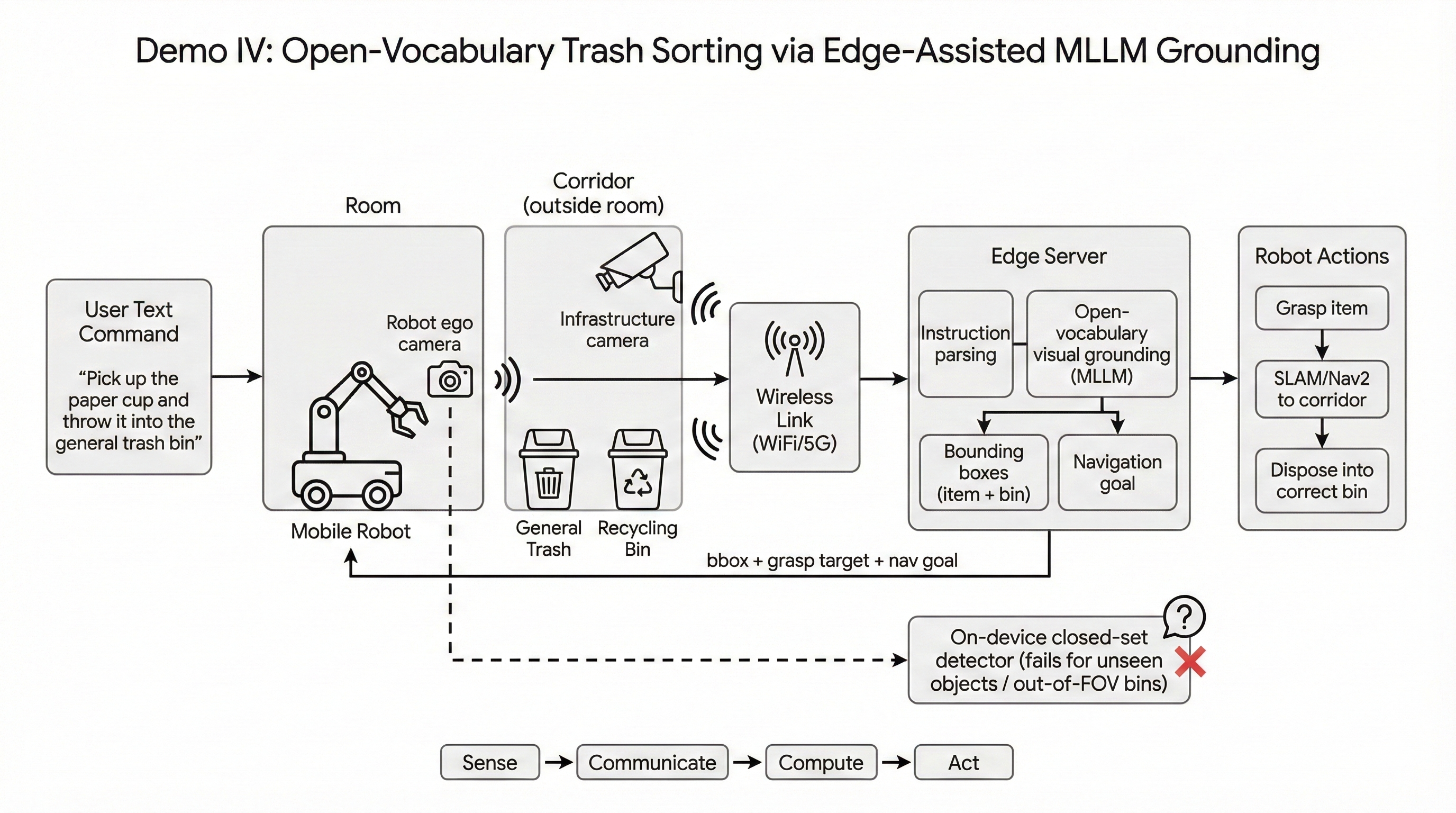}
    \caption{Demo~IV orchestration overview: multi-view sensing (robot camera + corridor camera) is uplinked to an edge server. The edge MLLM performs open-vocabulary grounding (returns bounding boxes) from text instructions, then downlinks grasp/navigation goals. The robot grasps the item and navigates (SLAM) to the correct bin for disposal.}
    \label{fig:demo4_system}
\end{figure*}

Figure~\ref{fig:demo4_system} illustrates the end-to-end loop.
A user provides a natural-language command, e.g.,
\emph{``Pick up the paper cup and throw it into the general trash bin''} or
\emph{``Pick up the can and throw it into the recycling bin''}.
An edge orchestrator parses the instruction into a structured goal 
(target item + target bin type), then runs open-vocabulary visual grounding to obtain bounding boxes for (i) the item in the room
and (ii) the correct bin in the corridor using a fixed corridor camera.
The robot uses depth + camera intrinsics to convert 2D boxes into 3D grasp/approach targets, executes manipulation, and navigates using SLAM.

\subsubsection{Hardware prototype and scene layout}
\begin{figure*}[t]
    \centering
    \includegraphics[width=0.80\textwidth]{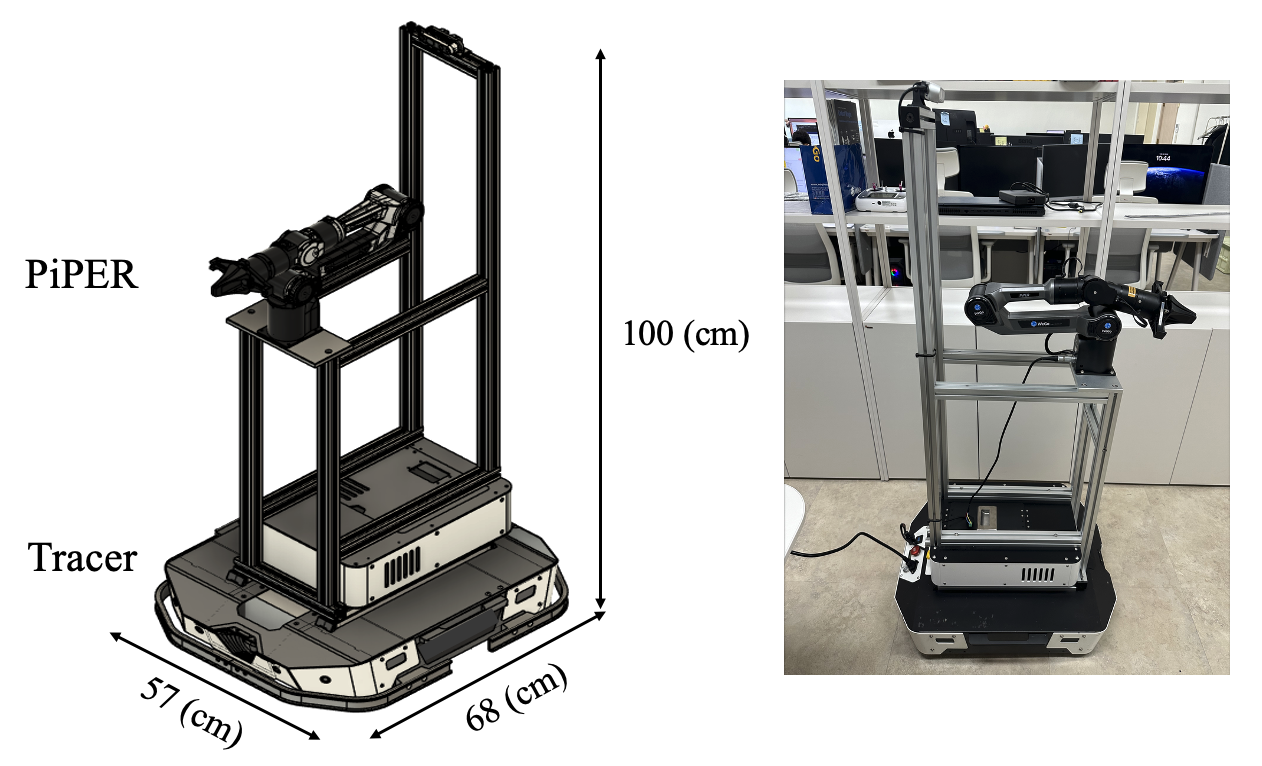}
    \caption{Demo~IV prototype mobile manipulation platform (mobile base + robot arm).}
    \label{fig:demo4_platform_photo}
\end{figure*}

We use a mobile manipulation platform comprising a mobile base and a robot arm (Fig.~\ref{fig:demo4_platform_photo}).
The room contains the target item (paper cup or can), while the corridor contains the bins (non-recyclables vs.\ recycling).
A fixed corridor camera provides an external view that is not available to the robot inside the room.

\begin{figure*}[t]
    \centering
    \includegraphics[width=0.88\textwidth]{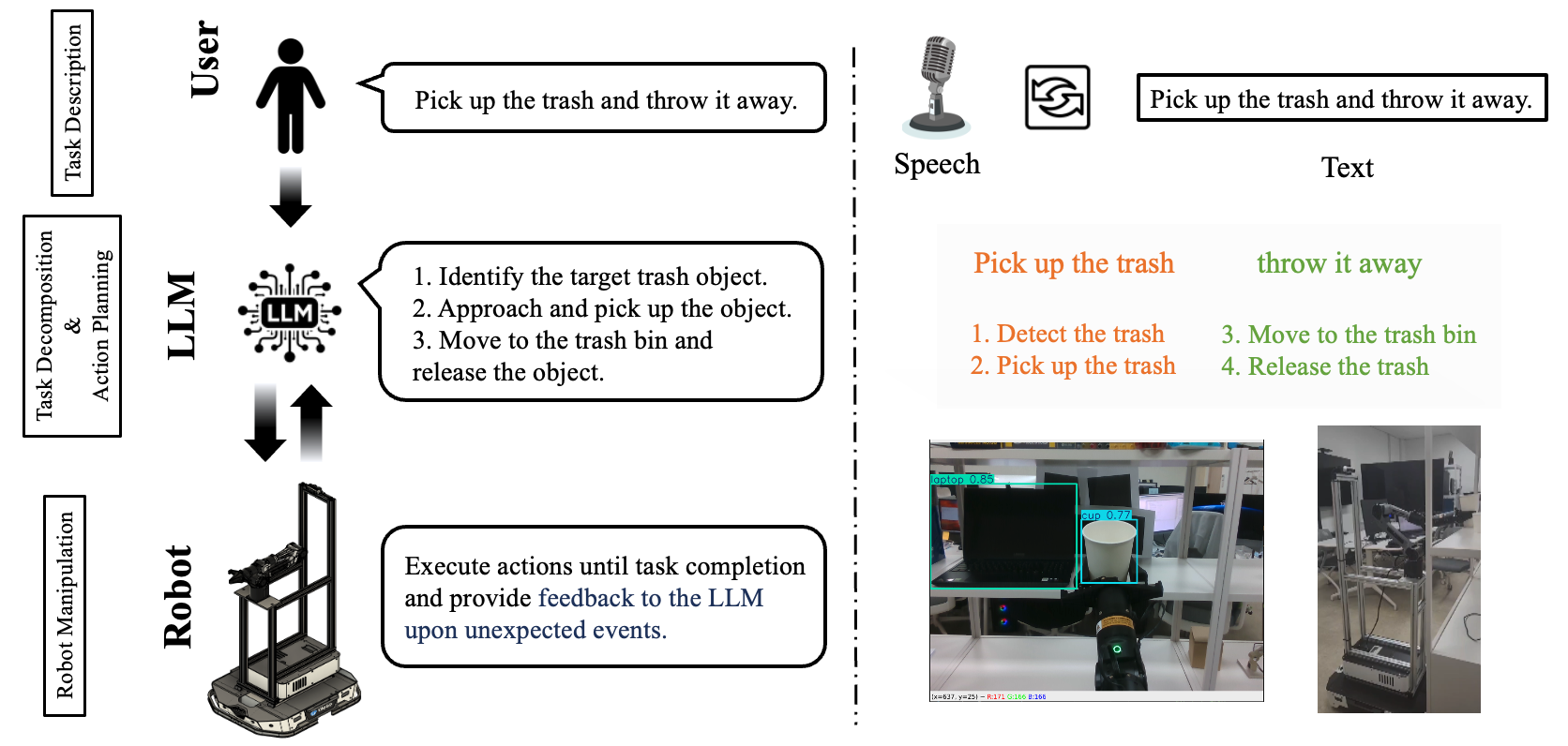}
    \caption{Demo~IV task concept: open-vocabulary pick \& dispose from a text instruction.}
    \label{fig:demo4_task_concept}
\end{figure*}

\begin{figure*}[t]
    \centering
    \includegraphics[width=0.90\textwidth]{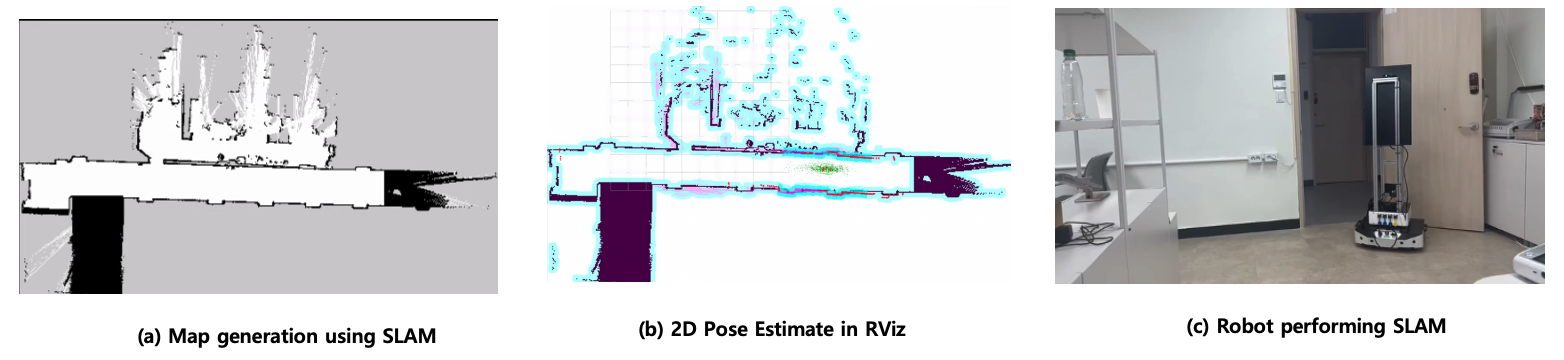}
    \caption{Demo~IV room-to-corridor navigation: representative SLAM/map view used for bin approach.}
    \label{fig:demo4_slam_example}
\end{figure*}

\subsubsection{Semantic-sensing switch: closed-set YOLO vs.\ open-vocabulary MLLM grounding}
\begin{figure*}[t]
    \centering
    \includegraphics[width=0.88\linewidth]{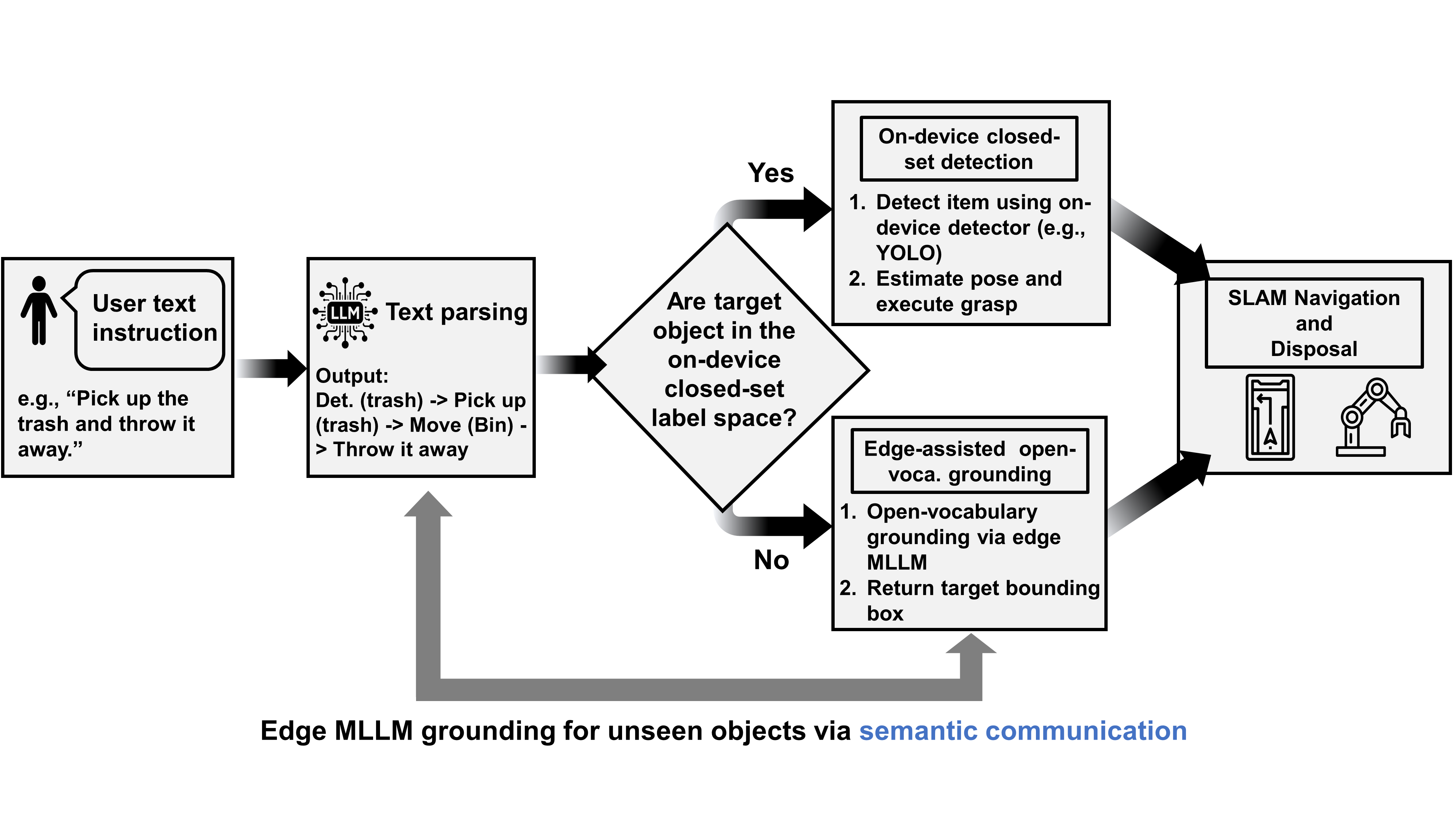}
    \caption{Demo~IV decision logic: closed-set on-device detection when labels exist; otherwise, switch to edge MLLM grounding with multi-view sensing (robot + corridor camera).}
    \label{fig:demo4_switch_logic}
\end{figure*}

A key design choice is whether the perception target is covered by the on-device detector's label set.
Figure~\ref{fig:demo4_switch_logic} summarizes the decision logic:
if the user-requested objects match the closed-set detector labels, the robot runs on-device YOLO; otherwise, it uplinks observations and invokes edge MLLM grounding to obtain bounding boxes for \emph{unseen} object names
(and, when needed, out-of-FOV bins via the corridor camera).

\paragraph{Why open-vocabulary grounding is essential.}
In our setting, the instruction can include objects that are \emph{not pre-defined} (e.g.,``trash bin'', ``headphones'', etc.).
Closed-set detectors may fail or mis-detect such targets, forcing the robot to either (i) search blindly or (ii) collect a new dataset and retrain.

\begin{figure*}[h!t]
    \centering
    \includegraphics[width=0.8\linewidth]{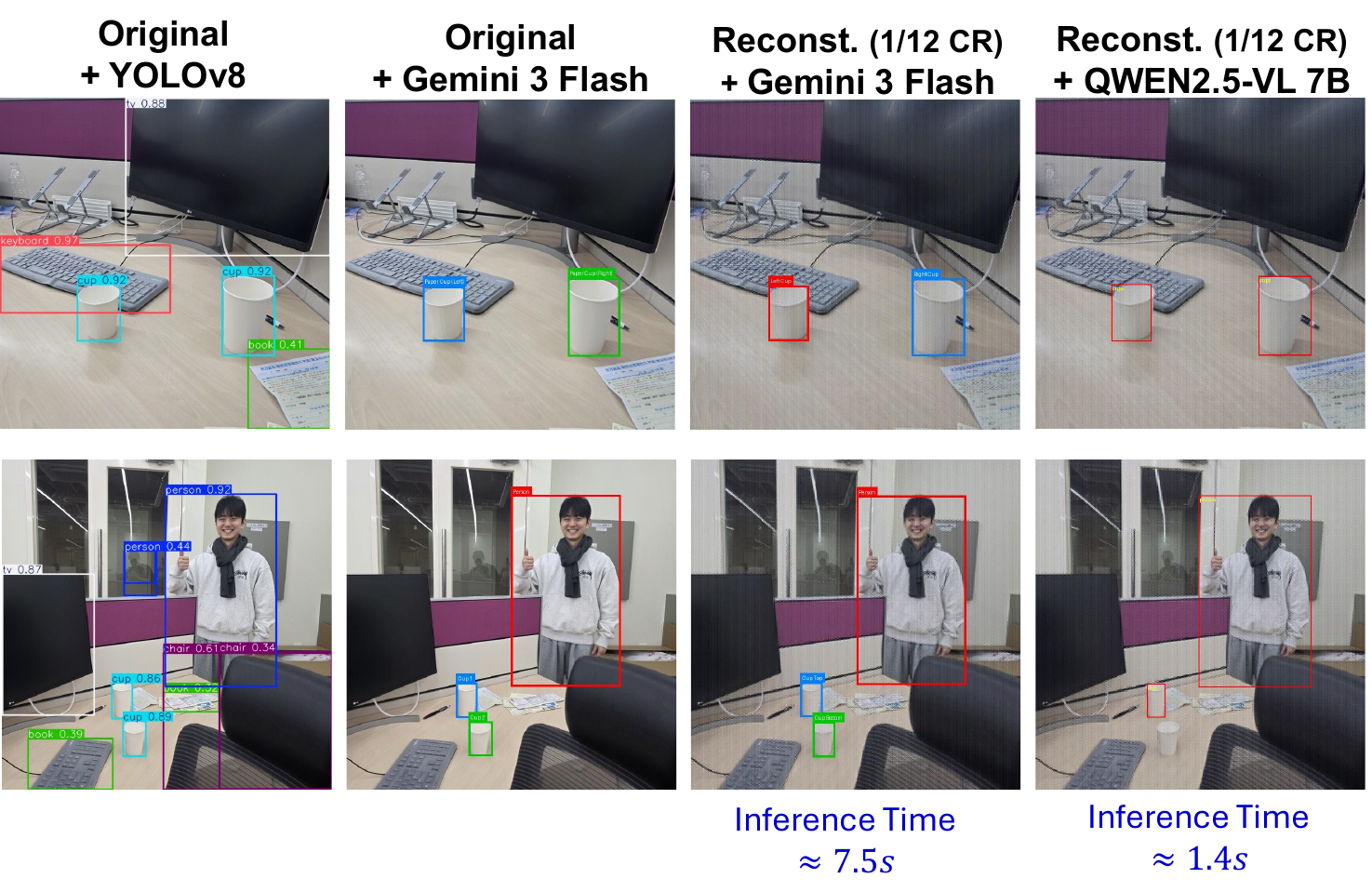}
    \caption{Open-vocabulary grounding example on a reconstructed image after semantic communication with a 1/12 compression ratio. Given a text query describing a target object that is classifiable by YOLO (e.g., a cup or a person), the MLLM localizes the corresponding region in the image and returns its bounding box coordinates.}
    \label{fig:demo4_grounding_examples_1}
\end{figure*}

\begin{figure*}[h!t]
    \centering
    \includegraphics[width=0.8\linewidth]{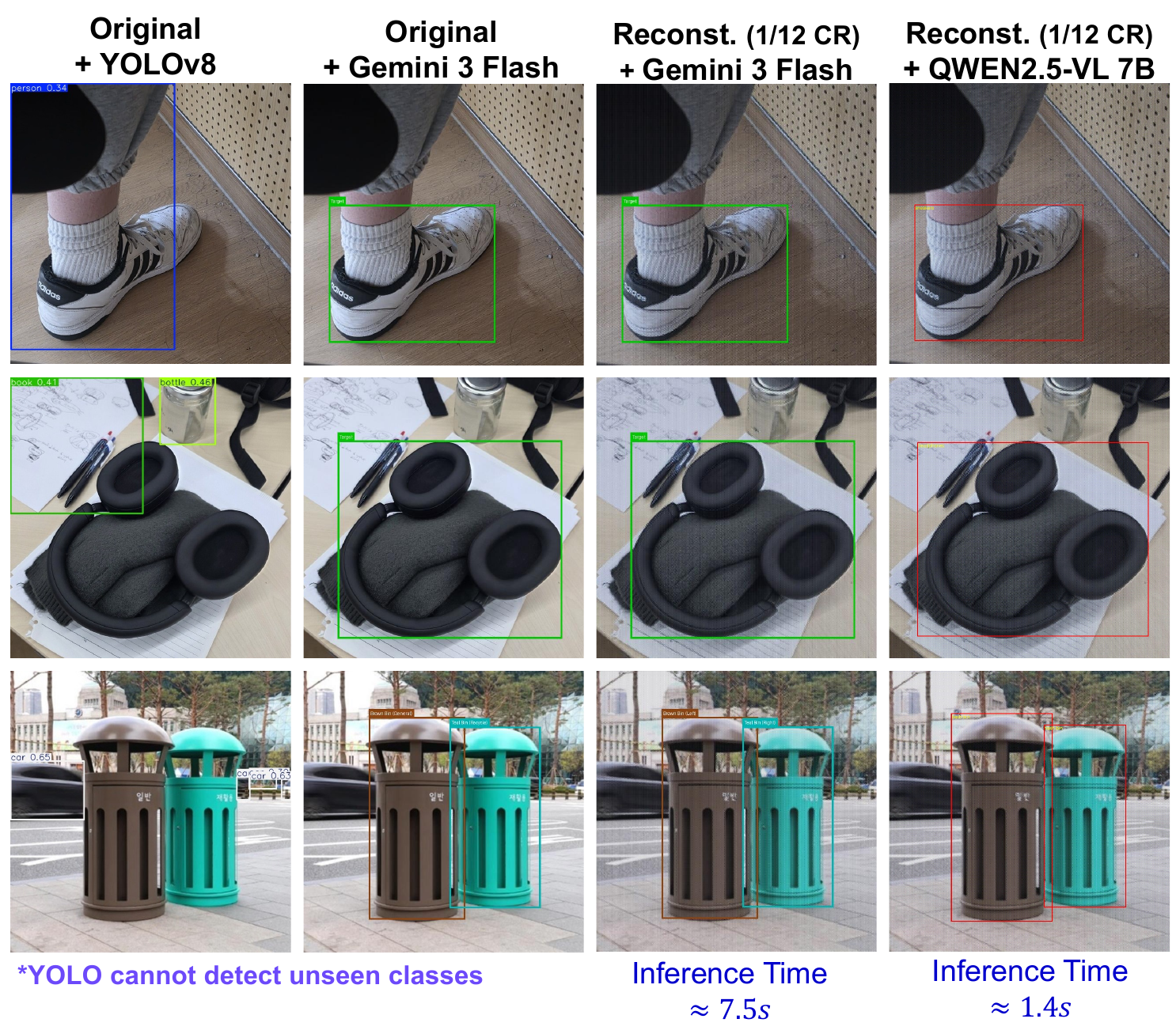}
    \caption{Open-vocabulary grounding example on a reconstructed image after semantic communication with a 1/12 compression ratio. Given a text query describing a target object that is not classifiable by YOLO (e.g., sneakers, headphones, or a trash bin), the MLLM localizes the corresponding region in the image and returns its bounding box coordinates.}
    \label{fig:demo4_grounding_examples_2}
\end{figure*}

Figures \ref{fig:demo4_grounding_examples_1} and \ref{fig:demo4_grounding_examples_2} illustrate representative grounding results for both common and long-tail objects. Figure \ref{fig:demo4_grounding_examples_1} shows that, for closed-set categories, performing grounding on images reconstructed via semantic communication—using Gemini 3 Flash and Qwen2.5-VL-7B—achieves qualitatively comparable accuracy to original-image inference, thereby enabling bandwidth-efficient uplinks.
Furthermore, Figure \ref{fig:demo4_grounding_examples_2} highlights the superiority of open-vocabulary models in grounding objects that are not reliably supported by a closed-set detector (e.g., YOLOv8n), proving that prompting and semantic reconstruction can overcome the limitations of fixed-label detectors.

\subsubsection{Baseline illustration: closed-set on-device detection and the retraining burden}
\begin{figure*}[t]
    \centering
    \begin{subfigure}[b]{0.49\textwidth}
        \centering
        \includegraphics[width=\linewidth]{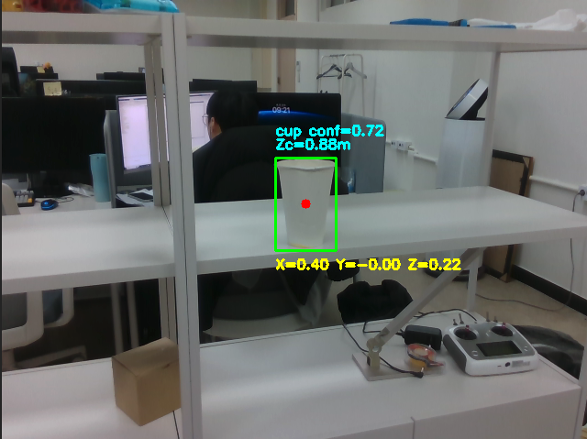}
        \caption{On-device closed-set detection (YOLO)}
        \label{fig:demo4_yolo_baseline}
    \end{subfigure}
    \hfill
    \begin{subfigure}[b]{0.49\textwidth}
        \centering
        \includegraphics[width=\linewidth]{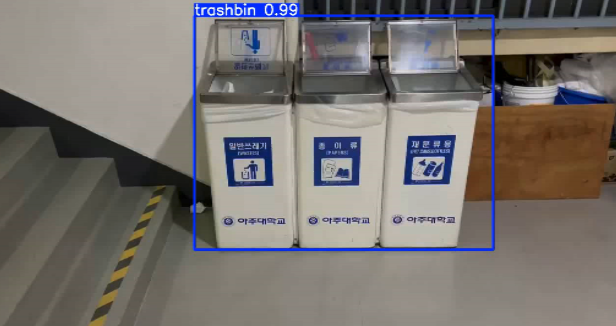}
        \caption{Closed-set retraining burden (bins/items)}
        \label{fig:demo4_bin_retrain}
    \end{subfigure}
    \caption{Baseline illustration: closed-set detection is fast when labels exist, but extending to new targets (e.g., bins) often requires additional labeling and retraining.}
    \label{fig:demo4_baseline_pair}
\end{figure*}

Figure~\ref{fig:demo4_yolo_baseline} shows an example of on-device YOLO detection.
While such detection is efficient when the label set matches the target, it does not naturally support open-vocabulary queries.
Moreover, recognizing new targets (e.g., bins) often requires collecting labeled images and retraining, as illustrated in Fig.~\ref{fig:demo4_bin_retrain}.

\subsubsection{End-to-end pipeline: from bounding boxes to grasp \& navigation goals}
Given a bounding box \((u_{\min},v_{\min},u_{\max},v_{\max})\), we select a grasp point (e.g., box center) and recover its 3D location using a depth map:
\begin{equation}
(u_c,v_c)=\Big(\tfrac{u_{\min}+u_{\max}}{2},\tfrac{v_{\min}+v_{\max}}{2}\Big),
\quad
\mathbf{p}_{\mathrm{cam}} = z \cdot \mathbf{K}^{-1}[u_c, v_c, 1]^\top,
\label{eq:demo4_3d_point}
\end{equation}
where \(z\) is depth at \((u_c,v_c)\) and \(\mathbf{K}\) is the camera intrinsic matrix.
The 3D grasp target is transformed into the robot base frame and passed to a motion planner for grasp execution.
After grasping the item, the robot navigates (SLAM) to the corridor and approaches the correct bin, which is localized by the corridor camera and grounded by the edge MLLM.

\subsubsection{Communication microbenchmarks (pilot) and implication}
To justify the edge-assisted design, we conducted a pilot uplink timing check comparing
(i) raw image streaming and (ii) compact representations (e.g., feature/semantic payloads).
In our pilot runs, compact payloads consistently reduced the uplink transmission time and its variance,
which is beneficial for stabilizing the closed-loop pipeline (grounding $\rightarrow$ grasp $\rightarrow$ navigation).
We therefore adopt semantic/compact uplinks as the default in the proposed edge-assisted setting,
and reserve raw-image streaming for cases where visual fidelity is required for failure analysis or debugging.

\subsubsection{Baselines and evaluation protocol}
We evaluate the following methods in the room-to-corridor trash sorting task:
\begin{itemize}
    \item \textbf{Baseline A (On-device closed-set + exploration):}
    a closed-set on-device detector (``object identifier'') is used; if the target is not recognized, the robot explores and attempts to proceed without reliable grounding. 
    Bins are out-of-FOV, so the robot must search the corridor without external vision; if bins are not detectable, disposal often fails.
    \item \textbf{Baseline B (Closed-set + retraining):}
    the detector is retrained with additional labeled data for bins/items. This improves detection when in-view, but still lacks open-vocabulary flexibility and can require significant dataset collection.
    \item \textbf{Proposed (Edge MLLM + multi-view + semantic-sensing switch):}
    the system uses open-vocabulary grounding on the edge for both the target item (robot camera) and the bins (corridor camera), optionally using semantic compression/reconstruction to reduce uplink cost.
\end{itemize}

\noindent We report the following task-facing metrics:
(i) \textbf{the use of open-vocabulary grounding},
(ii) \textbf{object detection success rate} for both target items and bins,
(iii) \textbf{end-to-end task completion time} (command \(\rightarrow\) disposal), and (iv) \textbf{success ratio} within a fixed time budget. 
Based on the task-facing metrics defined above, detection success rates are
measured as follows. Item detection success is defined as the case where
the target item is detected from the robot camera input and the
intersection-over-union (IoU) with a manually defined ground-truth
bounding box is at least 0.5. This evaluation includes objects that are
pre-labeled in YOLO, objects requiring additional training, and objects not
seen during YOLO training, thereby assessing open-vocabulary recognition
performance.
Bin detection success is evaluated using images collected in a corridor
environment. A detection is considered successful when a bin is detected
from either the robot’s field of view or the corridor camera and achieves
an IoU of at least 0.5 with respect to the ground-truth bounding box.
End-to-end task completion time is measured from the moment a user
command is issued to the moment the item is successfully disposed of into
the designated bin. The success ratio is defined as the fraction of trials in
which the task is completed within a predefined fixed time budget.

\subsubsection{End-to-end results}
Table~\ref{tab:demo4_projected} reports performance based on pilot
measurements and integrated system behavior. As shown in the table, the
proposed approach consistently outperforms closed-set detection baselines
in terms of end-to-end task performance. Edge-based MLLM-driven
open-vocabulary grounding reduces recognition failures for unseen object
names, while multi-view perception enabled by the corridor camera
shortens the search time for bins located outside the robot’s field of view
(out-of-FOV). The combination of these components reduces unnecessary
re-exploration and motion, leading to improvements in both task
completion time and success ratio. In contrast, although on-device
closed-set detectors can benefit from additional training, their
generalization to novel objects and environmental variations remains
limited, which constrains task-level reliability.

\begin{table*}[t]
\centering
\caption{Demo~IV End-to-end performance in the room-to-corridor trash sorting task.}
\label{tab:demo4_projected}
\scriptsize
\renewcommand{\arraystretch}{1.2}
\setlength{\tabcolsep}{3pt}
\begin{tabularx}{\textwidth}{
    >{\raggedright\arraybackslash}p{0.11\textwidth}
    >{\raggedright\arraybackslash}X
    >{\centering\arraybackslash}p{0.08\textwidth}
    >{\centering\arraybackslash}p{0.08\textwidth}
    >{\centering\arraybackslash}p{0.10\textwidth}
    >{\centering\arraybackslash}p{0.10\textwidth}
    >{\centering\arraybackslash}p{0.10\textwidth}
    >{\centering\arraybackslash}p{0.10\textwidth}
}
\toprule
\textbf{Method} &
\textbf{Perception stack} &
\textbf{Corridor cam} &
\textbf{Open-vocab} &
\textbf{Item det. success (\%)} &
\textbf{Bin det. success (\%)} &
\textbf{Task time (s)} &
\textbf{Success ratio (\%)} \\
\midrule
Baseline A &
On-device closed-set detector + exploration &
-- &
No &
40--50 &
0 &
Almost Fail &
0 \\
\addlinespace
Baseline B &
On-device closed-set retrained for bins/items &
-- &
No &
50--70 &
70--90 &
$70 \pm 15$ (Seldom Fail) &
40--50 \\
\addlinespace
\textbf{Proposed} &
\textbf{Edge MLLM grounding + multi-view (robot + corridor)} &
\checkmark &
Yes &
\textbf{90--95} &
\textbf{90--100} &
\textbf{$55 \pm 5$} &
\textbf{90--100} \\
\bottomrule
\end{tabularx}
\end{table*}

\subsubsection{Deployment notes}
This demo emphasizes a practical R2X advantage:
\textbf{open-vocabulary perception and out-of-FOV discovery} are naturally enabled when sensing and computation are lifted to the edge
and supported by communication.
Unlike closed-set on-device detectors, an edge MLLM can ground arbitrary text queries (``paper cup'', ``recycling bin'') and return usable bounding boxes,
thereby reducing training overhead and improving robustness to novel object names.
Moreover, a corridor camera provides a low-cost infrastructure extension that eliminates blind exploration for targets outside the robot's view.

\subsubsection{Limitations and future research directions}
The current results are validated through integrated hardware trials, and the reported end-to-end performance reflects reliable time-stamping, motion planning robustness, and repeated runs under diverse lighting and WiFi/5G conditions. Open-vocabulary grounding quality can also vary with prompting, model choice, and image compression artifacts.

A natural extension is \textbf{uncertainty-aware grounding} (e.g., multiple candidate boxes + confidence) combined with
risk-aware grasp planning and conservative navigation when the bin location is ambiguous.
Another direction is \textbf{multi-robot scaling}, where multiple robots share corridor cameras and edge compute; this introduces contention and requires semantic multiple access and compute-aware scheduling to maintain task-level success under load.

\subsection{Cross-Demo Synthesis: What R2X Enables and When Each Knob Matters}
\label{sec:demo_synthesis}

Across Demos~I--IV, we intentionally vary realism (digital-twin $\rightarrow$ simulated mobility $\rightarrow$ real hardware)
while keeping the same orchestration loop explicit: \emph{sense} $\rightarrow$ \emph{communicate} $\rightarrow$ \emph{compute} $\rightarrow$ \emph{act}.
A consistent takeaway is that R2X is not a single algorithmic block; rather, it is a \emph{policy layer} that selects
(i) \emph{what to transmit} (pixels vs.\ semantics), (ii) \emph{how to transmit} (predictive vs.\ reactive link adaptation / QoS),
and (iii) \emph{where to compute} (on-robot vs.\ edge) to satisfy a \emph{task-facing constraint} such as safe navigation or stable control.

\subsubsection{Cross-demo summary table (task, knobs, and measurable outcomes)}
\label{sec:demo_summary_table}
Table~\ref{tab:demo_summary} summarizes the four demonstrations in a unified format, emphasizing (i) the task and environment,
(ii) the orchestrated knobs that are actually exercised, and (iii) the \emph{measurable} system-level outcomes reported in each demo.
Our demos are not isolated case studies, but instances of the same closed-loop design logic
evaluated with latency/reliability/task metrics.

\begin{table*}[t]
\centering
\caption{Cross-demo synthesis: tasks, orchestrated knobs, and measurable outcomes across Demos~I--IV.}
\label{tab:demo_summary}
\small
\renewcommand{\arraystretch}{1.15}
\setlength{\tabcolsep}{6pt}
\begin{tabularx}{\textwidth}{l X X X}
\toprule
\textbf{Demo} &
\textbf{Task + environment} &
\textbf{Orchestrated knobs (sense/comm/compute)} &
\textbf{Measured (or reported) outcomes} \\
\midrule
Demo~I &
Warehouse navigation in a ray-tracing digital twin with dynamic humans; two robots must reach goals safely under link variability. &
\textbf{Sense:} raw RGB-D vs.\ compact semantic features; \textbf{Comm:} predictive path-gain-aware adaptation; \textbf{Compute:} server-side global replanning parameterized by schema-valid JSON. &
\textbf{Task completion time} across scenarios; \textbf{end-to-end RTT} feasibility relative to grid-transition time; qualitative trajectory stability (halt/wait reduction). \\
\addlinespace
Demo~II &
Mobility scenario with fast channel variation and delayed feedback; proactive MCS control under reliability constraint. &
\textbf{Sense:} uplink semantic features as link-prediction context; \textbf{Comm:} predictive MCS selection with target BLER; \textbf{Compute:} centralized \tcb{MLLM} infers near-future link proxy under delay. &
\textbf{Throughput/latency/BLER} time-series stability and SNR-conditioned performance; aggregate gains under delayed feedback. \\
\addlinespace
Demo~III &
Real FollowMe robot in an indoor corridor over WiFi; maintain smooth tracking/control under interference and range variation. &
\textbf{Sense:} JPEG streaming vs.\ ViT embedding switch; \textbf{Comm:} practical QoS choice (reliable vs.\ best-effort); \textbf{Compute:} edge tracking/control with closed-loop delay sensitivity. &
Codec microbenchmark (payload/latency/MSE); \textbf{CTA/UTFR} and run success. \\
\addlinespace
Demo~IV &
Real mobile manipulation: open-vocabulary trash sorting across room (item) and corridor (bins) using multi-view sensing. &
\textbf{Sense:} robot camera + corridor camera; semantic-sensing switch; \textbf{Comm:} multi-view uplink + bbox/goal downlink; \textbf{Compute:} edge MLLM grounding + goal parsing; \textbf{Act:} grasp + SLAM navigation + disposal. &
Open-vocabulary detection success (IoU-based); time-to-first item/bin localization; end-to-end task completion time; success ratio. \\
\bottomrule
\end{tabularx}
\end{table*}

\subsubsection{Generalizable design rules observed across demos}
\label{sec:demo_design_rules}
Across the demonstrations, we observe three system-level ``rules of thumb'' that are useful beyond the specific scenarios.
First, \textbf{semantic sensing is most valuable when the bottleneck is closed-loop stability rather than perceptual fidelity}:
when packet loss/jitter or uplink contention drives tail latency (Demo~III) or when payload size limits timely replanning (Demo~I),
compact semantics reduce the probability of control-unstable delays. Second, \textbf{prediction is most valuable when the feedback is stale}:
in mobility (Demo~II) or motion planning with changing geometry (Demo~I), predictive link context reduces mismatch and prevents
reactive policies from oscillating between over- and under-conservative actions. Third, \textbf{compute placement must follow the task metric}:
server/edge inference is beneficial when global reasoning or heavy models improve task outcomes (Demo~I/III), but it must be paired with
communication-aware sensing; otherwise, offloading can increase CTA tail latency and degrade behavior (Demo~III).

\subsection{Practical Deployment Considerations and Reproducibility Notes}
\label{sec:deployment_notes}

\subsubsection{Instrumentation: how to measure end-to-end latency and reliability in R2X loops}
\label{sec:instrumentation}
A recurring source of ambiguity in closed-loop networking papers is \emph{where} the latency is measured.
In our R2X framing, the most meaningful metric is an end-to-end control loop delay (e.g., RTT in Demo~I and CTA in Demo~III),
because it directly governs whether an action is applied before the environment changes.
To ensure reproducibility, each run should use a consistent time-stamping policy:
(i) stamp sensing time at capture on the robot, (ii) stamp edge arrival time when the payload is decoded/consumed, and
(iii) stamp actuation time when the command is applied by the robot controller.
Similarly, reliability should be reported in a task-facing form (e.g., BLER target in Demo~II, UTFR in Demo~III) rather than only
link-layer packet loss, since the same packet loss can have different consequences depending on buffering and QoS.

\subsubsection{Safety and fallback behaviors under orchestration failures}
\label{sec:safety_fallback}
Because orchestration introduces additional decision logic, safe deployment requires explicit fallbacks when the network or compute
cannot meet the loop deadline.
A practical policy consistent with all demonstrations is:
if the orchestrator cannot provide a fresh command/config within a bounded deadline, the robot transitions to a conservative local mode
(e.g., slow-down/stop-and-go, short-horizon obstacle avoidance, or last-command hold with speed cap) until the loop recovers.
This also mitigates occasional LLM/LMM failures by ensuring that the \emph{worst-case} behavior is safe even when the best-case behavior
is globally optimized.

\subsubsection{Scalability checklist: extending from single-robot to multi-robot deployments}
\label{sec:scalability_checklist}
Scaling R2X from one or two agents to fleets requires controlling shared resources (uplink scheduling and edge compute) in addition to
per-agent sensing/codec settings.
From the perspective of Section~IV, the main missing coupling is \emph{contention}: as the number of robots grows, the orchestrator must
jointly decide \emph{who sends what and when} (semantic multiple access), while maintaining stability constraints on each robot's control loop.
This suggests a natural extension of the demo methodology: keep the same task-facing metrics (completion time, BLER/latency, CTA/UTFR),
but increase the number of concurrent agents and report how orchestration policies trade off fairness, stability, and overall throughput.

\subsection{Broader Use Cases: Collaborative Humanoid Workcells and Emerging Directions}
\label{sec:broader_usecases}

While Section~IV validates R2X via four end-to-end demonstrations, the same \emph{sense} $\rightarrow$ \emph{communicate} $\rightarrow$ \emph{compute} $\rightarrow$ \emph{act} loop naturally extends to broader settings where \emph{system-level KPIs}---not isolated model accuracy---govern success. To bridge our empirical findings with broader real-world applications, this section provides a conceptual extension of the R2X loop by identifying essential performance metrics and articulating the fundamental necessity of orchestration.

\paragraph{Canonical example: collaborative humanoid workcells in factories}
A widely discussed near-term deployment is a factory workcell where multiple humanoids (and/or arms on mobile bases) collaboratively execute repetitive manipulation: part fetching, fixturing, hand-offs, and two-agent co-manipulation of bulky objects. In such settings, the primary objective is not only low average delay, but \emph{predictable cycle time and safety} under shared wireless and shared edge compute. Representative task-facing KPIs include (i) workcell cycle time (s/part) or throughput (parts/hour), (ii) safety intervention rate (e.g., emergency stops or speed-limit triggers per hour), (iii) task error rate (e.g., failed grasps / misalignments per 1k operations), and (iv) tail control-loop delay (e.g., CTA 95/99/99.9th percentiles) that correlates with oscillation, overshoot, or failed handoffs.

\paragraph{Multi-rate loops: what must be local vs.\ what can be orchestrated.}
Humanoid platforms typically run fast inner control loops (e.g., whole-body or torque/impedance control at $\mathcal{O}(10^2\!-\!10^3)$~Hz) that must remain \emph{on-robot} for stability. R2X targets the \emph{outer orchestration loops} (tens of Hz to a few Hz) that decide: what sensing representation to send, how to allocate wireless/compute across robots, and when to replan or reassign roles. The right way to evaluate such a workcell is therefore not a single ``latency'' number, but a \emph{stack} of tail metrics that match each traffic class and control timescale.

\begin{table*}[t]
\centering
\caption{Representative quantitative KPIs for collaborative humanoid workcells (illustrative orders of magnitude).}
\label{tab:broader_usecases_kpi}
\small
\renewcommand{\arraystretch}{1.15}
\setlength{\tabcolsep}{5pt}
\begin{tabularx}{\textwidth}{p{0.22\textwidth} p{0.14\textwidth} p{0.14\textwidth} X}
\toprule
\textbf{Signal / loop (examples)} &
\textbf{Typical rate} &
\textbf{Typical payload} &
\textbf{What to report (task-facing, measurable)} \\
\midrule
Safety interlock (e-stop, proximity alert, speed limit) &
event + 10--100~Hz beacons &
10--200~B &
Tail deadline reliability, e.g., $99.999\%$ delivery within $5$--$10$~ms; bounded jitter; verified safe fallback if deadline missed. \\
\addlinespace
Coordination state (pose, grasp state, handoff intent) &
30--100~Hz &
0.2--2~KB &
Age-of-information (AoI) distribution and sync error (ms); $99$th\% AoI within $\sim\!20$--$50$~ms to avoid stale handoff decisions. \\
\addlinespace
Semantic perception (tracks, keypoints, embeddings, scene graph updates) &
10--30~Hz &
1--50~KB &
End-to-end tail delay (e.g., $95/99$th\% within $50$--$150$~ms) and its impact on cycle time / error rate under contention. \\
\addlinespace
High-fidelity video / logging / remote QA  &
5--30~Hz &
0.1--3~MB &
Sustained goodput (Mbps) and graceful degradation (best-effort acceptable) without harming safety/coordination traffic. \\
\bottomrule
\end{tabularx}
\end{table*}

\paragraph{Communication implications beyond ``make delay small.''}
Collaborative workcells require \emph{heterogeneous QoS} rather than a single low-latency target: safety messages demand ultra-high deadline reliability; coordination demands low AoI and tight time sync; perception updates demand controlled tail latency; and logging prefers throughput. \tcb{To handle these complex requirements, we need a communication system that is easy for the orchestrator to manage. This ``orchestration-friendly'' approach focuses on four key areas to keep the system running smoothly:}
\begin{itemize}
    \item \tcb{\textbf{Smart Data Mixing (Mixed-criticality)}: Combining urgent safety alerts and regular data on the same network without interference.}
    \item \tcb{\textbf{Reliable Multi-linking}: Using multiple signal paths at once to stay connected even when obstacles block one path.}
    \item \tcb{\textbf{Predictive Speed Control}: Anticipating network changes to prevent sudden slowdowns before they happen.}
    \item \tcb{\textbf{Coordinated Scheduling}: Letting the orchestrator decide precisely when each robot sends data to maintain total system efficiency.}
\end{itemize}

\paragraph{Sensing and computing trends that strengthen the case for orchestration.}
On the sensing side, the trend is toward richer multi-modal representations (e.g., vision+tactile for contact-rich manipulation, event/radar modalities for robustness, and 3D scene representations such as neural implicit maps or Gaussian-based reconstructions) that can be \emph{selectively} transmitted as compact semantics. On the computing side, modern robot stacks increasingly incorporate large foundation models and learned visuomotor policies (e.g., transformer-based visuomotor models, diffusion-style policies, and retrieval-augmented or tool-using planners), which make latency/compute costs highly variable. R2X naturally accommodates this variability by orchestrating model selection and placement (on-robot vs.\ edge), using cascades/early-exit policies and compression (quantization/distillation) when tail latency threatens CTA/AoI budgets.

\paragraph{Lessons.}
For broader deployments such as collaborative humanoid workcells, the key contribution of R2X is not a single new algorithm, but an explicit \emph{closed-loop orchestration interface} that ties measurable network/compute/sensing knobs to task-facing metrics (cycle time, safety intervention rate, AoI/CTA tails). This provides a concrete blueprint for future experimental validation: report mixed-criticality traffic metrics and their direct impact on collaboration throughput and safety, rather than only average link throughput or average latency.

\section{Research Directions \& Discussions}

Advancing R2X communication systems for LLM-enabled robotics requires addressing several key research directions that stem from the unique challenges posed by integrating large AI models into robotic applications. Recent developments in M2M and V2X communications provide a foundation, but further innovations are necessary to meet the specific demands of R2X interactions involving robots, humans, infrastructure, and mobile devices.

To make these directions actionable (beyond a bullet list), we phrase each item as a concrete \emph{research question (RQ)} together with \emph{how it can be validated} using measurable protocols. We emphasize task-facing KPIs---e.g., end-to-end latency/CTA, reliability (BLER/deadline-miss), and task success---rather than only average link throughput.

\begin{itemize}

    \item \textbf{Development of Advanced Communication Protocols}\\
    Traditional M2M and V2X protocols need to be enhanced to support the high data rates, low latency, and reliability required for transmitting complex multimodal data used by LLM-enabled robots~\cite{Lee2023Towards6GHyperConnectivity}. According to 3GPP~\cite{3gppurllc}, URLLC targets latency $< 1$~ms, reliability $> 99.999\%$, and data rates up to 100~Mbps.
    NOMA techniques \cite{Ding2017NOMA} and grant-free access protocols \cite{Liu2018MTC_GrantFree} can support massive connectivity (up to $10^6$ devices/km$^2$) with reduced overhead.\\
    \textbf{RQ1 (Protocol design under mixed criticality):} How can an R2X link simultaneously support mixed traffic classes---(i) safety/control messages requiring ultra-high deadline reliability and (ii) high-rate multimodal perception streams---without harming control stability?\\
    \textbf{How to verify:} Use a mixed-criticality benchmark with (a) periodic control packets (e.g., 50--200~B at 50--100~Hz) and (b) bursty perception updates (KB--MB). Report (i) deadline-miss probability for control (e.g., $<10^{-5}$ within 5--10~ms), (ii) tail latency (95/99/99.9\%) for control, (iii) task KPIs (CTA/RTT and success) in a closed-loop robot demo (Section~IV-style), and (iv) throughput fairness under $N$-robot contention.

    \item \textbf{Edge Computing and Distributed AI Integration}\\
    Leveraging MEC can significantly reduce latency and improve the responsiveness of robotic systems~\cite{Taleb2017MobileEdge,Zhang2021MEC_V2X}. MEC frameworks integrated into the 5G system architecture can achieve end-to-end (E2E) latencies as low as 5--10~ms, compared to traditional cloud-based latencies of $>50$~ms. Research should focus on developing efficient AI workload distribution algorithms between robots and edge servers, with support for dynamic task offloading, real-time inference splitting, and resource orchestration.\\
    \textbf{RQ2 (When does offloading truly help?):} Under what measurable conditions does edge/cloud offloading \emph{improve} task outcomes compared to on-device processing, once uplink/downlink delay and queueing are included?\\
    \textbf{How to verify:} For each task, measure the full critical path distribution $\{T_{\mathrm{uplink}},T_{\mathrm{edge}},T_{\mathrm{downlink}}\}$ and compare against the on-device alternative (e.g., local compute delay or collision/exploration waiting). Report the probability that the offloading inequality holds, e.g., $\mathbb{P}[T_{\mathrm{uplink}}+T_{\mathrm{edge}}+T_{\mathrm{downlink}} < T_{\mathrm{local}}]$, and correlate it with task success and tail CTA/RTT (as in Demo~I/III/IV). Stress-test under varying contention (multi-robot uplinks) and edge saturation (limited GPU slots).

    \item \textbf{Adaptive and Intelligent Communication Strategies}\\
    Integrating AI and ML techniques into communication systems can enable adaptive resource management, predictive maintenance, and enhanced situational awareness \cite{Wang2020AI_MTC,Ye2018ML_V2X}. Developing intelligent communication strategies that adjust to network conditions and application requirements can improve performance and reliability. These approaches are aligned with the direction of 3GPP TR 38.843~\cite{3gppai}, which studies the application of AI/ML to the NR air interface, including MAC/RRC optimization, link adaptation, and dynamic beam management~\cite{Kim2026EnvironmentAwareBeamManagement,Kim2026LMMAidedScheduling}.\\
    \textbf{RQ3 (Prediction vs.\ reaction under stale feedback):} How much can semantic/context-aware prediction reduce performance loss caused by delayed/stale CSI feedback (mobility, blockage, interference), compared to purely reactive adaptation?\\
    \textbf{How to verify:} Use a delayed-feedback test where the baseline adapts using $\gamma(t-d)$ and the proposed method predicts $\hat{\gamma}(t)$ from semantics. Report reliability violations (BLER>$0.1$), latency spikes, and tail throughput/latency under multiple delays ($d=1,3,5,10,\dots$), following Demo~II-style evaluation. Include ablations: (i) no semantics, (ii) semantics without prediction, (iii) prediction without semantics, (iv) full method.

    \item \textbf{Feature Vector Encoding and Data Compression}\\
    To efficiently transmit the rich multimodal data used by LLM-enabled robots, feature vector encoding approaches can be employed, which transmit the meaning of information rather than raw data \cite{Xie2021DeepLearningSemanticComm}. For instance,~\cite{Xie2021DeepLearningSemanticComm} demonstrates that semantic encoders can reduce required transmission rates by up to $90\%$ while maintaining task-relevant fidelity. A 512-dimensional feature vector, when quantized to 4-bit representation and compressed via entropy coding, can be transmitted at rates below $100$kbps. Research into advanced data compression and encoding techniques can further reduce bandwidth requirements without compromising the quality of information.\\
    \textbf{RQ4 (Task-aligned compression):} How should semantic compression be optimized so that bandwidth reduction \emph{maximizes task success} (not just pixel fidelity), especially when the downstream consumer is an MLLM (grounding, planning, orchestration)?\\
    \textbf{How to verify:} Replace MSE/PSNR-only evaluation with task metrics: (i) detection/grounding success (IoU-based), (ii) plan quality or action success, (iii) end-to-end task completion time. Sweep bitrate/latency constraints and report Pareto fronts (task success vs.\ payload vs.\ CTA/RTT). Include robustness checks: performance under packet loss and compression artifacts (e.g., reconstructed frames).

    \item \textbf{Security, Trust, and Robustness for MLLM-Driven R2X} \\
    Unlike conventional sensing/communication pipelines, MLLM-driven R2X introduces new attack surfaces across \emph{sensing}, \emph{communication}, \emph{computation}, and \emph{orchestration}. Robust deployment therefore requires explicit defenses against (i) malicious inputs to multimodal perception (e.g., sensor spoofing, adversarial patches, and poisoned camera/LiDAR streams), (ii) language-side attacks such as prompt injection or tool/API misuse that can alter the orchestrator’s decisions, and (iii) communication-layer threats including impersonation, replay, man-in-the-middle attacks, and jamming/interference.\\
    From a communication-system perspective, authentication and encryption must be treated as first-class design constraints rather than add-ons. Practical directions include lightweight mutual authentication for R2R/R2I links (device identity and key management), integrity-protected control messaging for orchestration outputs (e.g., signed JSON commands), and secure boot/attestation for edge servers that host MLLM inference.\\
    Beyond security, fault tolerance is critical for safety: the system should remain stable under packet loss, delayed/stale commands, or edge/cloud failures. A robust R2X stack should therefore include deadline-aware fallbacks (e.g., local safe-mode control, last-command hold with speed caps), redundancy via multi-connectivity or multi-path routing, and anomaly detection modules that monitor distribution shifts in sensor features and command streams.\\
    \textbf{RQ5 (Robustness as a measurable KPI):} What robustness guarantees (or empirical bounds) can an R2X system provide under adversarial sensing/prompt inputs and under link/edge failures, while maintaining safety constraints?\\
    \textbf{How to verify:} Use stress-test protocols beyond average throughput/latency: (a) adversarial robustness tests on perception/grounding (targeted occlusion/patch/poison), (b) prompt-injection tests for tool-using orchestrators, (c) fault-injection tests for link outages/edge compute drops, and (d) end-to-end safety KPI reporting (deadline miss rate, worst-case CTA/RTT tails, and task success under attack).

    \item \textbf{Standardization and Interoperability}\\
    With the diverse range of devices and systems involved in R2X communications, standardization efforts are crucial for ensuring interoperability. Building upon existing M2M (e.g., LTE-M and NB-IoT) and V2X (e.g., cellular-V2X based on LTE and NR-V2X) standards \cite{3gppTR38913,3gpp.ts23287}, new standards specific to R2X interactions involving robots should be developed, facilitating seamless integration and collaboration.\\
    \textbf{RQ6 (Interoperability under orchestration):} What is the minimal interoperable interface between robots, edge orchestrators, and network infrastructure so that sensing/comm/compute knobs can be exposed in a standard, vendor-agnostic way?\\
    \textbf{How to verify:} Define a reference ``orchestration API'' (e.g., schema-valid control messages for sensing mode, QoS class, and compute placement). Demonstrate cross-vendor feasibility in a testbed (at least two robot platforms or simulators + one edge orchestrator). Measure integration overhead (latency to apply config, schema compatibility rate) and confirm that system-level KPIs (CTA/RTT, reliability) remain stable when swapping components.

    \item \textbf{Latency Reduction Techniques}\\
    For applications requiring real-time responses, such as autonomous driving and industrial automation, minimizing latency is critical. Research into techniques such as mmWave communications, massive MIMO, and network slicing can contribute to achieving ultra-low latency communications \cite{cheng2018vehicle}. For instance, mmWave systems operating in the 28~GHz and 60~GHz bands can achieve one-way latencies as low as $1-2$~ms in line-of-sight scenarios.\\
    \textbf{RQ7 (Tail-latency control):} Which mechanisms reduce \emph{tail} latency (95/99/99.9\%) for closed-loop robot control more effectively than simply improving average rate?\\
    \textbf{How to verify:} Report tail metrics under controlled interference/mobility. Compare (i) URLLC-grade scheduling/slicing, (ii) multi-connectivity redundancy, and (iii) semantic payload reduction. Evaluate against control KPIs (CTA/overshoot/instability events) rather than only link-layer latency.

    \item \textbf{Scalability and Massive Connectivity}\\
    As the number of connected robots increases, ensuring the scalability of the communication systems is essential. \tcb{Thus, research to support} massive device connectivity while maintaining performance, through techniques like NOMA and efficient spectrum utilization, \tcb{is of importance} \cite{Ding2017NOMA}. For example, NOMA can accommodate up to $10^6$ devices/km$^2$ by enabling power-domain multiplexing, and simulations in~\cite{Ding2017NOMA} demonstrate that NOMA can achieve up to 3$\times$ higher spectral efficiency compared to traditional orthogonal schemes under high-density deployment scenarios.\\
    \textbf{RQ8 (Semantic multiple access and compute contention):} How should ``who sends what and when'' be scheduled when both wireless resources and edge compute are shared among many robots, without violating per-robot control deadlines?\\
    \textbf{How to verify:} Scale the number of agents $N$ in a multi-robot benchmark and measure (i) fairness, (ii) deadline-miss probability per robot, and (iii) aggregate task throughput. Include compute-aware scheduling: model edge GPU slots/queues and show how joint comm+compute scheduling changes task success compared to comm-only scheduling.

\item \textbf{\tcb{Compressing and Optimizing MLLMs}}\\
\tcb{MLLM}-based robot control and communication systems should strictly follow the QoS guarantee; for example, planning and communication time should not exceed the required time for each action. To minimize the \tcb{MLLM}-based planning time, optimization of \tcb{MLLM} itself is critical. Deep learning model compression techniques—such as quantization, pruning, and knowledge distillation—can significantly reduce computational latency and memory footprint. For example, knowledge distillation has been shown to compress large-scale models like LLaMA-7B into lightweight variants (e.g., TinyLLaMA-1.1B) with over $85\%$ reduction in model size and over 3$\times$ speedup in inference time.\\
    \textbf{RQ9 (QoS-constrained model selection):} How can we select/compress/partition MLLMs so that inference meets strict loop deadlines while preserving task-level reasoning quality?\\
    \textbf{How to verify:} For a fixed control deadline budget (e.g., CTA $\le$ 100~ms), benchmark (i) distillation/quantization levels, (ii) early-exit or cascaded models, and (iii) split inference (on-device+edge). Report task success vs.\ deadline-miss probability and energy cost, not only perplexity or offline accuracy.

\end{itemize}

\tcb{Looking further ahead, the research directions outlined above collectively point toward a long-term vision for R2X: fully autonomous, language-conditioned multi-robot ecosystems in which fleets of heterogeneous agents---humanoids, drones, vehicles, and manipulators---collaborate on complex missions specified entirely in natural language. In this vision, the R2X orchestration layer acts as the universal nervous system that seamlessly translates high-level human intent into real-time, network-aware decisions across sensing, communication, and computation. Realizing this vision will require not only advances in individual components (compressed MLLMs, semantic multiple access, and predictive link adaptation) but also principled co-design frameworks that provably satisfy safety, reliability, and latency constraints across the full stack. We hope that the research questions and validation protocols presented in this section provide a concrete roadmap toward this goal.}

\section{Conclusion}
This paper presented a comprehensive survey and tutorial on the communication systems and architectural frameworks required for generative AI-empowered R2X networks in robotics. Unlike earlier paradigms such as M2M or V2X, R2X emphasizes a holistic integration of communication, sensing, and computing resources to support the increasingly complex and collaborative tasks performed by robots. We examined distributed, hybrid, and fully centralized approaches to perception and control, and highlighted how these strategies can meet stringent QoS demands in diverse applications—from precision manufacturing and healthcare robotics to disaster-response UAV swarms.

A key observation is that task requirements, ranging from ultra-low-latency operations to handling large multimodal data streams, dictate architectural decisions. For instance, tasks that demand instantaneous responses or highly sensitive operations, such as tele-surgery, often rely on centralized or edge-based solutions for \tcb{MLLM} computations to achieve sub-5 ms latency and near-perfect reliability. In contrast, warehouse sorting or UAV mapping can benefit from hybrid frameworks, where on-device inference handles immediate actions while a centrally hosted \tcb{MLLM} supports complex semantic reasoning and global decision-making.

Our exploration also underscored the importance of feature vector encoding, data compression, and advanced security measures. Building on foundational M2M and V2X research, the R2X paradigm requires intensified efforts in standardization, interoperability, and integration of AI/ML techniques into communication protocols. Looking ahead, key research directions include developing advanced scheduling and resource allocation schemes, further optimizing LLM/\tcb{MLLM} models for real-time constraints, and establishing scalable federated learning frameworks to maintain adaptability and efficiency.

In summary, the R2X paradigm, supported by LLM-assisted hybrid and centralized communication architectures, promises to unlock unprecedented levels of robot cooperation, intelligence, and autonomy. By addressing the technical challenges outlined in this paper, future R2X networks will enable robots to thrive in complex, dynamic environments and deliver transformative benefits across multiple industries.

Finally, we emphasize that security, adversarial robustness, and fault tolerance are not optional: they are essential system constraints for MLLM-driven R2X deployment. Section~V explicitly outlines attack/defense considerations and measurable stress-test protocols to guide robust real-world adoption.

\bibliographystyle{IEEEtran}
\bibliography{references}

\end{document}